\documentclass[osajnl,twocolumn,showpacs,superscriptaddress,10pt]{revtex4-1} 
\usepackage{amsmath,amssymb,graphicx}
\usepackage{subfig}
\usepackage{amssymb}
\usepackage[notcite,notref,final]{showkeys}%
\usepackage{url,hyperref,multirow}
\usepackage{caption}
\usepackage{latexsym,bm}
\usepackage{color}
\usepackage{extarrows}
\usepackage{array}
\usepackage{delarray}
\usepackage{showkeys}
\usepackage{natbib}
\usepackage{booktabs}
\usepackage[boxed]{algorithm2e}

\makeatletter
\newcommand{\rmnum}[1]{\romannumeral #1}
\newcommand{\Rmnum}[1]{\expandafter\@slowromancap\romannumeral #1@}
\makeatother

\begin{document}

\title{Truncated Nuclear Norm Minimization for Image Restoration Based On Iterative Support Detection}


\author{Yilun Wang}
\author{Xinhua Su}

\affiliation{University of Electronic Science and Technology of China, Chengdu, Sichuan, 611731 China}


\begin{abstract}Recovering a large matrix from limited measurements is a challenging task arising in many real applications,
such as image inpainting, compressive sensing and medical imaging, and this kind of problems are mostly formulated as low-rank matrix
approximation problems. Due to the rank operator being non-convex and discontinuous, most of the recent theoretical studies use the nuclear
norm as a convex relaxation and the low-rank matrix recovery problem is solved through minimization of the nuclear norm regularized problem. However,
a major limitation of  nuclear norm minimization is that all the singular
values are simultaneously minimized and the rank may not be well approximated \cite{hu2012fast}. Correspondingly, in this paper, we propose a new multi-stage algorithm, which  makes use of the concept of Truncated Nuclear Norm Regularization (TNNR) proposed in \citep{hu2012fast} and Iterative Support Detection (ISD) proposed in \citep{wang2010sparse} to overcome the above limitation. 
Besides matrix completion problems considered in \citep{hu2012fast}, the proposed method can be also extended to the general low-rank matrix recovery problems. 
 Extensive experiments well validate the superiority of our new algorithms over other state-of-the-art methods. 
\end{abstract}

\maketitle 

\section{Introduction}
In many real applications such as machine learning \citep{abernethy2006low,amit2007uncovering,evgeniou2007multi}, computer vision \citep{tomasi1992shape}, and control
\citep{mesbahi1998rank}, etc.,  we seek to recover an unknown (approximately) low-rank matrix from limited information. 
This problem can be naturally formulated as the following model
\begin{equation}\label{1}
\min_{X} \;rank(X)\;\;\text{s.t.}\;\mathcal{A}X=b,
\end{equation}
where $X \in {\mathcal{R}}^{m \times n}$ is the decision variable, the linear map\;$\mathcal{A}$: $\mathcal{R}^{m \times n } \to \mathcal{R}^{p}(p<mn)$ and vector $b \in \mathcal{R} ^{p}$ are given. However, this is usually NP-hard due to the non-convexity and discontinuous nature of the rank function.
In paper \citep{fazel2003log}, Fazel et.al firstly solved rank minimization problem by approximating the rank function using the nuclear norm (i.e. the sum of singular values of a matrix). Moreover, theoretical studies show that the nuclear norm is the tightest convex lower bound of the rank function of matrices \citep{recht2010guaranteed}. Thus, an unknown (approximately) low-rank matrix $\bar{X}$ can be perfectly recovered by solving the optimization problem
\begin{equation}\label{convexmodel}
\min_{X} \;{\|X\|}_{*}\;\;\text{s.t.}\;\mathcal{A}X=b\doteq \mathcal{A}\bar{X},
\end{equation}
where $\|X\|_{*}=\sum_{i=1}^{min (m,n)}\sigma_{i}(X)$ is the nuclear norm and $\sigma_{i}(X)$ is the $i-th$ largest singular value of $X$,  
 under some conditions on the linear transformation $\mathcal{A}$. 

As a special case, the problem \eqref{convexmodel} is reduced to the well-known matrix completion problem \eqref{matrixcompletionconvexmodel} \citep{candes2009exact,candes2010power}, when $\mathcal{A}$ is a sampling (or projection/restriction) operator.
\begin{equation}\label{matrixcompletionconvexmodel}
\min_{X} \;{\|X\|}_{*}\;\;\text{s.t.}\;X_{i,j}=M_{i,j}, (i,i)\in\Omega,
\end{equation}
where $M\in \mathcal{R}^{m\times n}$ is the incomplete data matrix and $\Omega$ is the set of locations corresponding to the observed entries.
To solve the kind of problems, we can refer to \citep{wright2009robust,toh2010accelerated,candes2009exact,candes2010power,cai2008restoration,cai2010singular} for some breakthrough results. Nevertheless, they may obtain suboptimal performance in real applications because the nuclear norm may not be a good approximation to rank-operator, because  all the non-zero singular values in rank-operator have the equal contribution, while the singular values in nuclear norm are treated differently by adding them together. 
Thus, to overcome the weakness of nuclear norm, Truncated Nuclear Norm Regularization (TNNR) was proposed for matrix completion, which only minimizes the smallest $ min(m, n)-r $  singular values \citep{hu2012fast}. The similar truncation idea was also proposed in our previous work \citep{wang2010sparse}.  Correspondingly, the problem can be formulated as
\begin{equation}\label{truconvexmodel}
\min_{X} \;{\|X\|}_{r}\;\;\text{s.t.}\;X_{i,j}=M_{i,j}, (i,i)\in\Omega,
\end{equation}
where ${\|X\|}_{r}$ is defined as the sum of $min(m, n)-r$ minimum singular values. In this way, ones can get a more accurate and robust approximation to the rank-operator on both synthetic and real visual data sets.

In this paper, we aim to extend the idea of TNNR from the special matrix completion problem to the general problem \eqref{convexmodel} and give the corresponding fast algorithm. More important, we will consider how to fast estimate $r$, which is usually unavailable in practice.

Throughout this paper, we use the following notation. We let $<\cdot,\cdot>$ be the standard inner product between two matrices in a finite dimensional Euclidean space, $\|\cdot\|$ be the 2-norm, and $\|\cdot\|_{F}$ be the Frobenius norm for matrix variables. The projection operator under the Euclidean distance measure is denoted by $\mathcal{P}$ and the transpose of a real matrix by $\intercal$. Let $X=U\Sigma V^{\intercal}$ is the singular value decomposition (SVD) for $X$, where $\Sigma=diag(\sigma_{i})$, $1\leq i\leq min\{m,n\}$, and $\sigma_{1}\geq\cdots\geq\sigma_{min\{m,n\}}$.

%
%
%
%
%
%

\subsection{Related Work}\label{rework}
The low-rank optimization problem \eqref{convexmodel} has attracted more and more interests in developing customized algorithms, particularly for lager-scale cases. We now briefly review some influential approaches to these problems.

The convex problem \eqref{convexmodel} can be easily reformulated into the semi-definite programming (SDP) problems \citep{fazel2001rank,srebro2004maximum} to make use of the generic SDP solvers such as   SDPT3 \citep{yuan2009sparse} and SeDuMi \citep{sturm1999using} which are based on the interior-point method. However, the interior-point approaches suffer from the limitation that they ineffectively handle large-scale problems which was mentioned in \citep{recht2010guaranteed,ji2009accelerated,pong2010trace}. The problem \eqref{convexmodel} can also be solved through a projected subgradient approach in \citep{recht2010guaranteed}, whose major computation is concentrated on singular values decomposition. The method can be used to solve large-scale cases of \eqref{convexmodel}. However, the convergence may be slow, especially when high accuracy is required. In \citep{recht2010guaranteed,wen2010alternating}, UV-parametrization $(X=UV^{\intercal})$ based on matrix factorization is applied in general low-rank matrix reconstruction problems. Specifically, the low-rank matrix $X$ is decomposed into the form $UV^{\intercal}$, where $U\in \mathcal{R}^{m\times r}$ and $V\in \mathcal{R}^{n\times r}$ are tall and thin matrices. The method reduces the dimensionality from $mn$ to $(m+n)r$. However, if the rank and the size are large, the computation cost may also be very high. Moreover, the rank $r$ is not known as a priori for most of applications, and it has to be estimated or dynamically adjusted, which might be difficult to realize. More recently, the augmented Lagrangian method (ALM) \citep{hestenes1969multiplier,Powe69a} and the alternating direction method of multipliers (ADMM) \citep{gabay1976dual} are very efficient for some convex programming problems arising from various applications. In \citep{yang2013linearized}, ADMM is applied to solve \eqref{convexmodel} with $\mathcal{A}\mathcal{A}^{*}=\mathcal{I}$.

As an important special case of problem \eqref{convexmodel},  the matrix completion problem \eqref{matrixcompletionconvexmodel} has been widely studied.  Cai et al.\citep{cai2008restoration,cai2010singular} used the shrinkage operator to solve the nuclear norm effectively. The shrinkage operator applies a soft-thresholding rule to singular values, as the sparse operator of a matrix, though it can be applied widely in mangy other approaches.
However, due to the above-mentioned limitation of nuclear norm, TNNR \eqref{truconvexmodel} is proposed to replace the nuclear norm. Since $\|X\|_{r}$ in \eqref{truconvexmodel} is non-convex, it can not be solved simply and effectively. So, how to change \eqref{truconvexmodel} into a convex function is critical. Obviously, it is noted that $\|X\|_{r}=\|X\|_{*}-\sum_{i=1}^{r}\sigma_{i}(X), Tr(L_{r}XR_{r}^{\intercal})=\sum_{i=1}^{r}\sigma_{i}(X)$, where $U\Sigma V^{\intercal}$ is the SVD of $X$, $U=(u_{1},\cdots, u_{m})\in \mathcal{R}^{m\times m}$, $\Sigma\in \mathcal{R}^{m\times n}$ and $V =(v_{1},\cdots, v_{n})\in \mathcal{R}^{n\times n}$. Then $L_{r} = {(u_{1},\cdots, u_{r})}^{T}, R_{r} = {(v_{1},\cdots, v_{r})}^{T} $ and the optimization problem \eqref{truconvexmodel} can be rewritten as:
\begin{equation}\label{promodle}
\min_{X} \;{\|X\|}_{*}-Tr(L_{r}XR_{r}^{\intercal})\;\;\text{s.t.}\;X_{i,j}=M_{i,j}, (i,j)\in\Omega,
\end{equation}
While the problem \eqref{promodle} is still non-convex, they can get a local minima by an iterative procedure proposed in \citep{hu2012fast} and we will review the procedure in more details later.

A similar idea of truncation in the context of the sparse vectors, is also implemented on the sparse signals by our previous work in \cite{wang2010sparse}, which tries to adaptively learn the information of the nonzeros of the unknown true signal. Specifically, we present a sparse signal reconstruction method, Iterative Support Detection (ISD, for short), aiming to achieve fast reconstruction and a reduced requirement on the number of measurements compared to the classical $l_{1}$ minimization approach. ISD alternatively calls its two components: support detection and signal reconstruction. From an incorrect reconstruction, support detection identifies an index set $I$ containing some elements of $supp(\bar{x}) = \{i : \bar{x}_{i}\neq 0\}$, and signal reconstruction solves
\begin{equation*}
\min_{x} \;{\|x_{T}\|}_{1}\;\;\text{s.t.}\;\mathcal{A}x=b\doteq \mathcal{A}\bar{x},
\end{equation*}
where $T = I^{C}$ and ${\|x_{T}\|}_{1}=\sum_{i\notin T}|x_{i}|$.  To obtain the reliable support detection from inexact reconstructions, ISD must take advantage of the features and prior information about the true signal $\bar{x}$. In \cite{wang2010sparse}, the sparse or compressible signals with components having a fast decaying distribution of nonzeros, are considered. 

\subsection{Contributions and Paper Organization}\label{1.B}

Our first contribution is the estimation of $r$, on which TNNR heavily depends  (in $\|X\|_{r}$).  Hu et.al \citep{hu2012fast} seeks the best $r$  by trying all the possible values and this leads high computational cost.
%
In this paper, motivated by Wang et.al \cite{wang2010sparse}, we propose singular value estimation (SVE) method to obtain the best $r$,  which can be considered as  a special implementation of iterative support detection of \cite{wang2010sparse} in case of matrices.

Our second contribution is to extend TNNR from matrix completion to the general low-rank cases. In \citep{hu2012fast}, they have only considered the matrix completion problem. 

The third contribution is based on the above two. In particular, a new efficient algorithmic framework is proposed for the low-rank matrix recovery problem. We name it LRISD, which iteratively calls its two
components: SVE  and solving the low-rank matrix reconstruction model based on TNNR.


The rest of this paper is organized as follows. In Section \ref{threemodels}, computing framework of LRISD and theories of SVE are introduced. In Section \ref{f-thridmodel}, Section \ref{semodel} and Section \ref{moxing3}, we introduce 
three algorithms to solve the problems mentioned in subsection \eqref{3.3}. Experimental results are presented in Section \ref{experiment}.  Finally,  some conclusions are made in Section \ref{Conclusion}.
\section{\bf{Iterative Support Detection for Low-rank Problems }}\label{threemodels}

In this section, we first give the outline of the proposed algorithm LRISD and then  elaborate the proposed SVE method which is a main component of  LRISD. 
\subsection{\bf{Algorithm Outline}}\label{3.3}

The main purpose of LRISD is to provide a better approximation to the model \eqref{1} than the common convex relaxation model \eqref{convexmodel}. The key idea is to make use of the Truncated Nuclear Norm Regularization (TNNR) defined in \eqref{truconvexmodel} and its variant \eqref{promodle} \cite{hu2012fast}. While ones can  passively try all the possible values of $r$ which is 
the number of largest few singular values, we proposed to actively estimate the value of $r$.

In addition, we will consider the general low-rank recovery problems beyond the matrix completion problem. Specifically,  we will  solve three models equality-model \eqref{fmodelNon-convex}, unconstrained-model \eqref{secondmodelNon-convex} and inequality-model \eqref{thridmodelNon-convex}:
\begin{equation}\label{fmodelNon-convex}
\min_{X} \;{\|X\|}_{*}-Tr(L_{r}XR_{r}^{\intercal} )\;\text{s.t.}\;\mathcal{A}X=b,
\end{equation}
\begin{equation}\label{secondmodelNon-convex}
\min_{X} \;{\|X\|}_{*}-Tr(L_{r}XR_{r}^{\intercal} )+\frac{\mu}{2}\|\mathcal{A}X-b\|^{2},
\end{equation}
\begin{equation}\label{thridmodelNon-convex}
\min_{X} \;{\|X\|}_{*}-Tr(L_{r}XR_{r}^{\intercal} )\;\text{s.t.}\;\|\mathcal{A}X-b\|\leq\delta,
\end{equation}
where $\mu>0$ and $\delta>0$ are parameters reflecting the level of noise.  The models \eqref{secondmodelNon-convex} and \eqref{thridmodelNon-convex} consider the case with noisy data. 
%
 Here $\mathcal{A}$ is a linear mapping such as partial discrete cosine transformation (DCT), partial discrete walsh hadamard transformation (DWHT), discrete fourier transform (DFT). 


The general framework of LRISD, as an iterative procedure,  
starts from the initial $r=0$, i.e. solving a plain nuclear norm minimization problem, and then estimates $r$ based on the recovered result. Based on the estimated $r$, we solve a resulted TNNR model \eqref{fmodelNon-convex}, or \eqref{secondmodelNon-convex} or \eqref{thridmodelNon-convex}. Using the new recovered result, we can update the $r$ value and solve a new TNNR model \eqref{fmodelNon-convex}, or \eqref{secondmodelNon-convex} or \eqref{thridmodelNon-convex}. Our algorithm iteratively calls the $r$ estimation and the solver of the TNNR model. 


As for solving \eqref{fmodelNon-convex}, \eqref{secondmodelNon-convex} and \eqref{thridmodelNon-convex}, we following the idea of \cite{hu2012fast}. Specifically,  A simple but efficient iterative procedure is adopted to decouple the $L_r$, $X$ and $R_r$. We set the initial guess $X_1$. In the $l$-th iteration, we first fix $X_l$ and compute $L_r^l$ and $R_r^l$ as described in  \eqref{promodle}, based on the SVD of $X_l$.
 Then fix $L_r^l$ and $R_r^l$ to update $X_{l+1}$ by solving the following problems, respectively:

\begin{equation}\label{fmodel}
\min_{X} \;{\|X\|}_{*}-Tr(L_{r}^l X{R_{r}^l}^{\intercal} )\;\text{s.t.}\;\mathcal{A}X=b,
\end{equation}
\begin{equation}\label{secondmodel}
\min_{X} \;{\|X\|}_{*}-Tr(L_{r}^l X{R_{r}^l}^{\intercal} )+\frac{\mu}{2}\|\mathcal{A}X-b\|^{2},
\end{equation}
\begin{equation}\label{thridmodel}
\min_{X} \;{\|X\|}_{*}-Tr(L_{r}^l X{R_{r}^l}^{\intercal} )\;\text{s.t.}\;\|\mathcal{A}X-b\|\leq\delta.
\end{equation}
In \cite{hu2012fast}, the authors have studied to solve the special case of matrix completion problems. For the general problems \eqref{fmodel}, \eqref{secondmodel} and \eqref{thridmodel}, we will extend the current state of the art algorithms to solve them in Section \ref{f-thridmodel}, Section \ref{semodel} and Section \ref{moxing3}, respectively.

In summary,  the procedure of LRISD, as a new multi-stage algorithm,  is summarized in the following Algorithm $1$.  By alternately running the SVE and solving the corresponding TNNR models, the iterative
scheme will converge to a solution of a TNNR model, whose solution is expected to be better than that of the plain nuclear norm minimization model \eqref{convexmodel}.
%
%
%
\vspace{2mm} \hrule \hrule
\vspace{2mm} {\bf  Algorithm 1: LRISD based on  \eqref{fmodel}, \eqref{secondmodel} and \eqref{thridmodel}}
\begin{enumerate}
\item[1.] Initialization:~ set $X_{re}=X_{0}$, which is the solution of pure nuclear norm regularized model \eqref{convexmodel}.
\item[2.] Repeat until $r$ reaches a stable value.
\begin{enumerate}
\itemsep1em
\item[Step 1.]  Estimate ${r}$ via SVE on $X_{re}$. 
\item[Step 2.]  Initialization:~$X_{1}=Data$ (the matrix form of $b$).\\
In the $l-th $ iteration:
\begin{enumerate}
\item[a)] Compute $L_{r}^l$ and ${R_{r}^l}$ of \eqref{fmodel} (\eqref{secondmodel} or \eqref{thridmodel}) according to the current $X_{l}$ in the same way as \eqref{promodle} mentioned.
\item[b)] Solve the model  \eqref{fmodel} (\eqref{secondmodel} or \eqref{thridmodel}) and obtain $X_{l+1}$. Goto a) until \\ $\|X_{l+1}-X_{l}\|_{F}^{2}/\|Data\|_{F}^{2}\leq\varepsilon_{1}$.
\item[c)] $l \leftarrow l+1$.

\end{enumerate}
\item[Step 3.]  Set $X_{re}=X_{l+1}$. 
\end{enumerate}
\item[3.]  Return the recovered matrix $X_{re}$.
\end{enumerate}
\vspace{2mm} \hrule \hrule \vspace{5mm}

In the following content, we will explain the implementation of  SVE of Step $1$ in more details, and  extend the existing algorithms for nuclear norm regularized models to TNNR based  models \eqref{fmodel}-\eqref{thridmodel} in the procedure b) of Step 2. 
\subsection{Singular Value Estimate} \label{Sec:SVE}

In this subsection, we mainly focus on Step $1$ of LRISD, i.e. describing the process of SVE to estimate $r$, which is the number of largest few singular values.  While it is feasible to find the best $r$ via trying all possible $r$ as done in \citep{hu2012fast},  this procedure is not computationally efficient. Thus,  we aim to quickly give an estimate of the best $r$. 

\begin{figure}[!th]
\centering
\subfloat[an image example]{\includegraphics[height=2.8cm,width=0.23\textwidth]{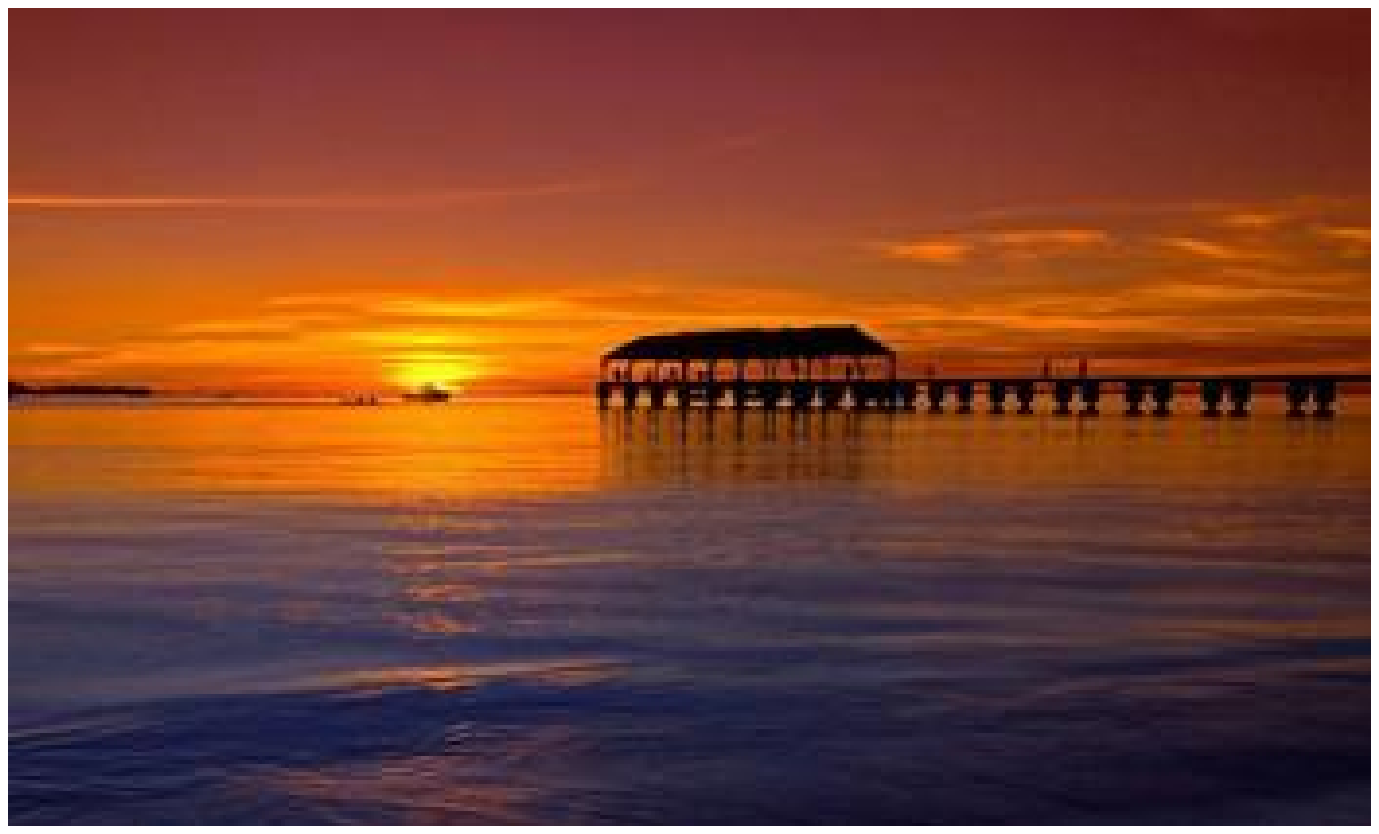}}
\hfil
\subfloat[red channel]{\includegraphics[height=2.8cm,width=0.23\textwidth]{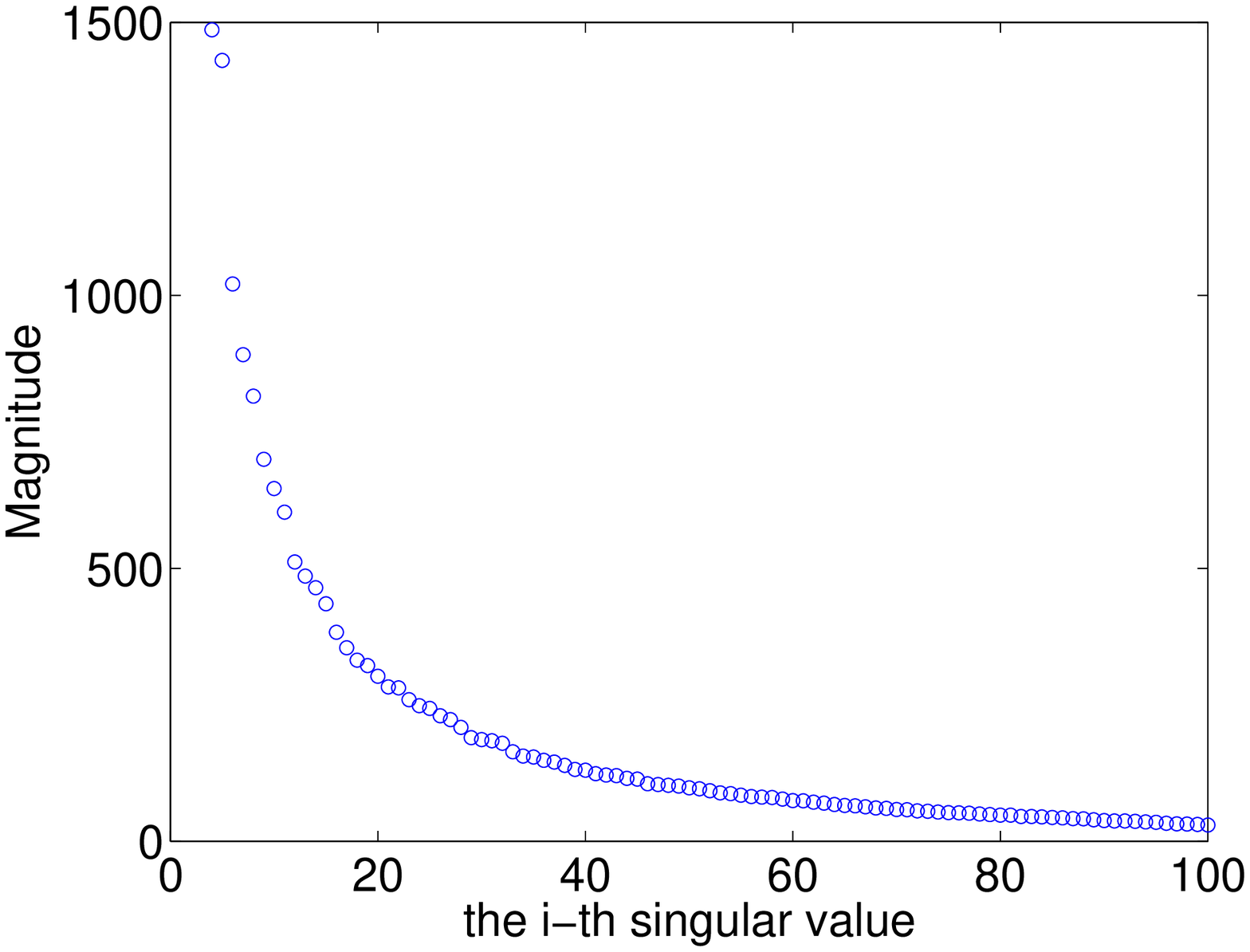}}\\
\hfil
\subfloat[green channel]{\includegraphics[height=2.8cm,width=0.23\textwidth]{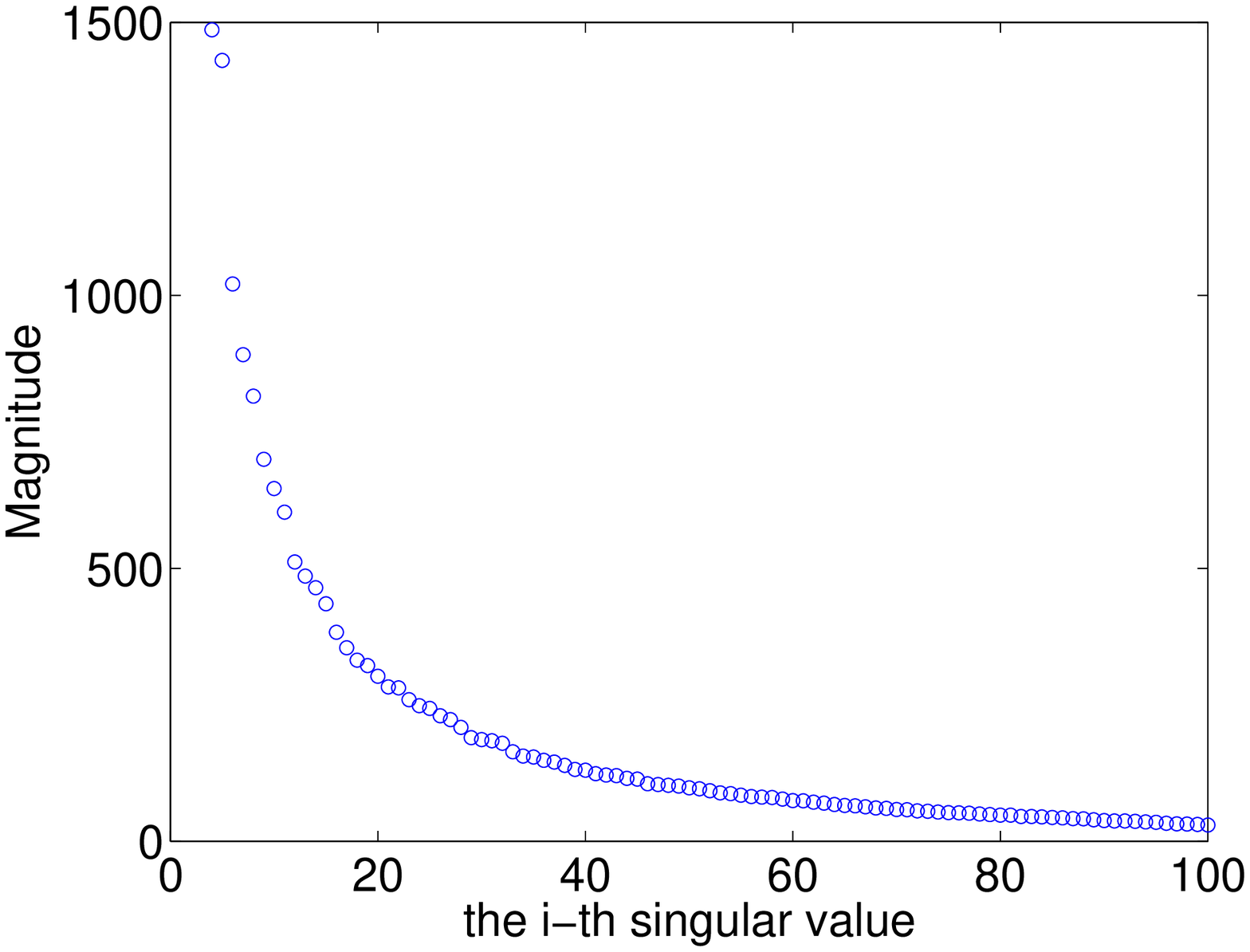}}
\hfil
\subfloat[blue channel]{\includegraphics[height=2.8cm,width=0.23\textwidth]{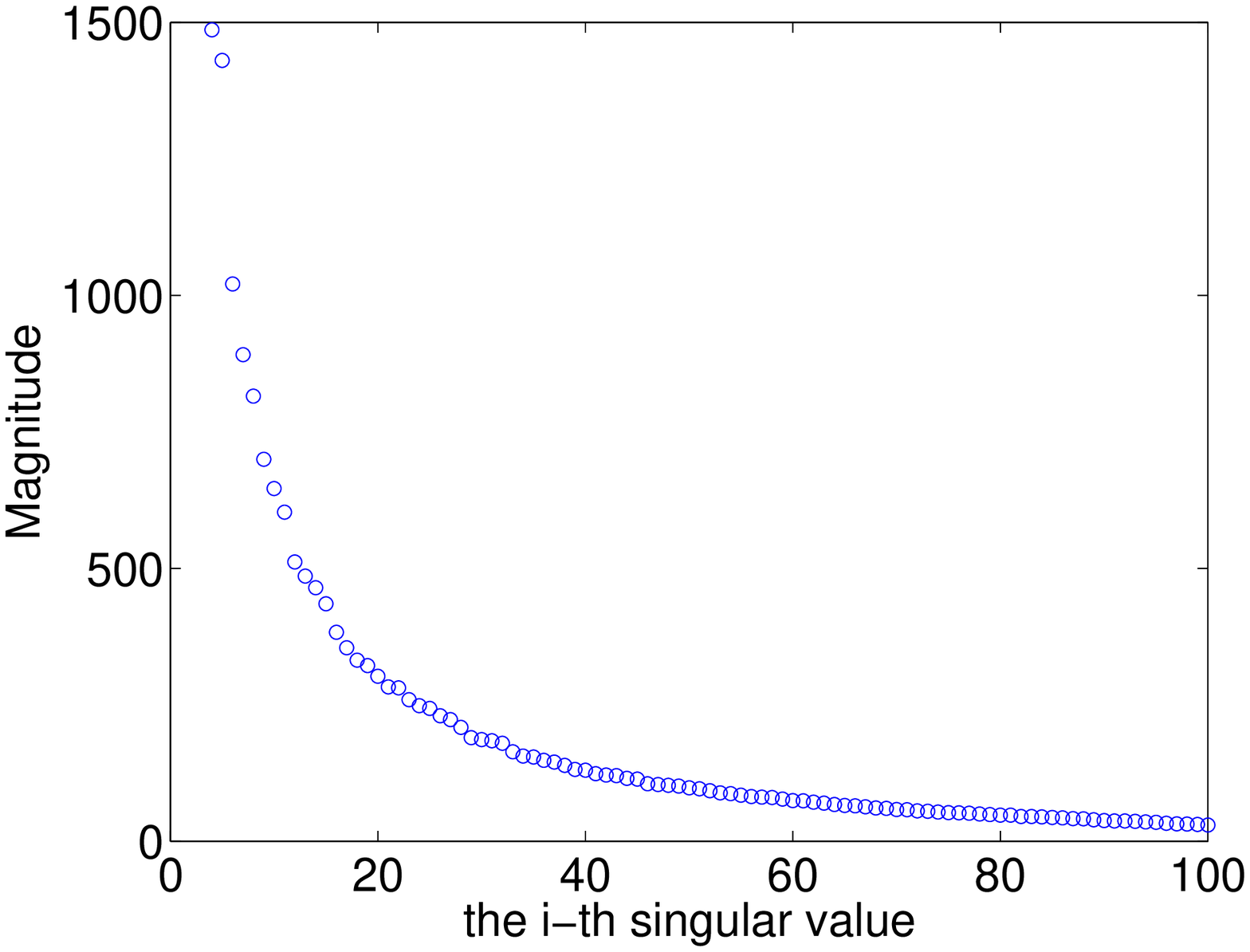}}
\caption{\small (a) A $350\times210$ image example. (b) The singular values of red channel. (c) The singular values of green channel. (d) The singular values of blue channel. In order to illustrate the distribution clearly , images (b) (c) (d)  are used for showing the magnitude of singular values from  the 4-th  to the 100-th in each channel.}
\label{fig:1}
\end{figure}

As we have known, for (approximately) low-rank matrices or images, these singular values often have a feature that they all have a fast decaying distribution (as showed in Fig  \ref{fig:1}). To take advantage of this feature,  we can extend our previous work ISD \citep{wang2010sparse} from detecting the large components of sparse vectors to the large singular values of low-rank matrices. In particular, 
SVE is nothing but  a specific implementation of support detection in cases of low-rank matrices, with the aim to acquire the estimation of the true $r$.

 Now we present the process of SVE and the effectiveness of SVE. It is noted that as showed in the Algorithm $1$,  SVE is repeated several times until a stable estimate $r$ is obtained. For each time, given the reference image $X_{re}$,
%
we can obtain the singular value vector $S$ of $X_{re}$ by SVD. A natural way to find the positions of the true large singular values based on $S$, which is considered as  an estimate of the singular value  vector of the true matrix $\bar{X}$, is based on thresholding
$$
I:=\{i: S_i > \epsilon \}
$$
due to the fast decaying property of the singular values. The choice of $\epsilon$ should be case-dependent. In the spirit of ISD, one can  use the so called ``last significant jump" rule to set the threshold value $\epsilon$ to detect the large singular values and minimized the false detections, if we assume that  the components of $S$ are sorted from large to small.  The straightforward way to apply the  ``last significant jump" rule is to look for the largest $i$ such that
$$ St_{(i)}\doteq |S_{i}-S_{i+1}|> \tau,$$ where $\tau$ is a prescribed value, and $St$ is defined as absolute values of the first order difference of $S$. This amounts to sweeping the decreasing sequence $\{S_{i}\}$ and look for the last jump larger than $\tau$. For example, the selected $i=4$, then we set $\epsilon=S_{4}$,

However, in this paper, unlike the original ISD paper \citep{wang2010sparse},  we propose to apply the ``last significant jump" rule on absolute values of the first order difference of $S$, i.e., $St$, instead of $S$. Specifically,
we look for the largest $i$ such that
$$ Stt_{(i)}\doteq |St_{i+1}-St_{i}|> \kappa,$$ where $\kappa$ will be selected below, and $Stt$ is defined as absolute values of the second order difference of $S$ .  This amounts to sweeping the decreasing sequence $\{St_{i}\}$ and look for the last jump larger than $\kappa$. For example, the selected $i=4$, then we set $\epsilon=S_{4}$. 
%
We set the estimation rank $\tilde{r}$ to be the cardinality of $I$, or a close number to it. 

Specifically, $St$ is computed to obtain jump sizes which count on the change of two neighboring components of $S$. Then, to reflect the stability of these jumps, the difference of $St$  need to be considered as we just do,  
because the few largest singular values jump actively, while the small singular values would not change much.  The cut-off threshold 
$\kappa$ is determined via certain  heuristic methods in our experiments: synthetic and real visual data sets.  Note that in subsection \ref{subsection6F}, we will present a reliable rule for determining threshold value $\kappa$.

\section{\bf{TNNR-ADMM For \eqref{fmodel} and \eqref{thridmodel}}}\label{f-thridmodel}

In this section, we extend the existing  ADMM method in \citep{gabay1976dual} originally for the nuclear norm regularized model to solve \eqref{fmodel} and \eqref{thridmodel} under common linear mapping $\mathcal{A}$ ($\mathcal{A}\mathcal{A}^{*}=\mathcal{I}$), and give closed-form solutions. 
 The extended verison of ADMM is named as TNNR-ADMM, and the original ADMM for the corresponding nuclear norm regularized low-rank matrix recovery model is denoted as LR-ADMM. In addition, we can deduce that the resulting subproblems are simple enough to have closed-form solutions and can be easily achieved to high precision. We start this section with some preliminaries which are convenient for the presentation of algorithms later.

When $\mathcal{A}\mathcal{A}^{*}=\mathcal{I}$, we present the following conclusions \citep{yang2013linearized}:
\begin{equation}\label{inverse}
(\mathcal{I}+\alpha \mathcal{A}^{*}\mathcal{A})^{-1}=\mathcal{I}-\frac{\alpha}{1+\alpha}\mathcal{A}^{*}\mathcal{A},
\end{equation}
where $(\mathcal{I}+\alpha \mathcal{A}^{*}\mathcal{A})^{-1}$ denotes the inverse operator of $(\mathcal{I}+\alpha \mathcal{A}^{*}\mathcal{A})$ and $\alpha >0$.\\
\textbf{Definition 3.1}(\citep{yang2013linearized}): When $\mathcal{A}$ satisfies $\mathcal{A}\mathcal{A}^{*}=\mathcal{I}$, for $\delta\geq0$ and $Y\in{\mathcal{R}}^{m \times n}$, the projection of $Y$ onto $B_{\delta}$ is defined as
\begin{equation*}
\mathcal{P}_{B_{\delta}}(Y)=Y+\frac{\eta}{\eta+1}\mathcal{A}^{*}(b-\mathcal{A}Y),
\end{equation*}
where
\begin{eqnarray*}
\eta=max\{\|\mathcal{A}Y-b\|/\delta-1,0\},\\
B_{\delta}=\{U\in{\mathcal{R}}^{m \times n}:\|\mathcal{A}U-b\|\leq\delta\}.
\end{eqnarray*}
In particular, when $\delta=0$,
\begin{equation*}
\mathcal{P}_{B_{0}}=Y+\mathcal{A}^{*}(b-\mathcal{A}Y),
\end{equation*}
where
\begin{equation*}
B_{0}=\{U\in{\mathcal{R}}^{m \times n}:\mathcal{A}U=b\}.
\end{equation*}
Then, we have the  following conclusion:
\begin{equation*}
\mathcal{P}_{B_{\delta}}(Y)=\arg\min_{X\in R^{m\times n}}\{\|X-Y\|_{F}^{2}:\|\mathcal{A}X-b\|\leq\delta\}.
\end{equation*}
\textbf{Definition 3.2}: For the matrix $X\in{\mathcal{R}}^{m \times n}$, $X$ have the singular value decomposition as following: $X=U\Sigma V^{\intercal}$, $\Sigma=diag(\sigma_{i})$. The shrinkage operator $\mathcal{D}_{\tau}(\tau>0)$ is defined:
\begin{equation*}
\mathcal{D}_{\tau}(X)=U\mathcal{D}_{\tau}(\Sigma)V^{\intercal}\;,\mathcal{D}_{\tau}(\Sigma)=diag(\{\sigma_{i}-\tau\}_{+}),
\end{equation*}
where $(s)_{+}=\max\{0,s\}$.\\
\textbf{Theorem 3.2}(\citep{cai2010singular}): For each $\tau\geq0$ and $Y\in{\mathcal{R}}^{m \times n}$, we have the following conclusion:
\begin{equation*}
\mathcal{D}_{\tau}(Y)=\arg\min_{X}\frac{1}{2}\|X-Y\|_{F}^{2}+\tau\|X\|_{*}.
\end{equation*}
\textbf{Definition 3.3}: Denote $\mathcal{A}^{*}$ be the adjoint operator of $\mathcal{A}$ satisfying the following condition:
\begin{equation}\label{adjoint}
\langle\mathcal{A}(X),Y\rangle=\langle X,\mathcal{A}^{*}(Y)\rangle.
\end{equation}

\subsection{\bf{Algorithmic Framework}}

The problems \eqref{fmodel} and \eqref{thridmodel} can be easily reformulated into the following linear constrained convex problem:
\begin{equation}\label{ffgmodel}
\begin{split}
\min_{X} \;{\|X\|}_{*}-Tr(L_{r}^lY{R_{r}^l}^{\intercal} )\\
\text{s.t.}\;X=Y,\;Y\in B_{\delta},
\end{split}
\end{equation}
where $\delta\geq0$. In particular, the above formulation is equivalent to \eqref{fmodel} when $\delta=0$. The augmented Lagrangian function of \eqref{ffgmodel} is:
\begin{equation} \label{fflmodel}
\begin{split}
L(X,Y,Z,\beta)&={\|X\|}_{*}-Tr(L_{r}^lY{R_{r}^l}^{\intercal})\\
&+\frac{\beta}{2}\|X-Y\|_{F}^{2}- \langle  Z, X-Y \rangle,
\end{split}
\end{equation}
where $Z\in{\mathcal{R}}^{m \times n}$ is the Lagrange multiplier of the linear constraint, $\beta>0$ is the penalty parameter for the violation of the linear constraint.

The idea of ADMM is to decompose the minimization task in \eqref{fflmodel} into three easier and smaller subproblems such that the involved variables X and Y can be minimized separately in altering order. In particular, we apply ADMM to solve \eqref{fflmodel}, and obtain the following iterative scheme:
\begin{equation}\label{frameadmm}
\left\{
\begin{split}
&X_{k+1}=\arg\min_{X\in \mathcal{R}^{m\times n}} \{L(X,Y_{k},Z_{k},\beta)\},\\
&Y_{k+1}=\arg\min_{Y\in B_{\delta}} \{L(X_{k+1},Y,Z_{k},\beta)\},\\
&Z_{k+1}=Z_{k}-\beta(X_{k+1}-Y_{k+1}).
\end{split}
\right.
\end{equation}

Ignoring constant terms and deriving the optimal conditions for the involved subproblems in \eqref{frameadmm}, we can easily verify that the iterative scheme of the TNNR-ADMM approach for \eqref{fmodel} and \eqref{thridmodel} is as follows.
\vspace{2mm} \hrule \hrule
\vspace{2mm} {\bf  Algorithm 2: TNNR-ADMM for \eqref{fmodel} and \eqref{thridmodel}}
\begin{itemize}
\item[1.] Initialization:~ set $X_{1}=Data$ (the matrix form of $b$), $Y_{1}=X_{1},Z_{1}=X_{1}$, and input $\beta$.
\item[2.] For $k=0,1,\cdots N$ (Maximum number of iterations),
\begin{itemize}
\item[(\rmnum{1})] Update $X_{k+1}$ by
\begin{equation}\label{ftsub1}
X_{k+1}=\arg\min_{X} \|X\|_{*}+\frac{\beta}{2}\|X-(Y_{k}+\frac{1}{\beta}Z_{k})\|_{F}^{2}.
\end{equation}
\item[(\rmnum{2})] Update $Y_{k+1}$ by
\begin{equation}\label{ftsub2}
Y_{k+1}=\arg\min_{Y\in B_{\delta}}\frac{\beta}{2}\|Y-(X_{k+1}+\frac{1}{\beta}({L_{r}^l}^{\intercal} {R_{r}^l}-Z_{k}))\|_{F}^{2}.
\end{equation}
\item[(\rmnum{3})] Update $Z_{k+1}$ by
\begin{equation}\label{ftsub3}
Z_{k+1}=Z_{k}-\beta(X_{k+1}-Y_{k+1}).
\end{equation}
\end{itemize}
\item[3.] End the iteration till\\
 $\|X_{k+1}-X_{k}\|_{F}^{2}/\|Data\|_{F}^{2}\leq\varepsilon_{2}$.
\end{itemize}
\vspace{2mm} \hrule \hrule \vspace{5mm}

\subsection{\bf{The Analysis of Subproblems}}

According to the analysis above, the computation of each iteration of TNNR-ADMM approach for \eqref{fmodel} and \eqref{thridmodel} is dominated by solving the subproblems \eqref{ftsub1} and \eqref{ftsub2}. We now elaborate on the strategies for solving these subproblems based on abovementioned preliminaries.

First, the solution of \eqref{ftsub1} can be obtained explicitly via \text{Theorem 3.2}:
\begin{equation*}
X_{k+1}=\mathcal{D}_{\frac{1}{\beta}}(Y_{k}+\frac{1}{\beta}Z_{k}),
\end{equation*}
which is the closed-form solution.

Second, it is easy to obtain:
\begin{equation}\label{20}
Y_{k+1}=X_{k+1}+\frac{1}{\beta}({L_{r}^l}^{\intercal }{R_{r}^l}-Z_{k}),\;\;Y_{k+1}\in B_{\delta}.
\end{equation}
Combining \eqref{20} and equipped with \text{Definition 3.1}, we give the final closed-form solution of the subproblem \eqref{ftsub2}:
\begin{equation*}
Y_{k+1}=Y_{k+1}+\frac{\eta}{\eta+1}\mathcal{A}^{*}(b-\mathcal{A}Y_{k+1}),
\end{equation*}
where
\begin{equation*}
\eta=max\{\|\mathcal{A}Y_{k+1}-b\|/\delta-1,0\}.
\end{equation*}
When $\delta=0$, it is the particular case of \eqref{fmodel} and can be expressed as:
\begin{equation*}
Y_{k+1}=Y_{k+1}+\mathcal{A}^{*}(b-\mathcal{A}Y_{k+1}).
\end{equation*}

Therefor, when the TNNR-ADMM is applied to solve \eqref{fmodel} and \eqref{thridmodel}, the generated subproblems all have closed-form solutions. Besides, some remarks are in order.
\begin{itemize}
\item $Z_{k+1}$ can be obtained via the following form
\begin{equation*}
Z_{k+1}=Z_{k}-\gamma\beta(X_{k+1}-Y_{k+1})
\end{equation*} where $0<\gamma<\frac{\sqrt{5}+1}{2}$ in \citep{glowinski1989augmented,glowinski2008lectures,chen2012matrix}. We make $\gamma=1$ to calculate $Z_{k+1}$ in our algorithms.
\item The convergence of the iterative scheme is well studied in \citep{boyd2011distributed}. Here, we omit the convergence analysis.
\end{itemize}

\section{\bf TNNR-APGL for \eqref{secondmodel}}\label{semodel}

In this section, we consider the model \eqref{secondmodel}, which has attracted a lot of attention in certain multi-task learning problems \citep{argyriou2008convex,abernethy2009new,obozinski2010joint,pong2010trace}. 
While TNNR-ADMM can be applied to solve this model, it is preferred for the noiseless problems. 
For the simple version of the model \eqref{secondmodel}, i.e. the one based on the common nuclear norm regularization,
%
many accelerated gradient techniques \citep{nesterov2007gradient,beck2009fast} based on \citep{nesterov1983method} are proposed. Among them, an accelerated proximal gradient line search (APGL) method proposed by Beck et al.\citep{beck2009fast} has been extended to solve TNNR based matrix completion model in \citep{hu2012fast}. In this paper,  we can extend APGL to solve the more general TNNR based the low-rank recovery problem \eqref{secondmodel}.

\subsection{\bf{TNNR-APGL with Noisy Data}}

For completeness, we give a short overview of the APGL method. The original model aims to solve the following problem:
\begin{equation*}
\min\{T(X)=F(X) + G(X) : X\in R^{m\times n}\},
\end{equation*}
where $G(X), \; T(X)$ meet these conditions:
\begin{itemize}
\item $G: \mathcal{R}^{m\times n}\rightarrow \mathcal{R} $ is a continuous convex function, possibly nondifferentiable function.
\item $F: \mathcal{R}^{m\times n}\rightarrow \mathcal{R} $ is a convex and differentiable function. In other words, it is continuously differentiable with Lipschitz continuous gradient $L(F)$\;($L(F) > 0$ is the Lipschitz constant of $\nabla F$).
\end{itemize}

By linearizing $F(X)$ at $Y$ and adding a proximal term, APGL constructs an approximation of $T(X)$. More specially, we have
\begin{equation*}
Q(X,Y)=F(Y)+\langle X-Y,g\rangle+\frac{1}{2\tau}\|X-Y\|_{F}^{2}+G(X),
\end{equation*}
where $\tau>0$ is a proximal parameter and $g=\nabla F(Y)$ is the gradient of $F(X)$ at Y.

\subsubsection{Algorithmic Framework}

For convenience, we present the model \eqref{secondmodel} again and define $F(X)$ and $G(X)$ as follows
\begin{equation*}
\begin{split}
&\min_{X} \;{\|X\|}_{*}-Tr(L_{r}^lX{R_{r}^l}^{\intercal} )+\frac{\mu}{2}\|\mathcal{A}X-b\|^{2}.\\
&F(X)=-Tr(L_{r}^lX{R_{r}^l}^{\intercal} )+\frac{\mu}{2}\|\mathcal{A}X-b\|^{2},\\
&G(X)=\|X\|_{*}.
\end{split}
\end{equation*}

Then, we can conclude that each iteration of the TNNR-APGL for solving  model  \eqref{secondmodel} requires solving the following subproblems.
\begin{equation}\label{firtite}
\left\{
\begin{split}
&X_{k+1}=\arg\min_{X\in\mathcal{ R}^{m\times n}} \{Q(X,Y_{k})\},\\
&\tau_{k+1}=\frac{1+\sqrt{1+4\tau_{k}^{2}}}{2},\\
&Y_{k+1}=X_{k+1}+\frac{\tau_{k}-1}{\tau_{k+1}(X_{k+1}-X_{k})}.
\end{split}
\right.
\end{equation}

During the above iterate scheme, we update $\tau_{k+1}$ and $Y_{k+1}$ via the approaches mentioned in \citep{beck2009fast,ji2009accelerated}. Then, based on \eqref{firtite}, we can easily drive the TNNR-APGL algorithmic framework as follows.
\vspace{2mm} \hrule \hrule
\vspace{2mm} {\bf Algorithm 4: TNNR-APGL for \eqref{secondmodel} }
\begin{itemize}
\item[1.] Initialization:~ set $X_{1}=Data$ (the matrix form of $b$)\;,$Y_{1}=X_{1},\pi_{1}=1$.
\item[2.] For $k=0,1,\cdots N,$ (Maximum number of iterations),
\begin{itemize}
\item[(\rmnum{1})] Update $X_{k+1}$ by
\begin{equation}\label{sssub1}
X_{k+1}=\arg\min_{X} \|X\|_{*}+\frac{1}{2\tau_{k}}\|X-(Y_{k}-\tau_{k}\nabla F(Y_{k}))\|_{F}^{2}.
\end{equation}
\item[(\rmnum{2})] Update $\tau_{k+1}$ by
\begin{equation*}
\tau_{k+1}=\frac{1+\sqrt{1+4\tau_{k}^{2}}}{2}
\end{equation*}
\item[(\rmnum{3})] Update $Y_{k+1}$ by
\begin{equation*}
Y_{k+1}=X_{k+1}+\frac{\tau_{k}-1}{\tau_{k+1}(X_{k+1}-X_{k})}.
\end{equation*}
\end{itemize}
\item[3.] End the iteration till \\
$\|X_{k+1}-X_{k}\|_{F}^{2}/\|Data\|_{F}^{2}\leq\varepsilon_{2}$.
\end{itemize}
\vspace{2mm} \hrule \hrule \vspace{5mm}

\subsubsection{The Analysis of Subproblems}

Obviously, the computation of each iteration of the TNNR-APGL approach for \eqref{secondmodel} is dominated by the subproblem \eqref{sssub1}. According to \text{Theorem 3.2}, we get
\begin{equation*}
X_{k+1}=\mathcal{D}_{\tau_{k}}(Y_{k}-\tau_{k}\nabla F(Y_{k})),
\end{equation*}
where
\begin{equation*}
\nabla F(Y_{k})=-{L_{r}^l}^{\intercal}{R_{r}^l}+\mu\mathcal{A}^{*}(\mathcal{A}Y_{k}-b).
\end{equation*}
Then, the closed-form solution of \eqref{sssub1} is given by
\begin{equation*}
X_{k+1}=\mathcal{D}_{\tau_{k}}(Y_{k}-\tau_{k}(\mu\mathcal{A}^{*}(\mathcal{A}Y_{k}-b))-{L_{r}^l}^{\intercal}{R_{r}^l}).
\end{equation*}


By now, we have applied TNNR-APGL to solve the problem \eqref{secondmodel} and obtain closed-form solutions. In addition, the convergence of APGL is well studied in \citep{beck2009fast} and it has a convergence rate of $O(\frac{1}{k^{2}})$. In our paper, we also omit the convergence analysis.
\section{\bf{ TNNR-ADMMAP for \eqref{fmodel} and \eqref{thridmodel}}}\label{moxing3}

While the TNNR-ADMM is usually very efficient for solving the TNNR based models \eqref{fmodel} and \eqref{thridmodel},  its  convergence could become slower with more constraints in \citep{he2012alternating}.  Inspired by \citep{hu2012fast}, the alternating direction method of multipliers with adaptive penalty (ADMMAP) is applied to reduce the constrained conditions, and adaptive penalty \citep{lin2011linearized} is used to speed up the convergence. The resulted algorithm is named as ``TNNR-ADMMAP", whose subproblems  can also get closed-form solutions.

\subsection{\bf{Algorithm Framework}}

Two kinds of constrains have been mentioned as before: $X=Y,\;\mathcal{A}Y=b$ and $X=Y,\;\|\mathcal{A}Y-b\|\leq\delta$. Our goal is to transform \eqref{fmodel} and \eqref{thridmodel} into the following form:
\begin{equation*}
\min _{x,y} F(x)+G(y),\;\text{s.t}\;P(x)+Q(y)=c,
\end{equation*}
where $P$ and $Q$ are linear mapping, $x, y$ and $c$ could  be either vectors or matrices, and $F$ and $G$ are convex functions.

In order to solve problems easily, $b$ was asked to be a vector formed by stacking the columns of matrices. On the contrary, if $\mathcal{A}$ is a linear mapping containing sampling process, we can put $\mathcal{A}Y=b$ into a matrix form sample set. Correspondingly, we should flexibly change the form between matrices and vectors in the calculation process. Here, we just provide the idea and process of TNNR-ADMMAP. Now, we match the relevant function to get the following results:
\begin{equation}\label{ladmmap-func}
F(X)=\|X\|_{*},\;\;G(Y)=-Tr(L_{r}^lYR_{r}^{\intercal}).
\end{equation}

\begin{equation*}
P(X) \;= \left(
\begin{array}{ccc}
X & 0\\
0 & 0
\end{array}
\right),\;\;
Q(Y) \;= \left(
\begin{array}{ccc}
-Y & 0\\
0 & \mathcal{A}Y
\end{array}
\right),
\end{equation*}

\begin{equation*}
C1\;= \left(
\begin{array}{ccc}
0 & 0\\
0 & Data
\end{array}
\right),\;\;
C2 \;= \left(
\begin{array}{ccc}
0 & 0\\
0 & \xi
\end{array}
\right).
\end{equation*}
where $P$ and $Q$ : $\mathcal{R}^{m\times n}\rightarrow \mathcal{R}^{2m\times 2n}$ and $C=C1+C2$. Denote $B_{\delta,2}=\{\zeta \in \mathcal{R}^{p}:\|\zeta\|\leq \delta\}$ and $\xi\in \mathcal{R}^{m\times n}$ that is the matrix form of $\zeta=\mathcal{A}X-b$. When $\delta=0$, it reflects the problem \eqref{fmodel}.

Then, the problems \eqref{fmodel} and \eqref{thridmodel} can be equivalently transformed to
\begin{equation}\label{admmap}
\min_{X} \;{\|X\|}_{*}-Tr(L_{r}^lY{R_{r}^l}^{\intercal}),\;\; s.t\;\;P(X)+Q(Y)=C.
\end{equation}
So the augmented Lagrangian function of \eqref{admmap} is:
\begin{equation}\label{lgadmmap}
\begin{split}
\mathcal{L}(X,Y,Z,\xi,\beta)&=\|X\|_{*}-\langle Z,P(X)+Q(Y)-C\rangle\\
&-Tr(L_{r}^lY{R_{r}^l}^{\intercal})+\frac{\beta}{2}\|P(X)+Q(Y)-C\|_{F}^{2},
\end{split}
\end{equation}
where
\begin{equation*}
Z \;= \left(
\begin{array}{ccc}
Z_{11} & Z_{12}\\
Z_{21} & Z_{22}
\end{array}
\right)\in\mathcal{ R}^{2m\times 2n}.
\end{equation*}

The Lagrangian form can be solved via linearized ADMM and a dynamic penalty parameter $\beta$ is preferred in \citep{lin2011linearized}. In particular, due to the special property of  $\mathcal{A}$ ($\mathcal{A}\mathcal{A}^{*}=\mathcal{I}$), here, we use ADMMAP in order to  handle the problem \eqref{lgadmmap} easily. Similarly, we use the following adaptive updating rule on $\beta$ \citep{lin2011linearized}:
\begin{equation}
\beta_{k+1}=\min(\beta_{max},\rho\beta_{k}),
\end{equation}
where $\beta_{max}$ is an upper bound of $\{\beta_{k}\}$. The value of $\rho$ is defined as
\begin{equation*} \rho\;= \left\{
\begin{split}
&\rho_{0}, if \;\frac{\beta_{k}\max\{\|X_{k+1}-X_{k}\|_{F},\|Y_{k+1}-Y_{k}\|_{F}\}}{\|C\|_{F}}<\varepsilon,\\
&1, otherwise.
\end{split}
\right.
\end{equation*}
where $\rho_{0}\geq1$ is a constant and $\varepsilon$ is a proximal parameter.

In summary, the iterative scheme of the TNNR-ADMMAP is as follows:
\vspace{2mm} \hrule \hrule
\vspace{2mm} {\bf Algorithm 5: TNNR-ADMMAP for \eqref{fmodel} and \eqref{thridmodel}  }
\begin{itemize}
\item[1.] Initialization:~ set $X_{1}=Data$ (the matrix form of b)\;,$Y_{1}=X_{1},Z_{1}=zeros(2m,2n)$, and input $\beta_{0},\varepsilon,\rho_{0}$.
\item[2.] For $k=0,1,\cdots N$ (Maximum number of iterations),
\begin{itemize}
\item[(\rmnum{1})] Update $X_{k+1}$ by
\begin{equation}\label{apsub1}
\begin{split}
X_{k+1}=&\arg\min_{X} \|X\|_{*}\\
&+\frac{\beta}{2}\|P(X)+Q(Y_{k})-C-\frac{1}{\beta}Z_{k}\|_{F}^{2}.
\end{split}
\end{equation}
\item[(\rmnum{2})] Update $Y_{k+1}$ by
\begin{equation}\label{apsub2}
\begin{split}
Y_{k+1}=&\arg\min_{Y}-Tr(L_{r}^lY{R_{r}^l}^{\intercal})\\
&+\frac{\beta}{2}\|P(X_{k+1})+Q(Y)-C-\frac{1}{\beta}Z_{k}\|_{F}^{2}.
\end{split}
\end{equation}
\item[(\rmnum{3})] Update $Z_{k+1}$ by
\begin{equation}\label{apsub3}
Z_{k+1}=Z_{k}-\beta(P(X_{k+1})+Q(Y_{k+1})-C).
\end{equation}
\item[(\rmnum{4})] The step is  calculated with $\delta>0$. Update $\xi$ by
\begin{equation}\label{apsub4}
\begin{split}
C2_{k+1}&=P(X_{k+1})+Q(Y_{k+1})-C1-\frac{1}{\beta}Z_{k+1}.\\
\xi_{k+1}&=\mathcal{P}_{B_{\delta,2}}(C2_{k+1})_{22}.
\end{split}
\end{equation}
\end{itemize}
\item[3.] End the iteration till \\
$\|X_{k+1}-X_{k}\|_{F}^{2}/\|Data\|_{F}^{2}\leq\varepsilon_{2}$.
\end{itemize}
\vspace{2mm} \hrule \hrule \vspace{5mm}

\subsection{\bf{The Analysis of Subproblems}}

Since the computation of each iteration of the  TNNR-ADMMAP method is dominated by solving the subproblems \eqref{apsub1} and \eqref{apsub2}, we now elaborate on the strategies for solving these subproblems.

First, we compute $X_{k+1}$.  Since the special form of $P$ and $Q$, we can give the following solution by ignoring the constant term:
\begin{equation*}
\begin{split}
X_{k+1}&=\arg\min_{X}\|X\|_{*}+\frac{\beta}{2}\|X-Y_{k}-\frac{1}{\beta}(Z_{k})_{11}\|_{F}^{2},\nonumber\\
X_{k+1}&=\mathcal{D}_{\frac{1}{\beta}}(Y_{k}+\frac{1}{\beta}(Z_{k})_{11}).
\end{split}
\end{equation*}

Second, we concentrate on computing $Y_{k+1}$. Obviously, $Y_{k+1}$  obeys the following rule:
\begin{equation*}
\begin{split}
0&\in\partial[-Tr(L_{r}^lY_{k+1}{R_{r}^l}^{\intercal})\\
&+\frac{\beta}{2}\|P(X_{k+1})+Q(Y_{k+1})-C-\frac{1}{\beta}Z_{k}\|_{F}^{2}].
\end{split}
\end{equation*}
It can be solved as:
\begin{equation}\label{wsolve}
Q^{*}Q(Y)=\frac{1}{\beta}{L_{r}^l}^{\intercal}{R_{r}^l}-Q^{*}
[P(X_{k+1})-C-\frac{1}{\beta}Z_{k}],
\end{equation}
where $Q^{*}$ is the adjoint operator of $Q$ which is mentioned in \eqref{adjoint}.

Let
\begin{equation*}
W \;= \left(
\begin{array}{ccc}
W_{11} & W_{12}\\
W_{21} & W_{22}
\end{array}
\right)\in \mathcal{R}^{2m\times 2n},
\end{equation*}
where $W_{ij}\in \mathcal{R}^{m\times n}$, according to \eqref{adjoint}, we have $\langle Q(Y),W\rangle=\langle Y,Q^{*}(W)\rangle$. More specially,
\begin{equation*}
\begin{split}
\langle Q(Y),W\rangle
&= Tr
\left(
\begin{array}{ccc}
-Y & 0\\
0 & \mathcal{A}Y
\end{array}
\right)
\left(
\begin{array}{ccc}
W_{11} & W_{12}\\
W_{21} & W_{22}
\end{array}
\right)^{\intercal}\\
&=Tr
\left(
\begin{array}{ccc}
-YW_{11}^{\intercal} & -YW_{21}^{\intercal}\\
\mathcal{A}YW_{12}^{\intercal} & \mathcal{A}YW_{22}^{\intercal}
\end{array}
\right)\\
&=Tr(-YW_{11}^{\intercal})+Tr(\mathcal{A}YW_{22}^{\intercal})\\
&=\langle Y,-W_{11}\rangle+\langle \mathcal{A}Y,W_{22}\rangle\\
&=\langle Y,-W_{11}\rangle+\langle Y,\mathcal{A}^{*}W_{22}\rangle\\
&=\langle Y,-W_{11}+\mathcal{A}^{*}W_{22}\rangle\\
&=\langle Y,Q^{*}(W)\rangle.
\end{split}
\end{equation*}
Thus, the adjoint operator $Q^{*}$ is denoted as
\begin{equation*}
Q^{*}(W)=-W_{11}+\mathcal{A}^{*}W_{22}.
\end{equation*}
The left side in \eqref{wsolve} can be shown as
\begin{equation}\label{lequ}
Q^{*}Q(Y)\;= Q^{*}\left(
\begin{array}{ccc}
-Y & 0\\
0 & \mathcal{A}Y
\end{array}
\right)
=Y+\mathcal{A}^{*}\mathcal{A}Y.
\end{equation}
Then, we apply the linear mapping $\mathcal{A}$ ($\mathcal{A}\mathcal{A}^{*}=\mathcal{I}$) on both sides of \eqref{lequ}, and we obtain
\begin{equation*}
\mathcal{A}(Q^{*}Q(Y))= \mathcal{A}Y+\mathcal{A}\mathcal{A}^{*}\mathcal{A}Y=2\mathcal{A}Y.
\end{equation*}
\begin{equation}\label{useA}
\mathcal{A}Y=\frac{1}{2}\mathcal{A}(Q^{*}Q(Y)).
\end{equation}
Combining \eqref{lequ} and \eqref{useA}, we get
\begin{equation*}
Y_{k+1}=Q^{*}Q(Y)-\mathcal{A}^{*}\mathcal{A}Y=Q^{*}Q(Y)- \frac{1}{2}\mathcal{A}^{*}\mathcal{A}(Q^{*}Q(Y)).
\end{equation*}
Similarly, according to the property of $Q^{*}$ in \eqref{lequ}, we can get the transformation for the right side in \eqref{wsolve}
\begin{equation}\label{requ}
\begin{split}
Q^{*}Q(Y)&=X_{k+1}-\frac{1}{\beta}(Z_{k})_{11}+\frac{1}{\beta}{L_{r}^l}^{\intercal}{R_{r}^l}\\
&+\mathcal{A}^{*}(Data+\xi+\frac{1}{\beta}(Z_{k})_{22})
\end{split}
\end{equation}
Based on the above from \eqref{lequ} and \eqref{requ}, we achieve
\begin{equation*}
\begin{split}
Y_{k+1}&=Q^{*}Q(Y)-\mathcal{A}^{*}\mathcal{A}Y\\
&=Q^{*}Q(Y)- \frac{1}{2}\mathcal{A}^{*}\mathcal{A}(Q^{*}Q(Y))\\
&=X_{k+1}-\frac{1}{2\beta}\mathcal{A}^{*}\mathcal{A}({L_{r}^l}^{\intercal}{R_{r}^l}-(Z_{k})_{11}+\beta X_{k+1})\\
&+\frac{1}{\beta}({L_{r}^l}^{\intercal}{R_{r}^l}-(Z_{k})_{11})+\frac{1}{2\beta}\mathcal{A}^{*}(\beta Data+\beta\xi+(Z_{k})_{22}).
\end{split}
\end{equation*}

Some remarks are in order.
\begin{itemize}
\item The compute of $\xi_{k+1}$ begins with  $\xi_{1}>0$. In other words, the problem matches \eqref{thridmodel} when $\xi_{1}>0$,.
\item The convergence of the iterative schemes of ADMMAP is well studied in \citep{he2000alternating,lin2011linearized}. In our paper, we omit the convergence analysis.
\end{itemize}

Overall, when TNNR-ADMM and TNNR-APGL are applied to solve \eqref{fmodel}-\eqref{thridmodel}, the generated subproblems all have closed-form solutions. As mentioned before, TNNR-ADMMAP is used to speed up the convergence of \eqref{fmodel} and \eqref{thridmodel} when there are too many constraints.  When one problem can be solved simultaneously with the three algorithms, TNNR-ADMMAP is in general more efficient, in the case of matrix completion \citep{hu2012fast}, and in our test problems.
\section{Experiments and Results}\label{experiment}

In this section, we present numerical results to validate the effectiveness of SVE and LRISD. In summary,  there are two parts certified by the following experiments. On one hand, we illustrate the effectiveness of SVE on both real visual and synthetic data sets. On the other hand, we also illustrate the effectiveness of LRISD which solves TNNR based low-rank matrix recovery problems on both synthetic and real visual data sets. 
%
%
Since the space is limited, we only discuss the model \eqref{fmodel} using LRISD-ADMM in our experiments. If necessary,  you can refer to \citep{hu2012fast} for extensive numerical results to understand that ADMMAP is much faster than APGL and ADMM without sacrificing the recovery accuracy, in cases of matrix completion. Similarly, for the low-rank matrix recovery, we have the same conclusion according to our experiments. Since the main aim of the paper is to present the effectiveness of SVE and LRISD, here, we omit the detailed explanation.

All experiments were performed under Windows 7 and Matlab v7.10 (R2010a), running on a HP laptop with an Intel Core 2 Duo CPU at 2.4
GHz and 2GB of memory.

\subsection{\bf{Experiments and Implementation Details} }\label{qianti}

We conduct the numerical experiments under the following four classes, where two representative linear mapping $\mathcal{A}$: matrix completion and partial DCT, are used. The first two cases are to illustrate the effectiveness of SVE. Here we compared our algorithm LRISD-ADMM with that proposed in \citep{hu2012fast}, which we name as ``TNNR-ADMM-TRY", on the matrix completion problem. The main difference between
  LRISD-ADMM and TNNR-ADMM-TRY is the way of determining the best $r$. The former is to estimate the best $r$ via SVE while the latter one is to try all the possible $r$ values and pick the one of the best performance.

The last two  are to show the better recovery quality of LRISD-ADMM compared with the solution of the common nuclear norm regularized low-rank recovery models, for example, \eqref{convexmodel}, whose corresponding algorithm is denoted as  LR-ADMM as above.  

\begin{itemize}
\item [(1)] Compare LRISD-ADMM with TNNR-ADMM-TRY on matrix completion problems. These experiments are conducted on real visual data sets.
\item [(2)] Compare the real rank $r$ with  $\tilde{r}$ which is estimated by SVE under different situations, where  $\mathcal{A}$ is a two-dimensional partial DCT operator ($\mathcal{A}\mathcal{A}^{*}= \mathcal{I}$). These experiments are conducted on synthetic data sets.
\item [(3)] Compare LRISD-ADMM with LR-ADMM on the generic low-rank situations, where $\mathcal{A}$ is also a two-dimensional partial DCT operator. These experiments are conducted on synthetic data sets under different problem settings.
\item [(4)] Compare LRISD-ADMM with LR-ADMM on the generic low-rank situations, where $\mathcal{A}$ is also a two-dimensional partial DCT operator. These experiments are conducted on real visual data sets.
\end{itemize}

In all synthetic data experiments, we generate the sample data as follows: $b=\mathcal{A}X^{*}+\omega$, where $\omega$ is Gaussian  white noise of mean zeros and standard  deviation $std$. The MATLAB script for generating $X^{*}$ as follows
\begin{equation}\label{makex}
X^{*}=randn(m,r)*randn(r,n).
\end{equation}
where $r$ is a prefixed integer. Moreover, we generate the index set $\Omega$ in \eqref{promodle} randomly in matrix completion experiments. And, the partial DCT is also generated randomly. 

In the implementation of all the experiments, we use the criterion：$\|X_{l+1}-X_{l}\|_{F}^{2}/\|Data\|_{F}^{2}\leq10^{-2}$ to terminate the iteration of \text{Step 2} (in Algorithm 1) in LR-ADMM and LRISD-ADMM. In addition, we terminate the iteration of b) in \text{Step 2} by the criterion: $\|X_{k+1}-X_{k}\|_{F}^{2}/\|Data\|_{F}^{2}\leq10^{-4}$. In our experiments, we set $\beta=0.001$ empirically, which works quit well for the tested problems. The other parameters in TNNR-ADMM are set to their default values (we use the Matlab code provided online by the author of \citep{hu2012fast}). Besides, we use the PSNR (Peak Signal to Noise Ratio) to evaluate the quality of an image. As color images have three channels (red, green, and blue), PSNR is defined as $10\times\log_{10}(\frac{255^{2}}{MSE})$. Where $MSE=SE/3T$, $SE=error_{red}^{2}+error_{green}^{2}+error_{blue}^{2}$ and $T$ is the total number of missing pixels. For grayscale images, PSNR has the similar definition.

\subsection{\bf{The Comparison between LRISD-ADMM and TNNR-ADMM-TRY on  Matrix Completion} }\label{completion experiment}

In this subsection, to evaluate the  effectiveness of SVE, we compare the proposed LRISD-ADMM with TNNR-ADMM-TRY as well as LD-ADMM on matrix completion problems. As the better recovery quality of TNNR-ADMM-TRY than  LR-ADMM on the matrix completion problem has been demonstrated in  \citep{hu2012fast}, we will show that
the final estimated $r$ of LRISD-ADMM via SVE is very close to the one of TNNR-ADMM-TRY. 

\begin{figure*}[!th]
\centering

\subfloat[Original image]{\includegraphics[width=0.185\textwidth]{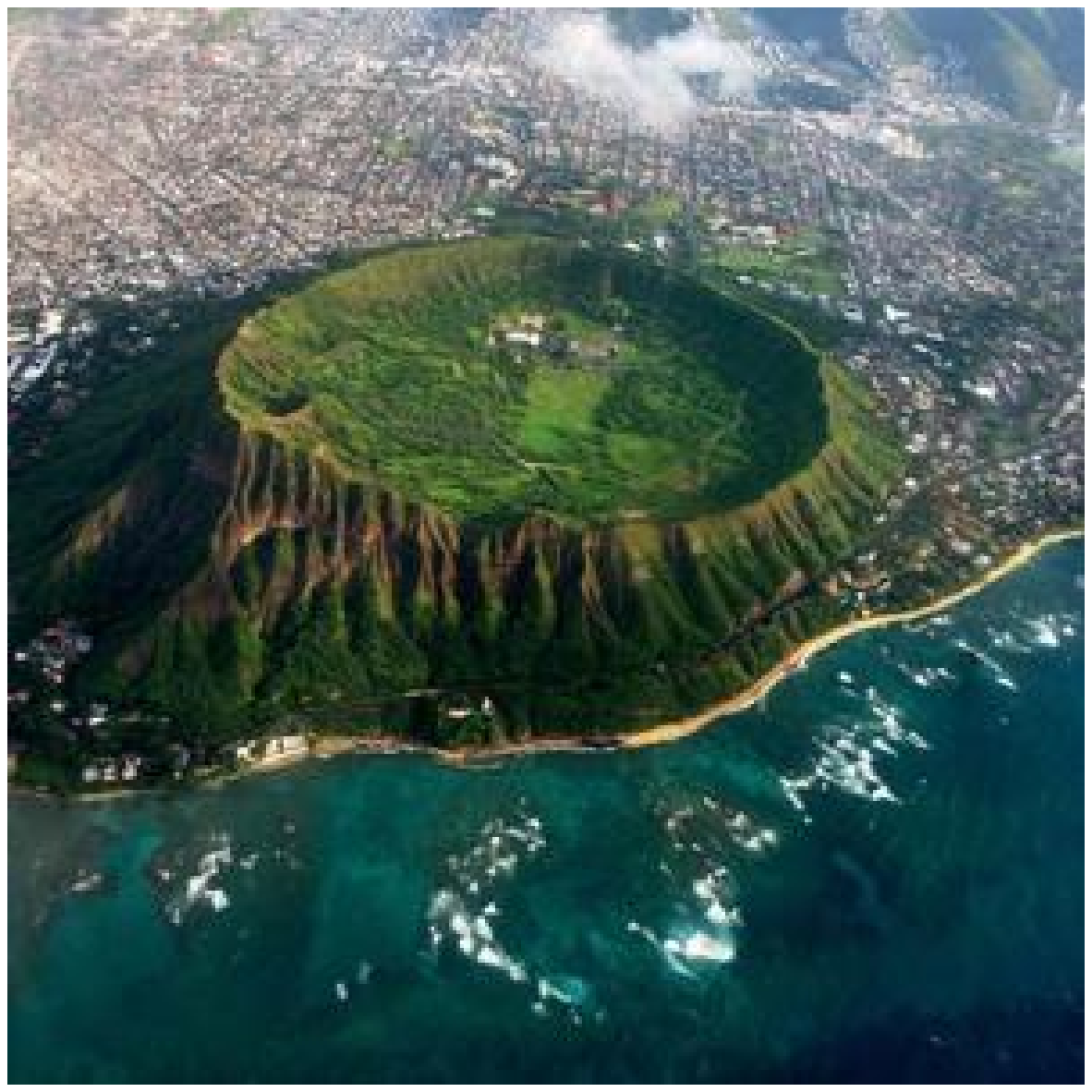}}
\hfil
\subfloat[Masked image]{\includegraphics[width=0.185\textwidth]{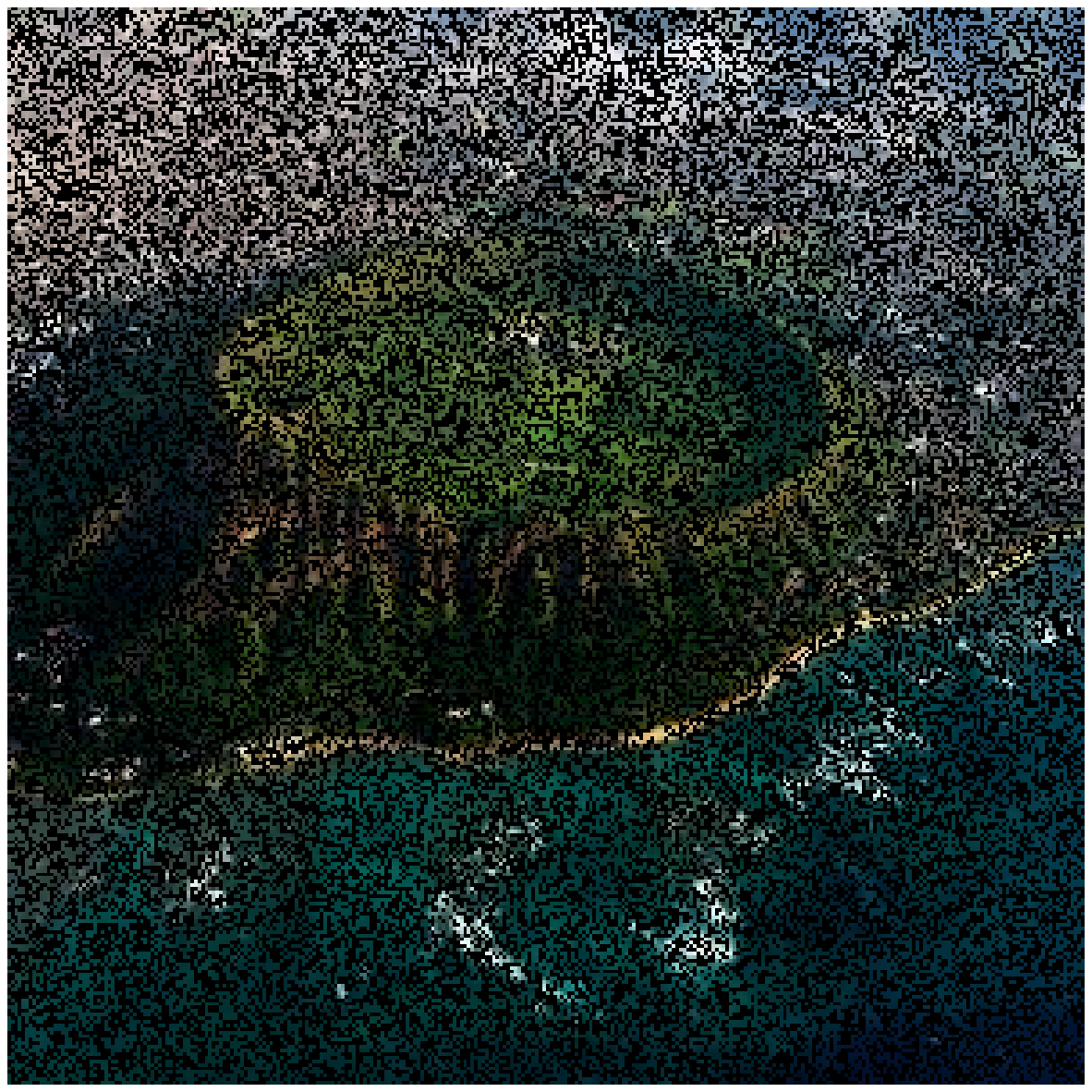}}
\hfil
\subfloat[LR-ADMM(r=0)]{\includegraphics[width=0.185\textwidth]{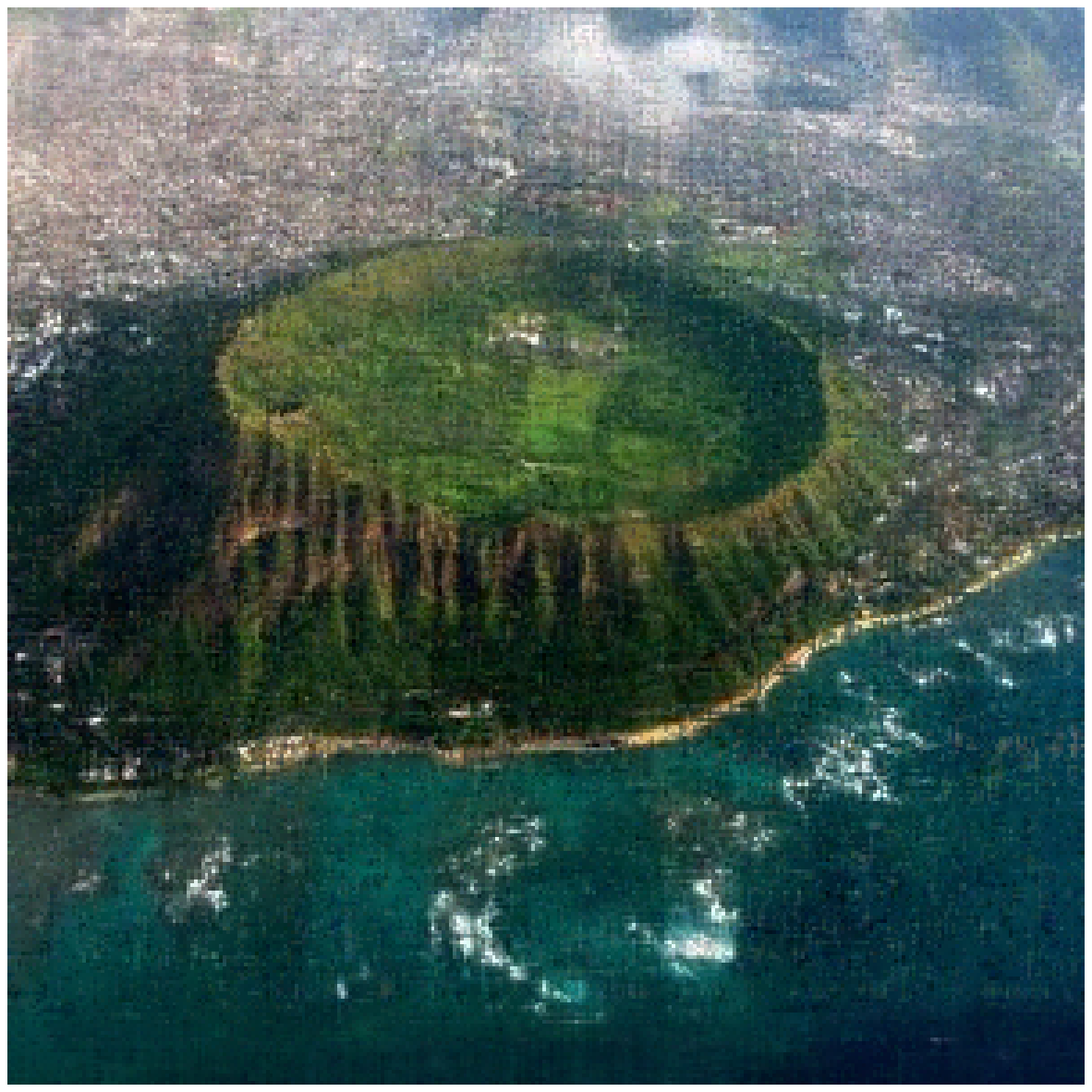}}
\hfil
\subfloat[TNNR-ADMM-TRY(r=6)LRISD-ADMM-ADJUST(r=6)]{\includegraphics[width=0.185\textwidth]{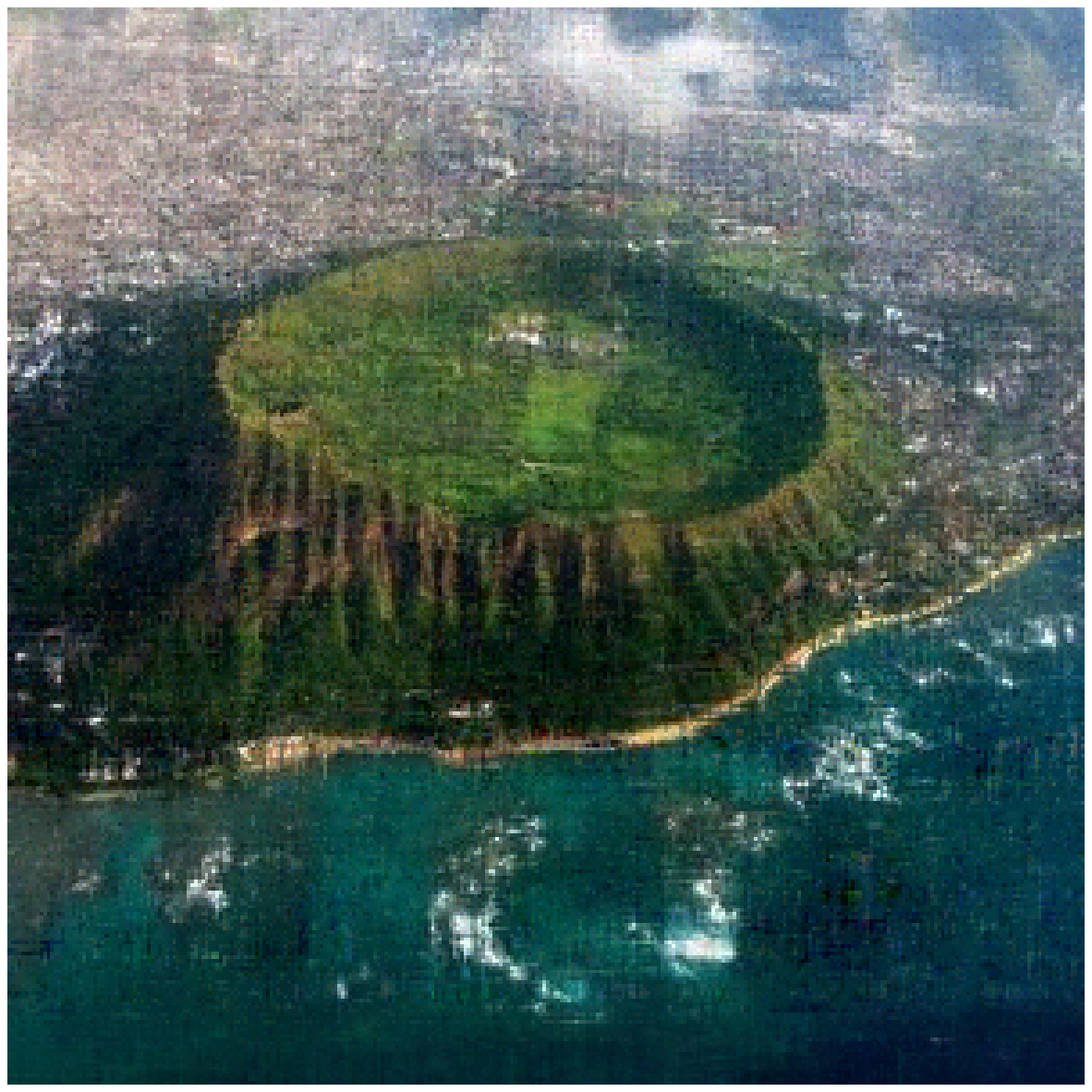}}
\hfil
\subfloat[LRISD-ADMM(r=7)]{\includegraphics[width=0.185\textwidth]{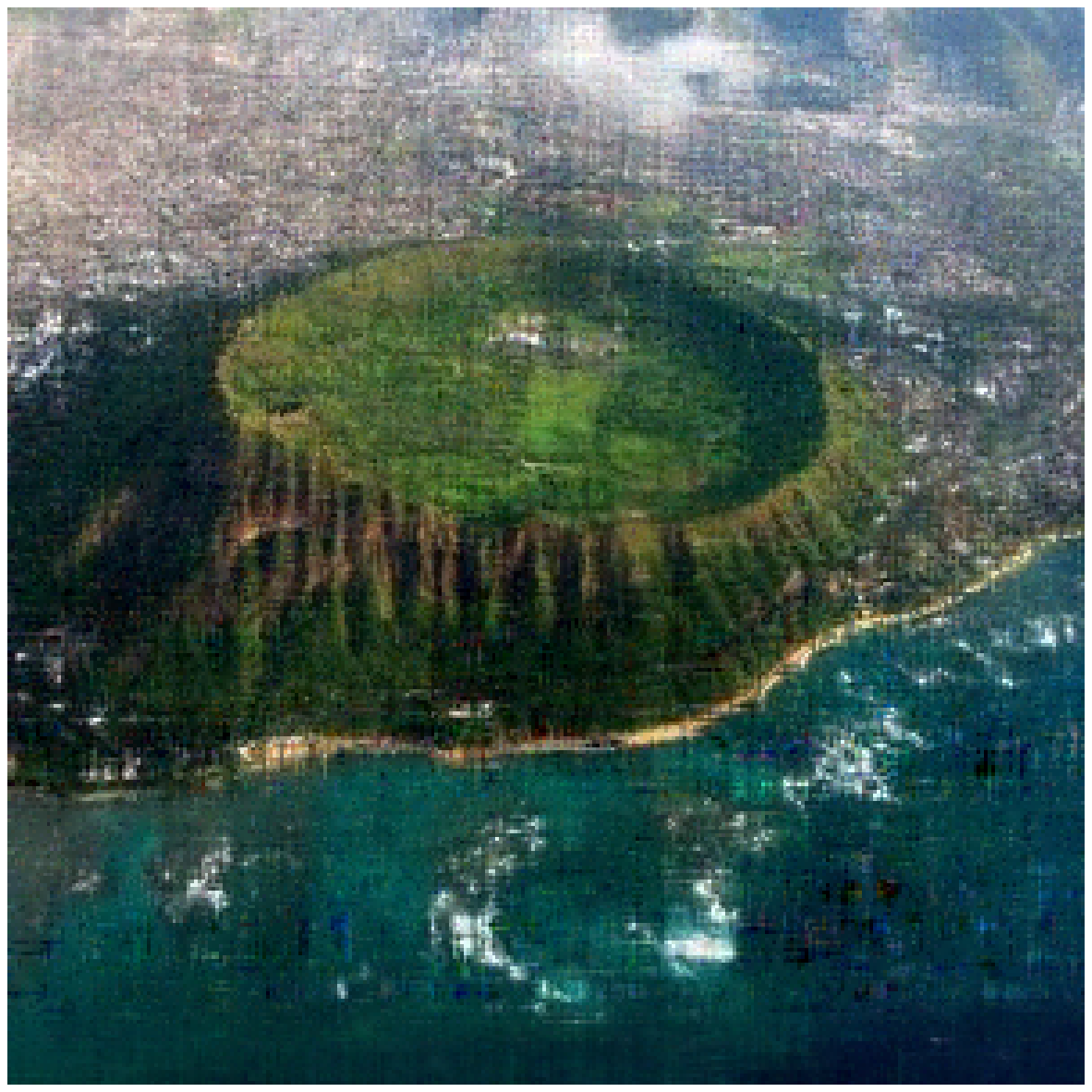}}\\
\subfloat[Original image]{\includegraphics[width=0.185\textwidth]{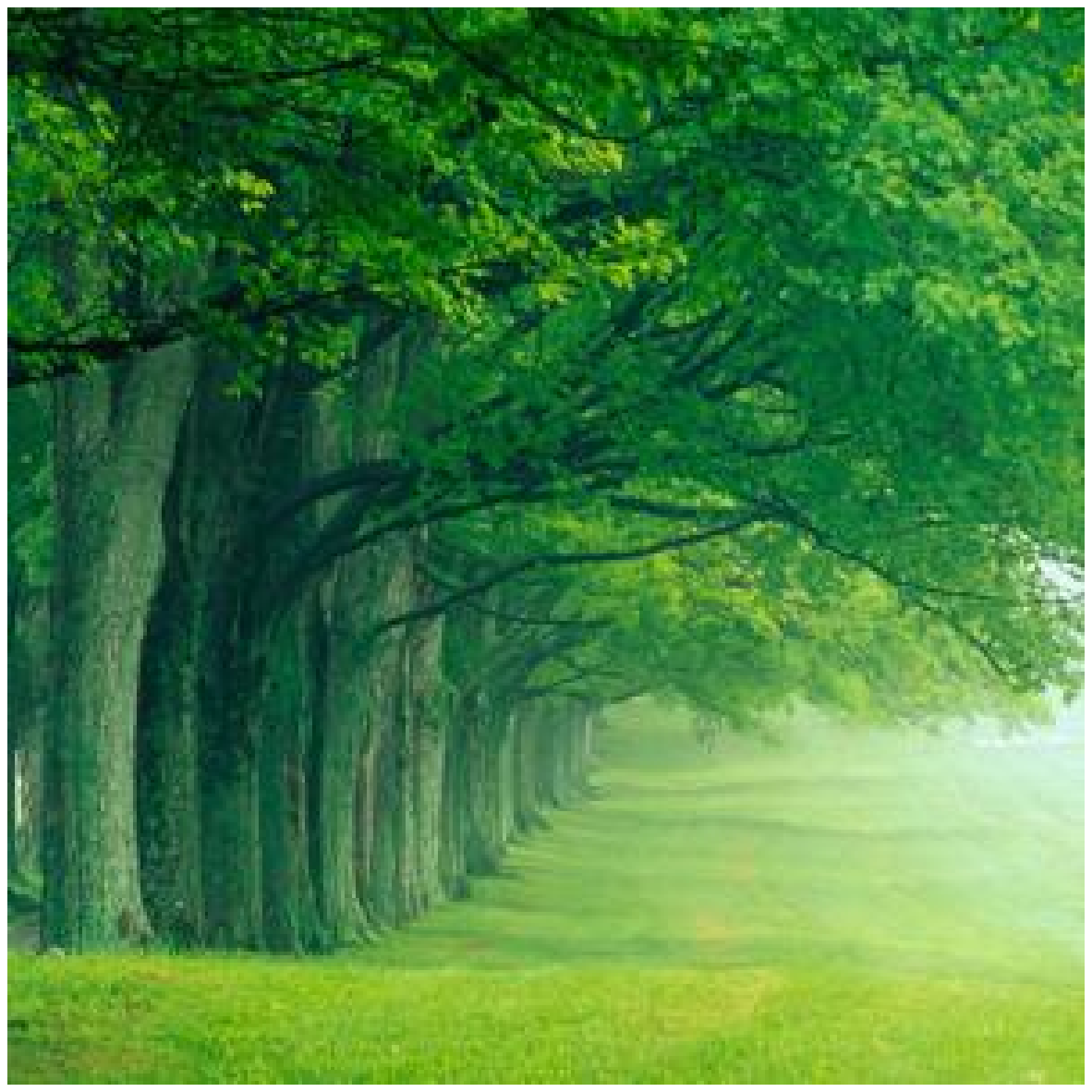}}
\hfil
\subfloat[Masked image]{\includegraphics[width=0.185\textwidth]{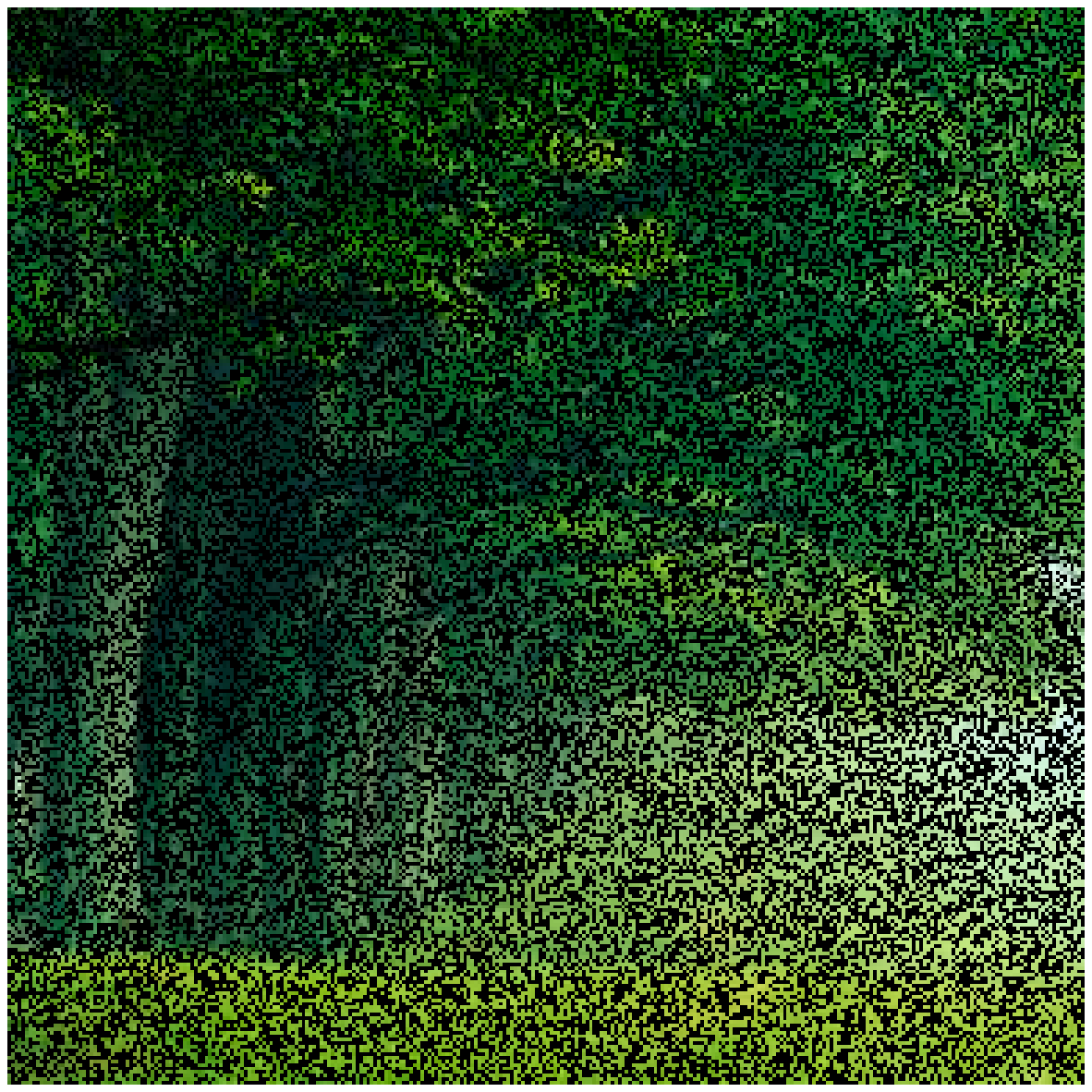}}
\hfil
\subfloat[LR-ADMM(r=0)]{\includegraphics[width=0.185\textwidth]{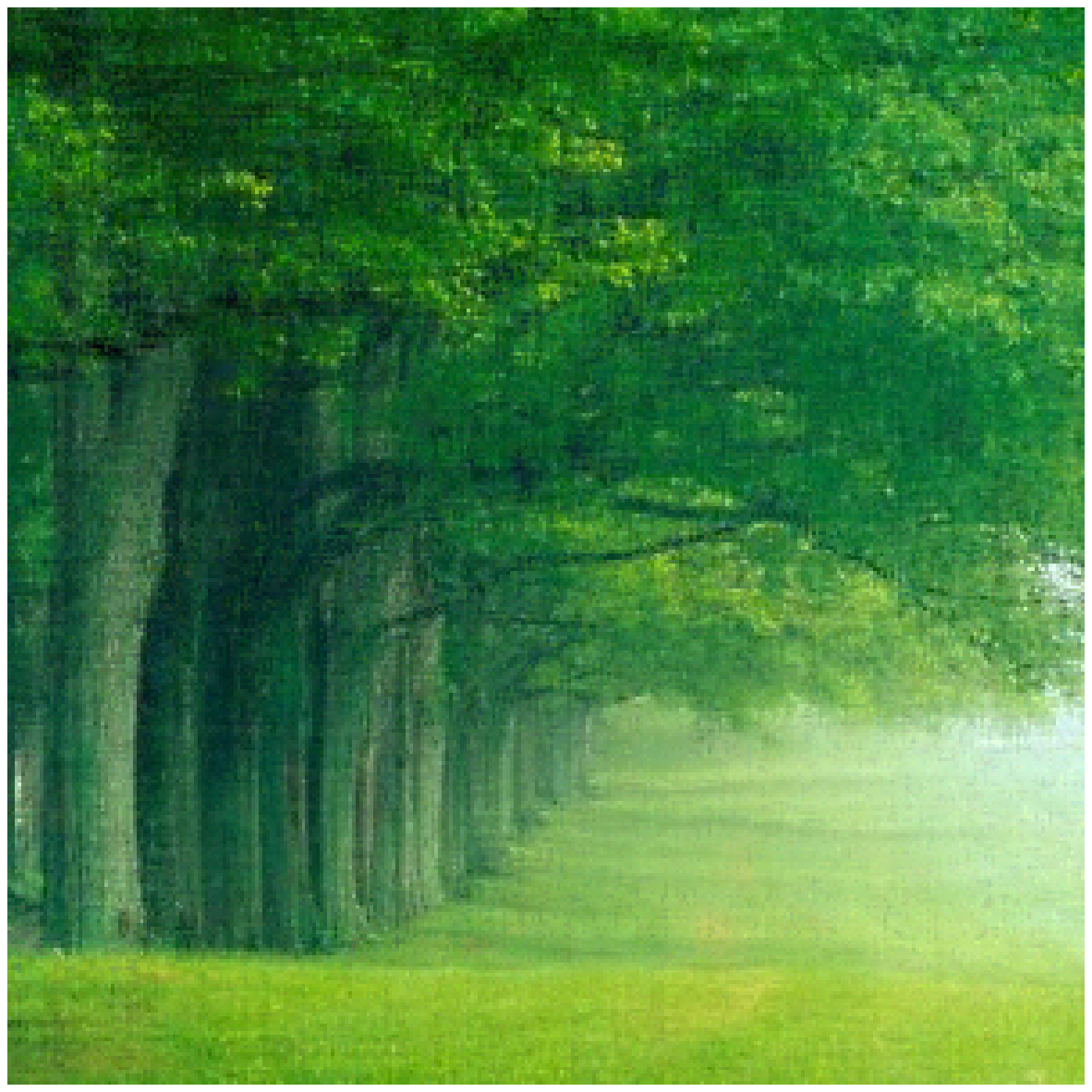}}
\hfil
\subfloat[TNNR-ADMM-TRY(r=6)LRISD-ADMM-ADJUST(r=6) ]{\includegraphics[width=0.185\textwidth]{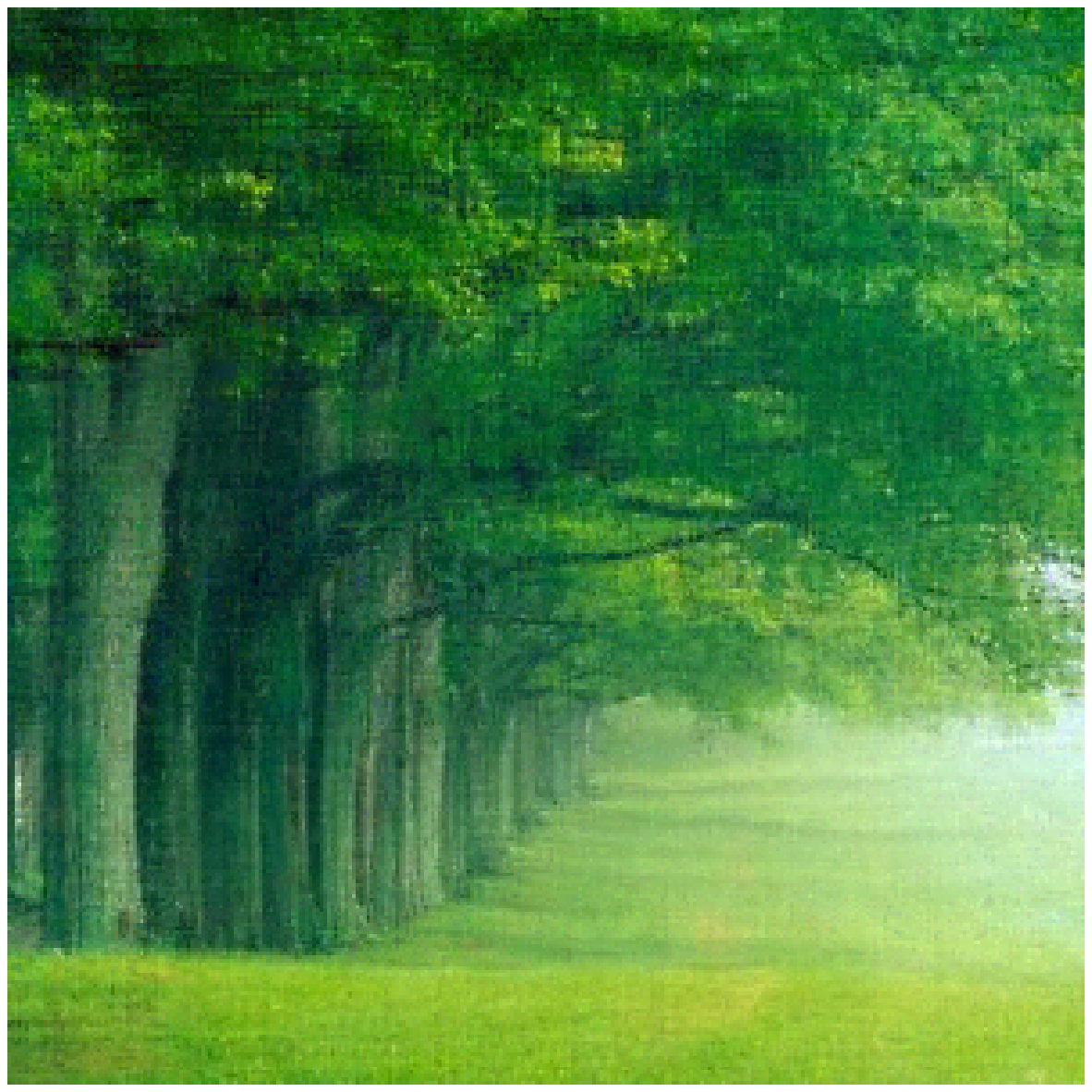}}
\hfil
\subfloat[LRISD-ADMM(r=5)]{\includegraphics[width=0.185\textwidth]{0atcam5-3.eps}}\\
\subfloat[Original image]{\includegraphics[width=0.185\textwidth]{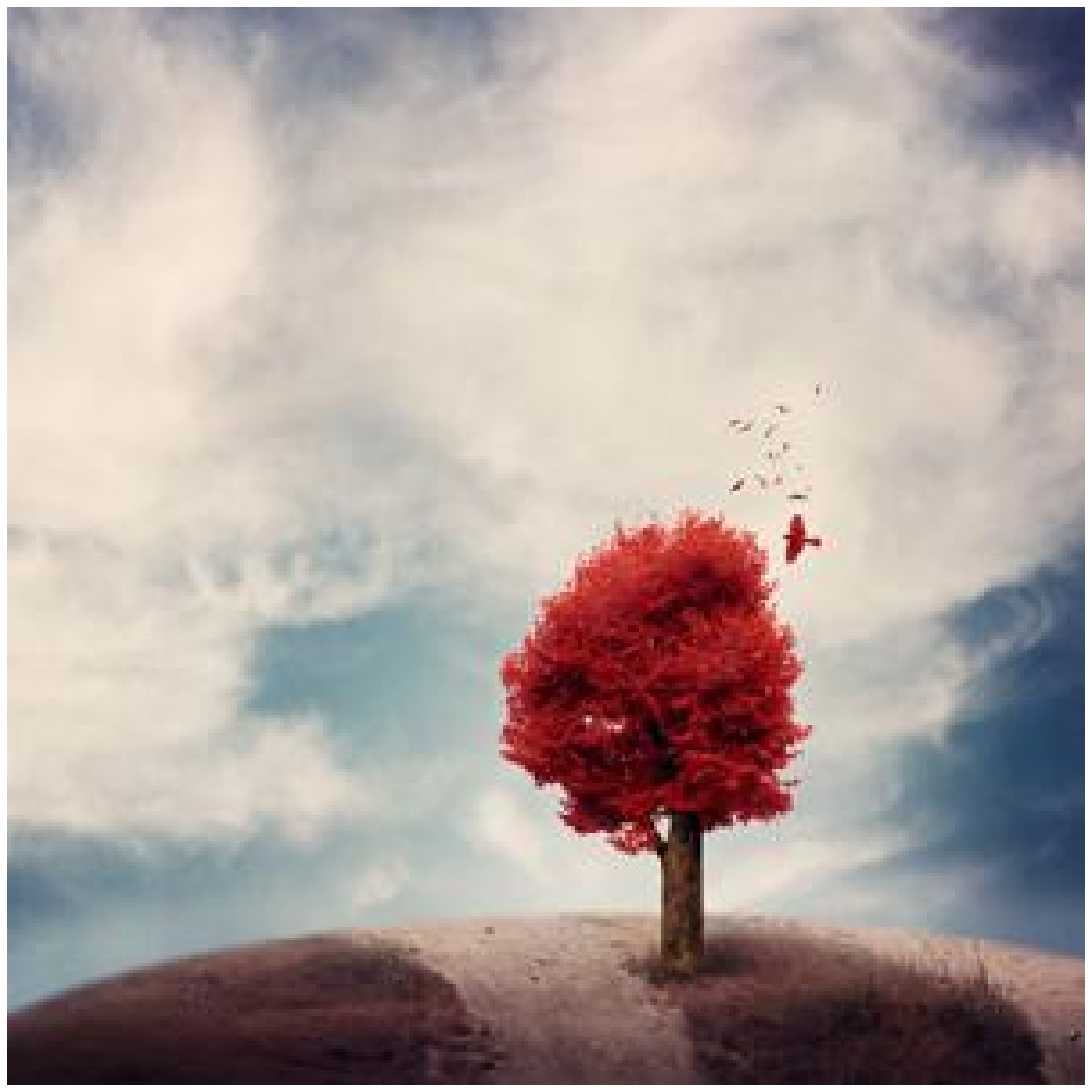}}
\hfil
\subfloat[Masked image]{\includegraphics[width=0.185\textwidth]{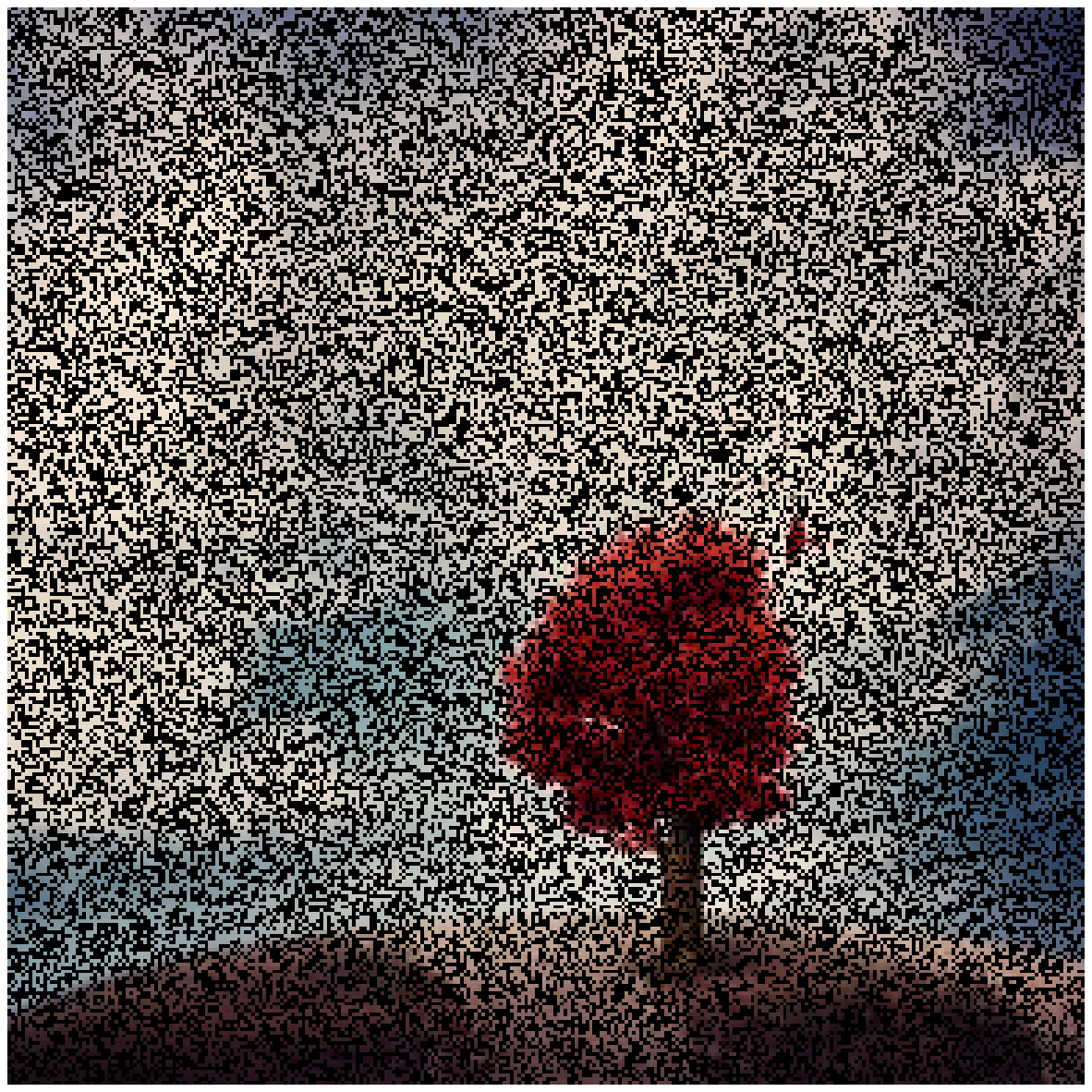}}
\hfil
\subfloat[LR-ADMM(r=0)]{\includegraphics[width=0.185\textwidth]{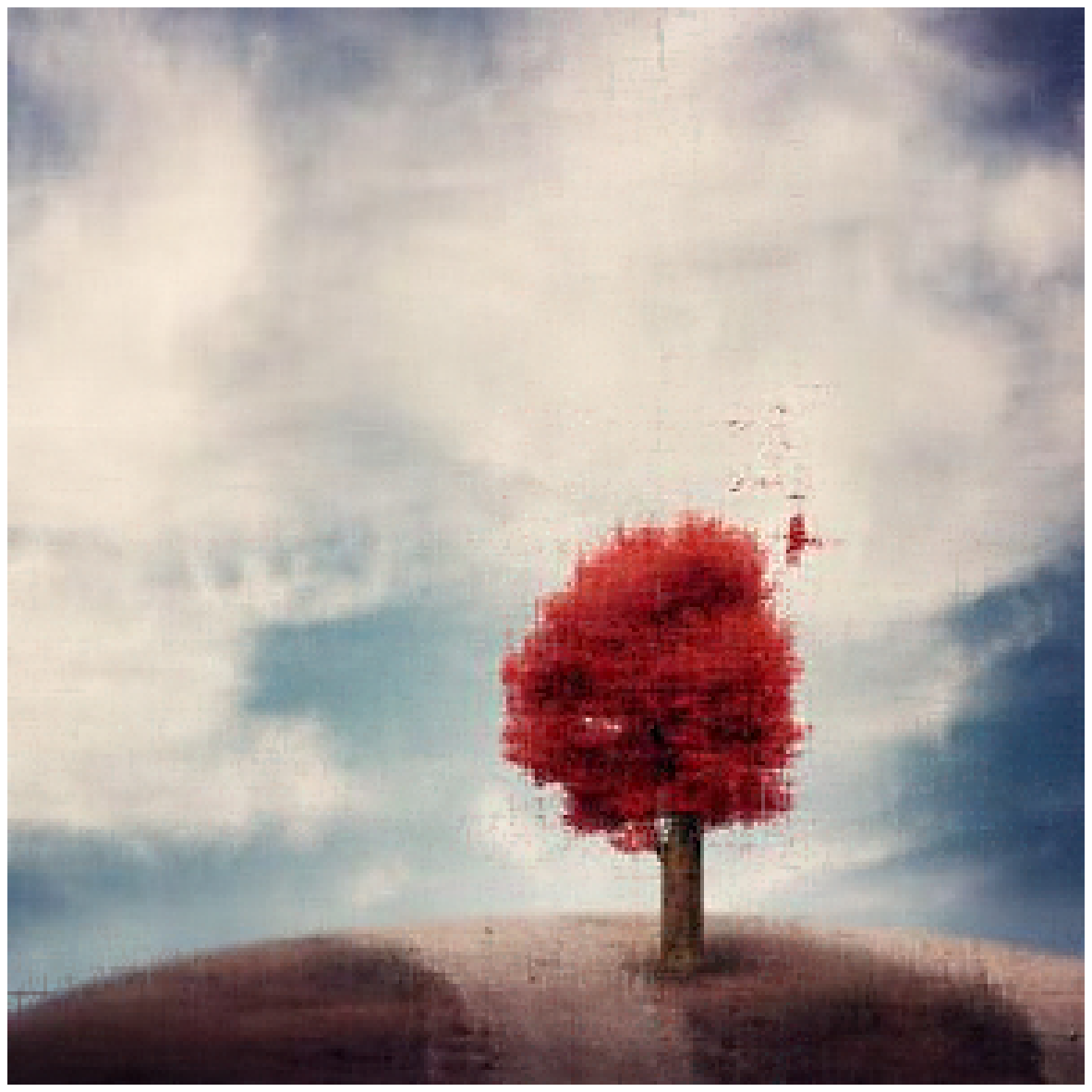}}
\hfil
\subfloat[TNNR-ADMM-TRY(r=15)LRISD-ADMM-ADJUST(r=15)]{\includegraphics[width=0.185\textwidth]{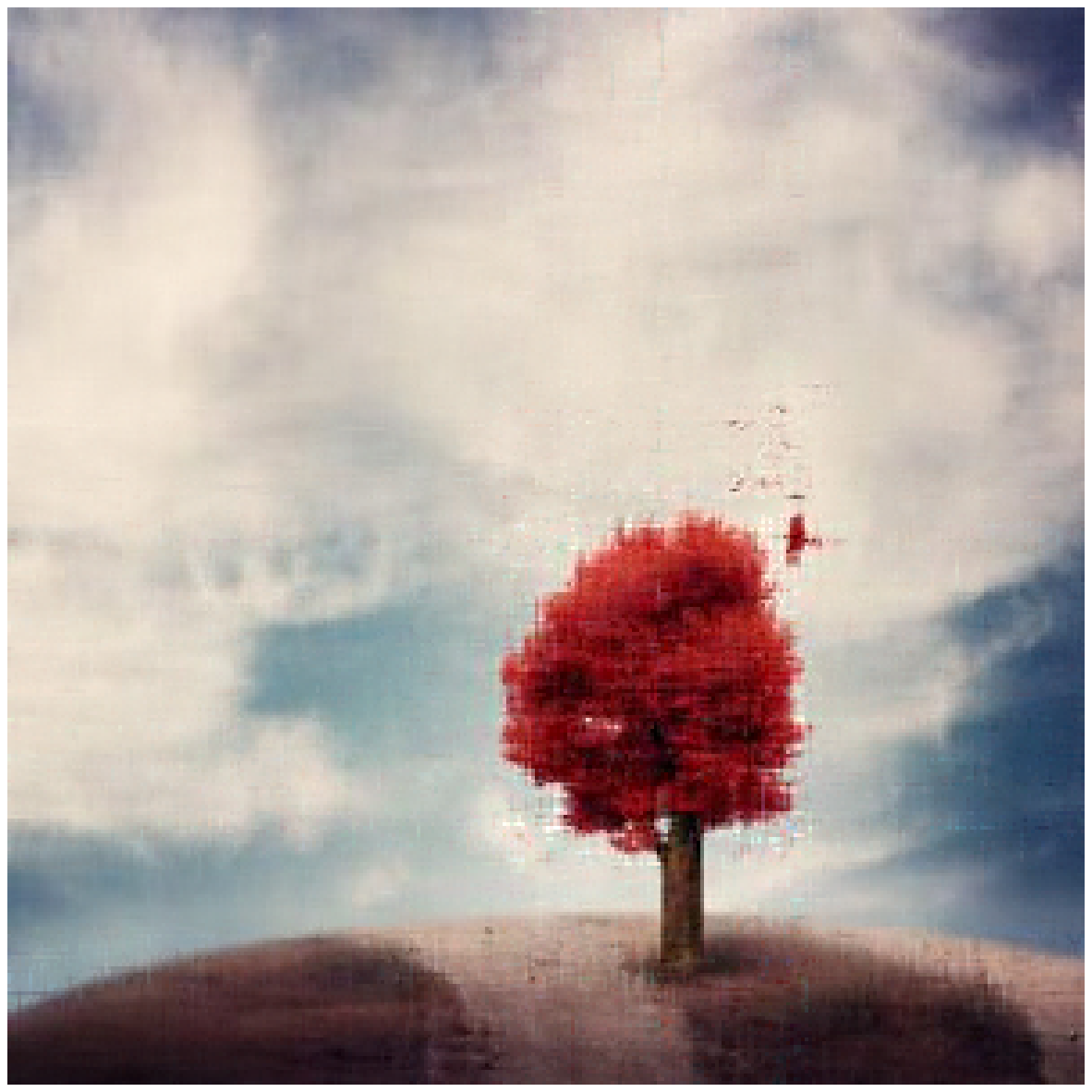}}
\hfil
\subfloat[LRISD-ADMM(r=11)]{\includegraphics[width=0.185\textwidth]{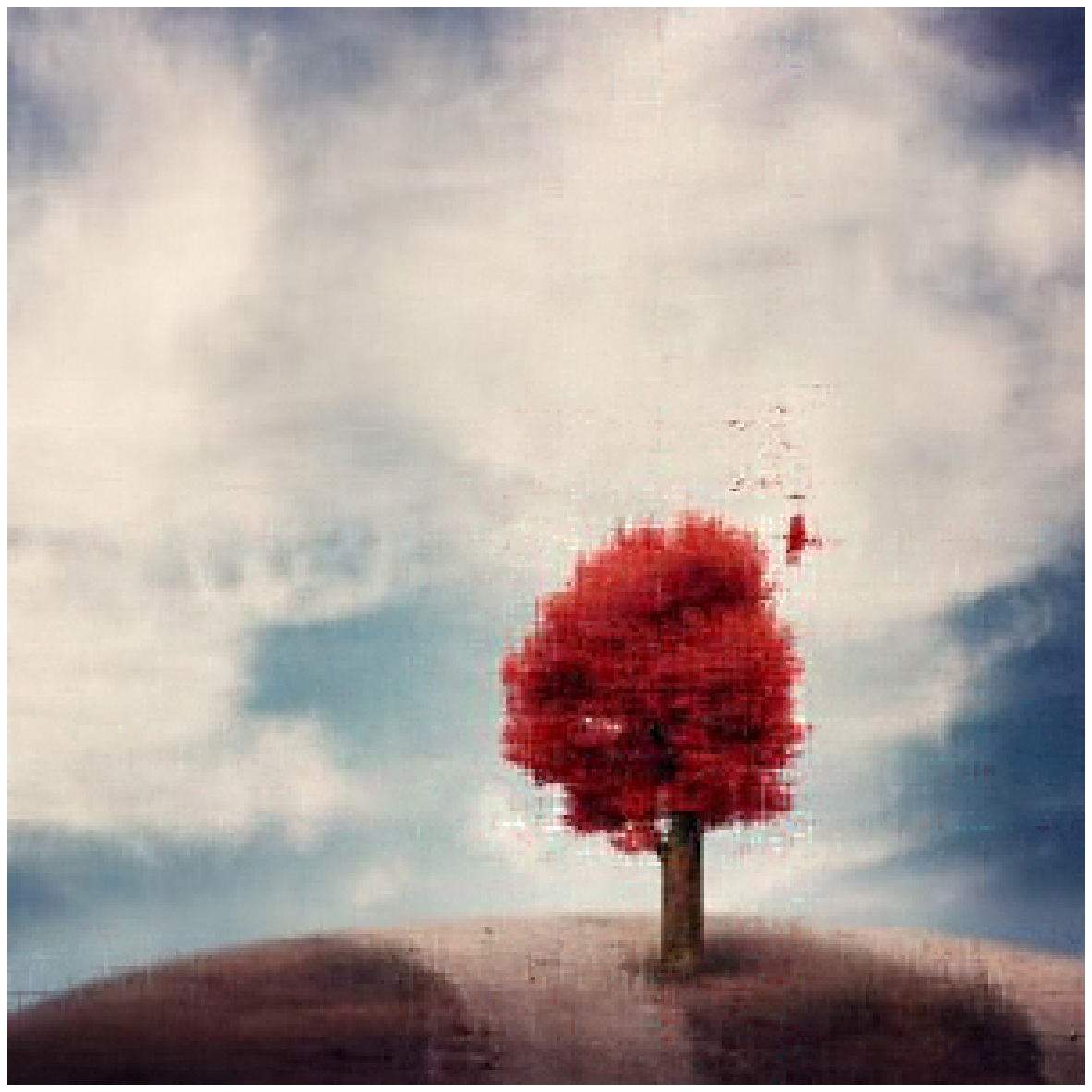}}
\caption{\small Comparisons results of LR-ADMM, TNNR-ADMM-TRY, LRISD-ADMM and LRISD-ADMM-ADJUST, we use three images here. The first column are original images. The second column are masked images. The masked images is obtained by covering $50\%$ pixels of the original image in our test. The third column depicts images recovered by LR-ADMM.  The fourth column depicts images recovered by TNNR-ADMM-TRY and LRISD-ADMM-ADJUST. The fifth column depicts images recovered by LRISD-ADMM where we just use the estimated $\tilde{r}$ directly. Noticing the fourth column, we get the same image by applying two different methods TNNR-ADMM-TRY and LRISD-ADMM-ADJUST. The reason is the values of $r$ calculated in the two methods are the same. But the procedure they find $r$ is different. In TNNR-ADMM-TRY, they search the best $r$ via testing all possible values (1-20). In LRISD-ADMM-ADJUST, we use the estimated rank $\tilde{r}$ as a reference to search around for the best $r$.}
\label{fig:2}
\end{figure*}

We test three real clear images and present the input images, the masked images and the results calculated via three different algorithms: LR-ADMM, TNNR-ADMM-TRY and LRISD-ADMM. The recovery images are showed in Fig \ref{fig:2} and the numerical value comparison about time and  PSNR are shown in Table $1$. 
We can see that compared to LR-ADMM,  both TNNR-ADMM-TRY and LRISD-ADMM achieve better recovery quality as expected.
While the TNNR-ADMM-TRY achieves the best recovery quality as expected due to its trying every possible $r$ value, its running time is extremely longer than LR-ADMM. Our proposed LRISD-ADMM can achieve almost the same recovery quality as TNNR-ADMM-TRY, but with a significant reductions of computation cost.
%
%
In fact, if the best precision is  expected, 
we use the estimated $\tilde{r}$ by LRISD-ADMM as a reference to search the best $r$ around it, with the reasonably extra cost of computation. This will  we can also find the best $r$, which is the same as TNNR-ADMM-TRY. Here, for convenience in notation, we name the process LRISD-ADMM-ADJUST in Table 1 and Fig \ref{fig:2}.   
\begin{table*}
\caption{\small We compare the PSNR and time between LR-ADMM, TNNR-ADMM-TRY, LRISD-ADMM and LRISD-ADMM-ADJUST. In TNNR-ADMM-TRY, they search the best $r$ via testing all possible values of $r$. We use the estimated rank $\tilde{r}$ as the best $r$ in LRISD-ADMM. In LRISD-ADMM-ADJUST, we use the estimated rank $\tilde{r}$ as a reference to search the best $r$.}
\begin{center}
\begin{tabular}{ccccccccccccccccccccccccc}
\hline
\multirow{2}{*}{image}&&\multicolumn{5}{c}{LR-ADMM}&&\multicolumn{5}{c}{TNNR-ADMM-TRY}& & \multicolumn{5}{c}{LRISD-ADMM} & & \multicolumn{5}{c}{LRISD-ADMM-ADJUST} \\
  \cline{3-7}\cline{9-13}\cline{15-19} \cline{21-25}
   & & $r$ & & time & & PSNR & & $r$ & & time& & PSNR & & $r=\tilde{r}$& & time& & PSNR & & $r$& & time & & PSNR \\
\hline
 1 & & 0 & &73.8s& & 21.498   && 6& &5030s && 21.645& & 7 && 221s && 21.618 & & 6 && 683s &&  21.645 \\
 2 & &0 & &98.3s & & 24.319 &&6 && 5916s && 24.366 && 5 & &150s && 24.357 && 6 & &799s && 24.366\\
 3 & &0 & &106.3s & &29.740 &&15 && 3408s& & 30.446 & &11&&148s && 30.342 & &15&&1433s && 30.446\\
\hline
\end{tabular}
\end{center}
\label{table1}
\end{table*}

\subsection{\bf{The Effectiveness Analysis of SVE in Two-dimensional Partial DCT: Synthetic Data}}

In subsection \ref{completion experiment}, we showed that the estimated  $\tilde{r}$ by LRISD-ADMM is very close to the best $r$ by TNNR-ADMM-TRY, on matrix completion problems. Here we further confirm the effectiveness of the proposed SVE of LRISD-ADMM,  by conducting experiments for the generic low-rank  operator $\mathcal{A}$ on synthetic data, where $\mathcal{A}$ is a two-dimensional partial DCT (discrete cosine transform) operator. We compare the best $r$ (true rank) with the estimated $\tilde{r}$  under different settings.

\begin{figure*}[htbp]
\centering
\subfloat[]{\includegraphics[height=5.2cm,width=0.33\textwidth]{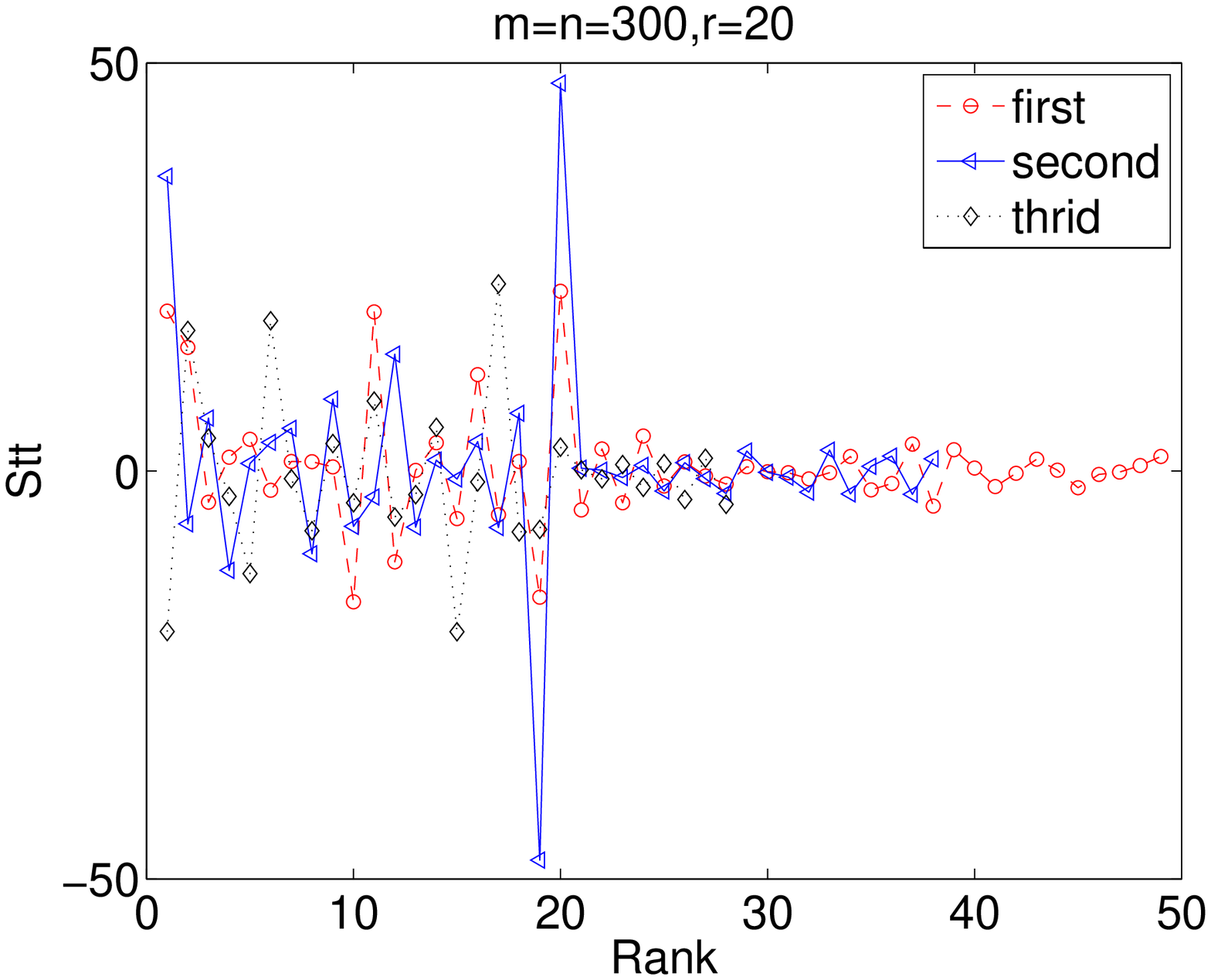}}
\hfil
\subfloat[]{\includegraphics[height=5.2cm,width=0.33\textwidth]{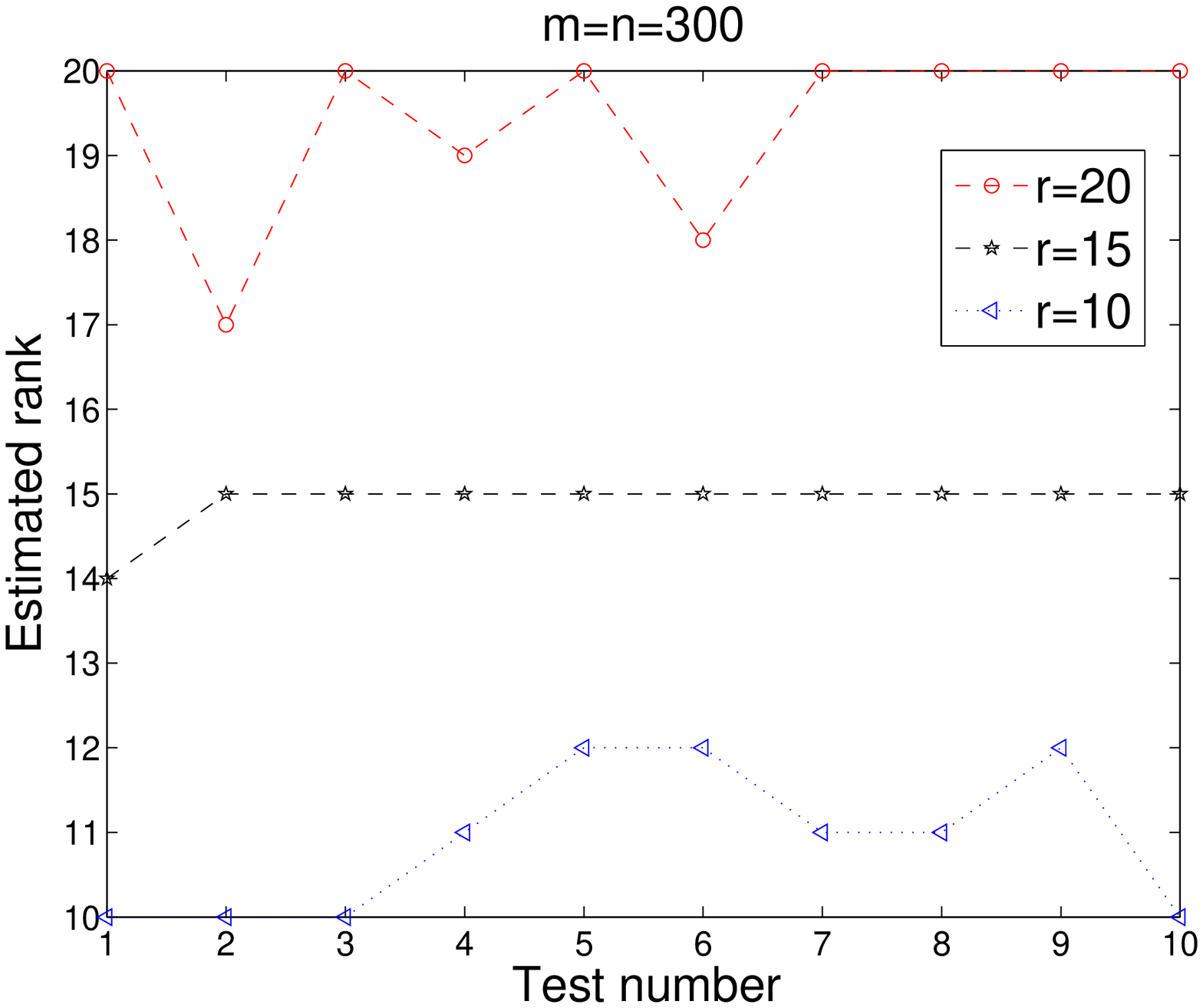}}
\hfil
\subfloat[]{\includegraphics[height=5.2cm,width=0.33\textwidth]{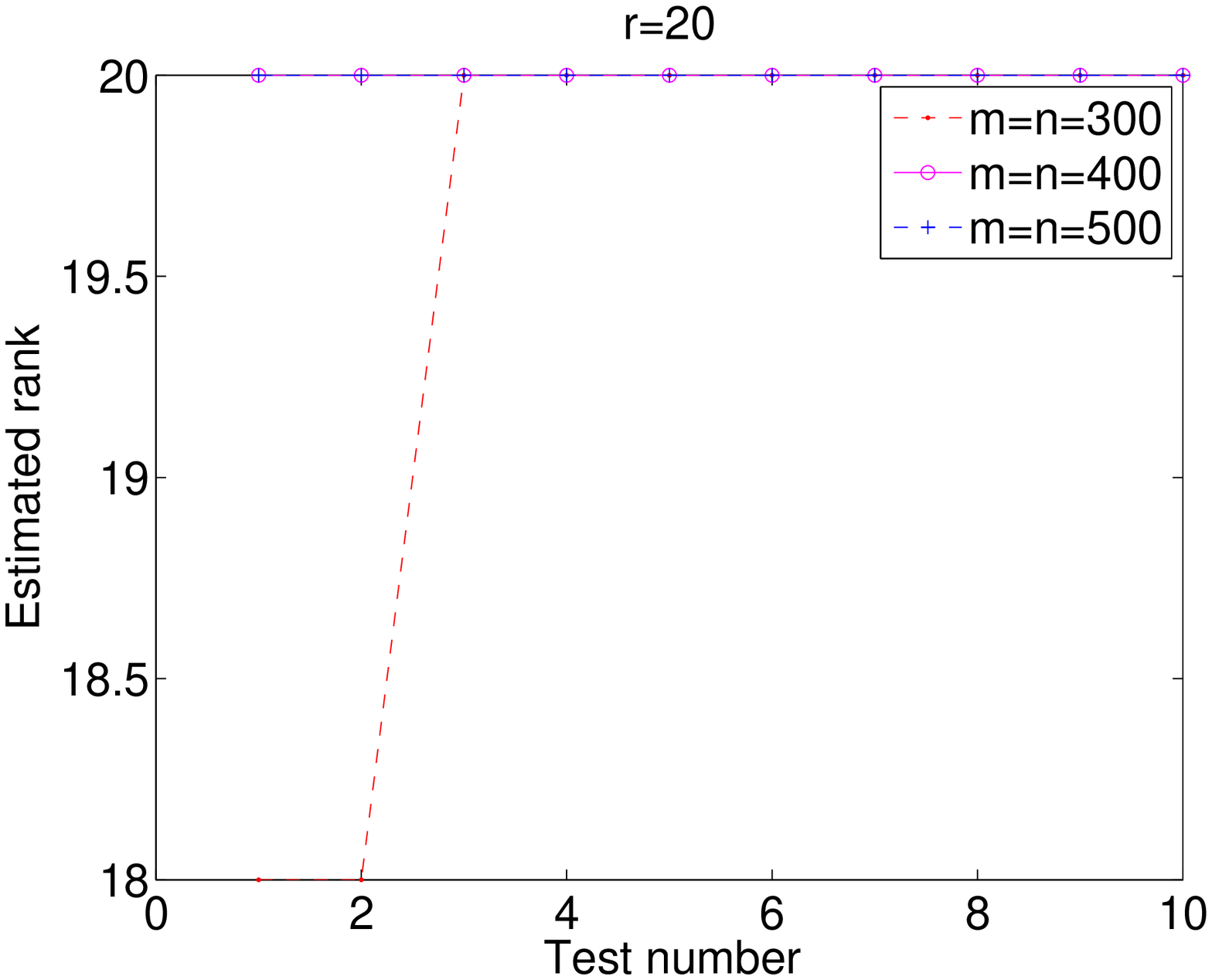}}
\caption{We give the relationship between $Stt$ and $r$ (tested three times) in $(a)$. Comparisons between the true rank $r$ and the estimated rank $\tilde{r}$ under different ranks  are shown in chart $(b)$ and different matrix sizes in chart $(c)$.}
\label{fig:3}
\end{figure*}

In the results below, $r$, $\tilde{r}$, and $sr$ denote,  the rank of the matrix $X^{*}$, estimated rank, and sample ratio taken, respectively. We set the noise level $std=0.9$,  the sample ratios $sr=0.5$ and choose $\kappa=10$ in all of the tests. The reason of setting $std=0.9$ is that we want to well illustrate the robustness of SVE to noise. Next, we compare $r$ with $\tilde{r}$ under different settings. For each scenario, we generated the model by 3 times and report the results.
\begin{itemize}
\item We fix the matrix size to be $m=n=300, r=20$ and run LRISD-ADMM to indicate the relationship between $Stt$ and $r$. The results are showed in Fig \ref{fig:3}$(a)$
\item We fix the matrix size to be $m=n=300$ and run LRISD-ADMM under different $r$. The results are showed in  Fig \ref{fig:3}$(b)$
\item We fix $r=20$ and run LRISD-ADMM under different matrix sizes. The results are showed in  Fig \ref{fig:3}$(c)$
\end{itemize}


As shown in Fig \ref{fig:3}, the proposed SVE performs the rationality and effectiveness to estimate $\tilde{r}$ in \text{Step 1} in Algorithm 1. Even if there is much noise, this method is still valid, namely, $\tilde{r}$ is (approximately) equivalent to the real rank.  That is to say, the proposed \text{SVE} is pretty robust to the corruption of noise on sample data. In practice, we can achieve the ideal results in other different settings. To save space, we only illustrate the effectiveness of SVE using the above-mentioned situations.
\subsection{\bf{The Comparison between LRISD-ADMM and LR-ADMM: Synthetic Data}}

In this subsection, we compare the proposed LRISD-ADMM with LR-ADMM on partial DCT data in general low rank matrix recovery cases.
We will illustrate some numerical results to show the advantages of the proposed LRISD-ADMM in terms of better recovery quality. 

\begin{figure*}[!t]
\centering
\subfloat[]{\includegraphics[height=4.5cm,width=0.33\textwidth]{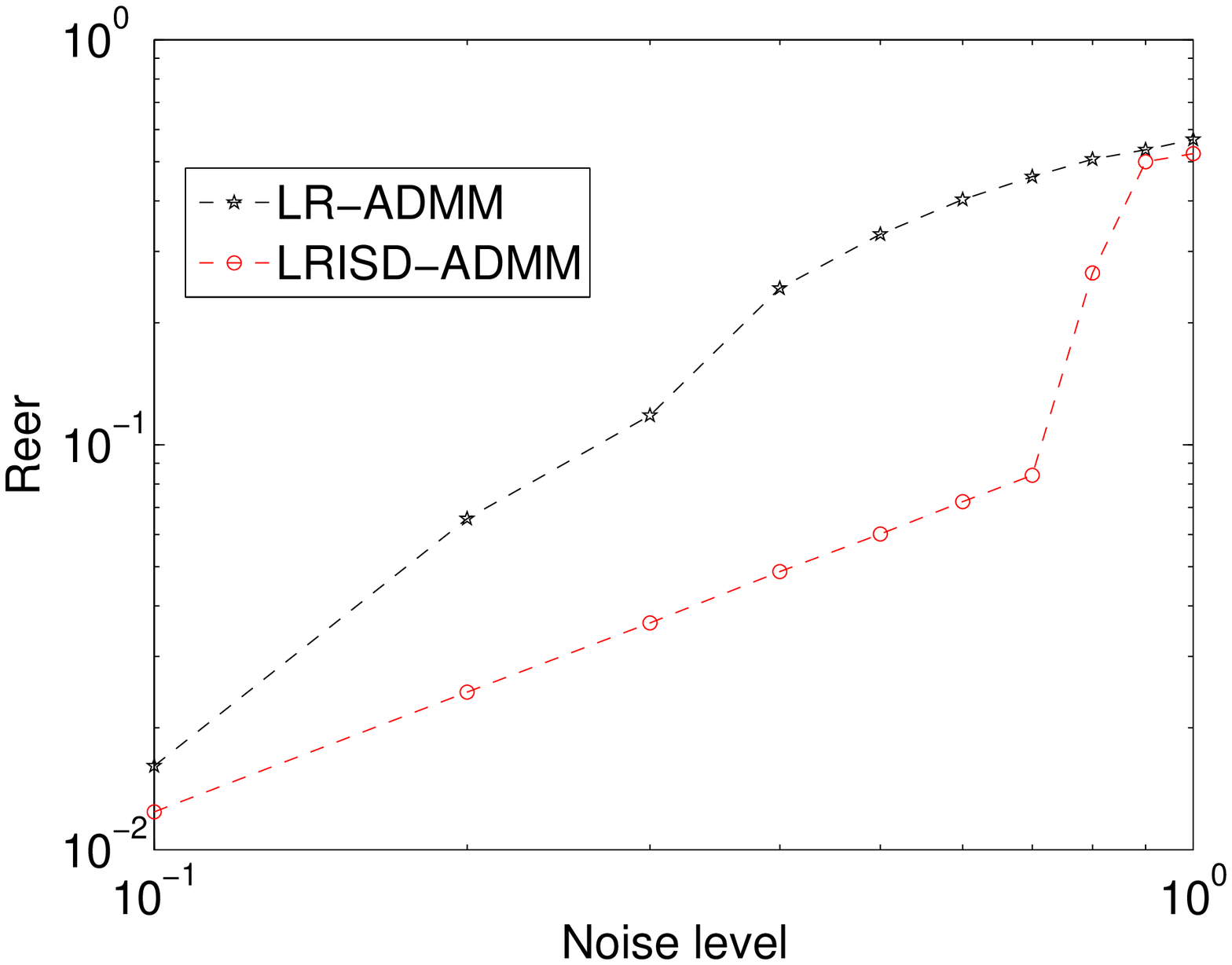}}
\hfil
\subfloat[]{\includegraphics[height=4.5cm,width=0.33\textwidth]{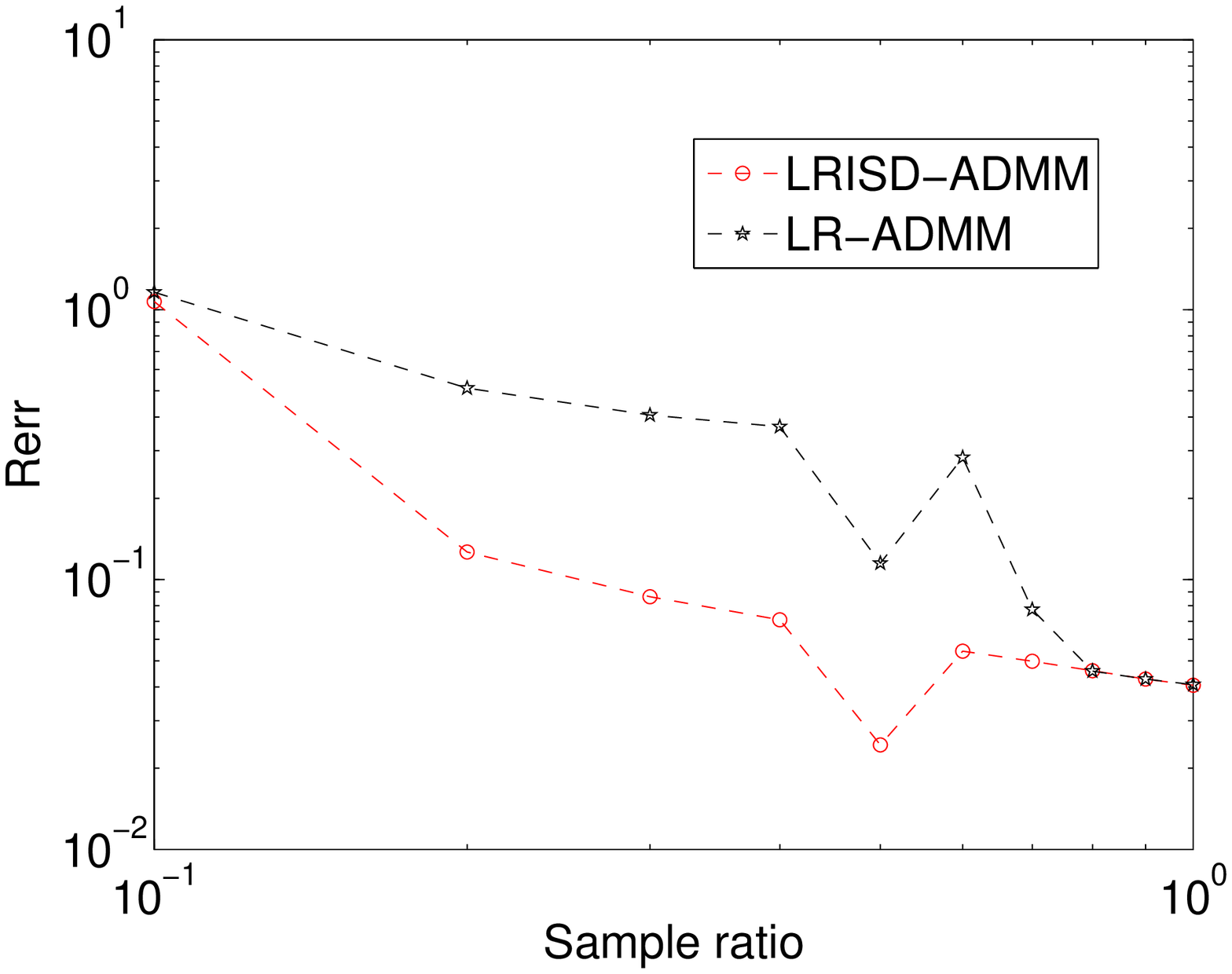}}
\hfil
\subfloat[]{\includegraphics[height=4.5cm,width=0.33\textwidth]{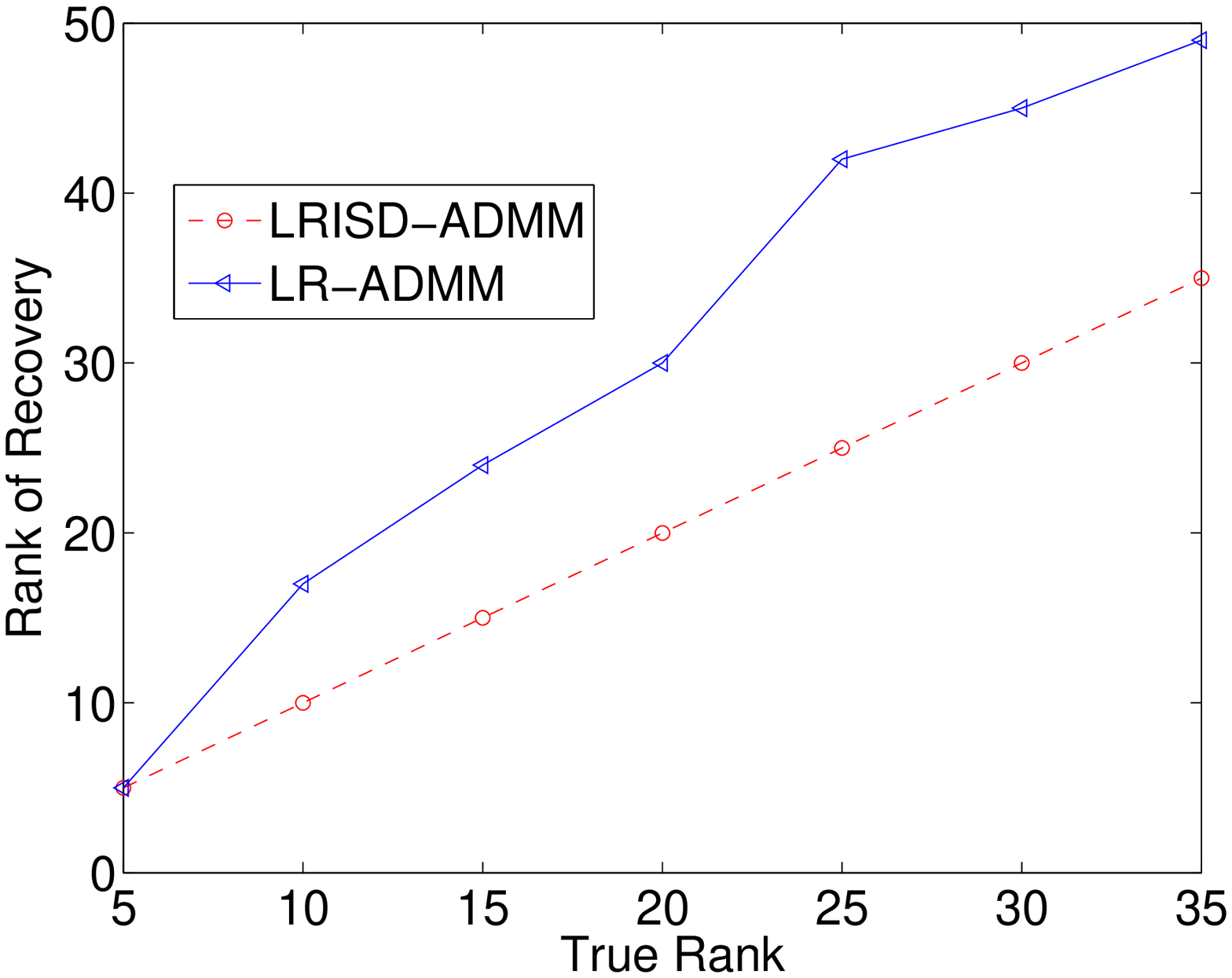}}
\caption{Comparison results of  LRISD-ADMM and LR-ADMM on synthetic data. Different noise levels ($std$) are shown in (a). (b) gives the results under different sample ratios($sr$). (c) shows the recovery ranks under different ranks ($r$). }
\label{fig:4}
\end{figure*}

We evaluate the recovery performance by the Relative Error as $Reer=\|X_{re}-X^{*}\|_{F}/\|X^{*}\|_{F}$. We compare the reconstruction error under different conditions: different noise levels ($std$) and different sample ratios ($sr$) taken which are shown respectively in Fig \ref{fig:4}(a) and Fig \ref{fig:4}(b). In addition, we compare the recovery ranks which are obtained via the above two algorithms in Fig \ref{fig:4}(c). For each scenario, we generated the model by 10 times and report the average results.
 \begin{itemize}
\item We fix the matrix size to be $m=n=300, r=15, sr=0.5$, and run LRISD-ADMM and LR-ADMM under different noise levels $std$. The results are shown in Fig \ref{fig:4}(a)
\item We fix the matrix size to be $m=n=300, std=0.5$, and run LRISD-ADMM and LR-ADMM under different $sr$. The results are shown in Fig \ref{fig:4}(b)
\item We set the matrix size to be $m=n=300, sr=0.5, std=0.5$, and run LRISD-ADMM and LR-ADMM under different $r$. The results are shown in Fig \ref{fig:4}(c)
\end{itemize}

It is easy to see from Fig \ref{fig:4}(a), as the noise level $std$ increases, the total $Reer$ becomes larger. Even so, LRISD-ADMM can achieve much better recovery performance than LR-ADMM. This is because the LRISD model better approximate the rank function than the nuclear norm. Thus, we illustrate that LRISD-ADMM is more robust to noise when it deals with low-rank matrices. With the increasing of sample ratio $sr$, the total $Reer$ reduces in Fig \ref{fig:4}(b). Generally, LRISD-ADMM does better than LR-ADMM, because LRISD-ADMM can approximately recover the rank of the matrix as showed in Fig \ref{fig:4}(c).
\subsection{\bf{The Comparison between LRISD-ADMM and LR-ADMM: Real Visual Data}}

\begin{figure*}[!th]
\centering
\subfloat[door: the first outer iteration of LRISD]{\includegraphics[height=4.2cm,width=0.33\textwidth]{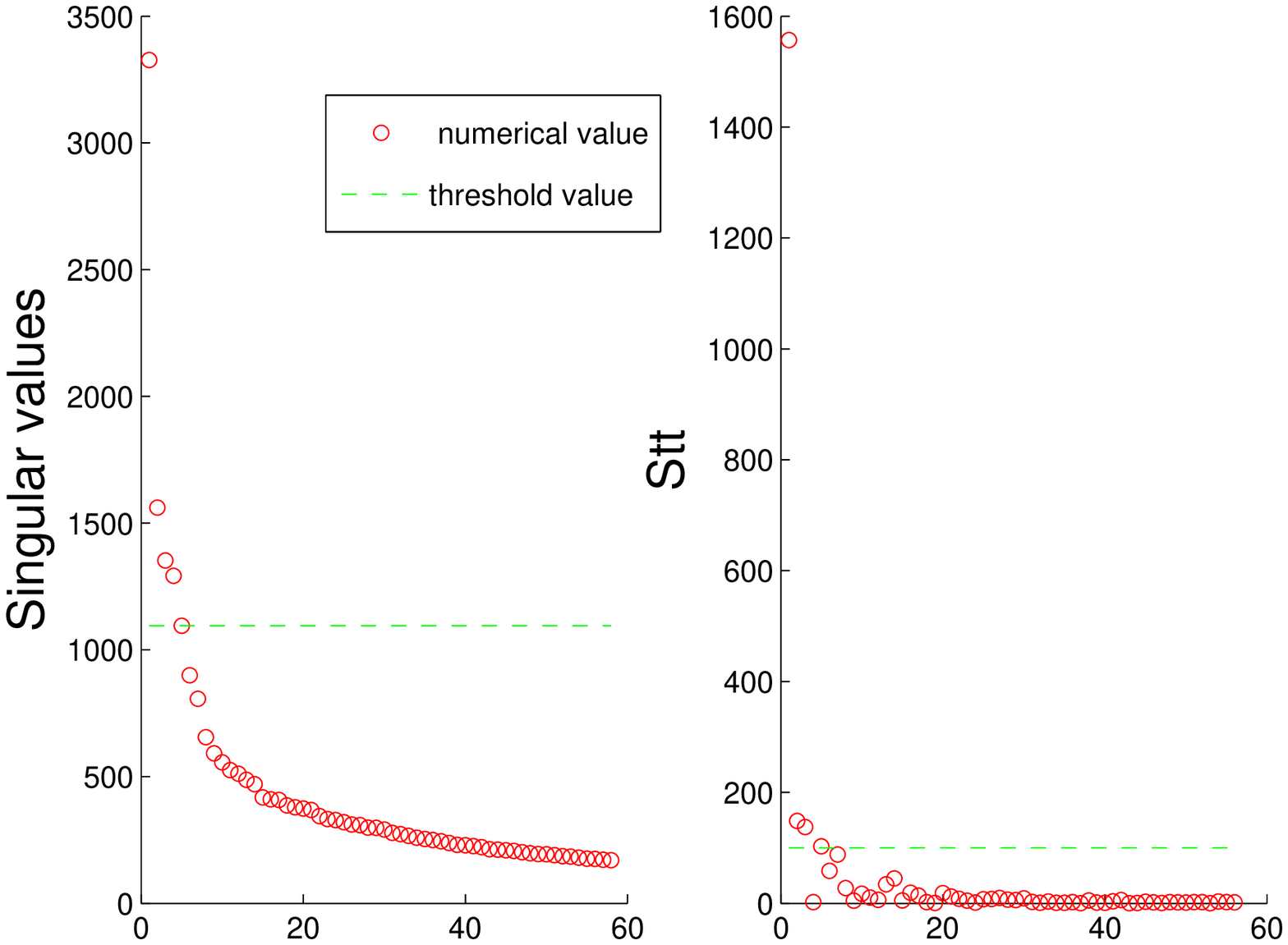}}
\hfil
\subfloat[door: the second outer iteration of LRISD]{\includegraphics[height=4.2cm,width=0.33\textwidth]{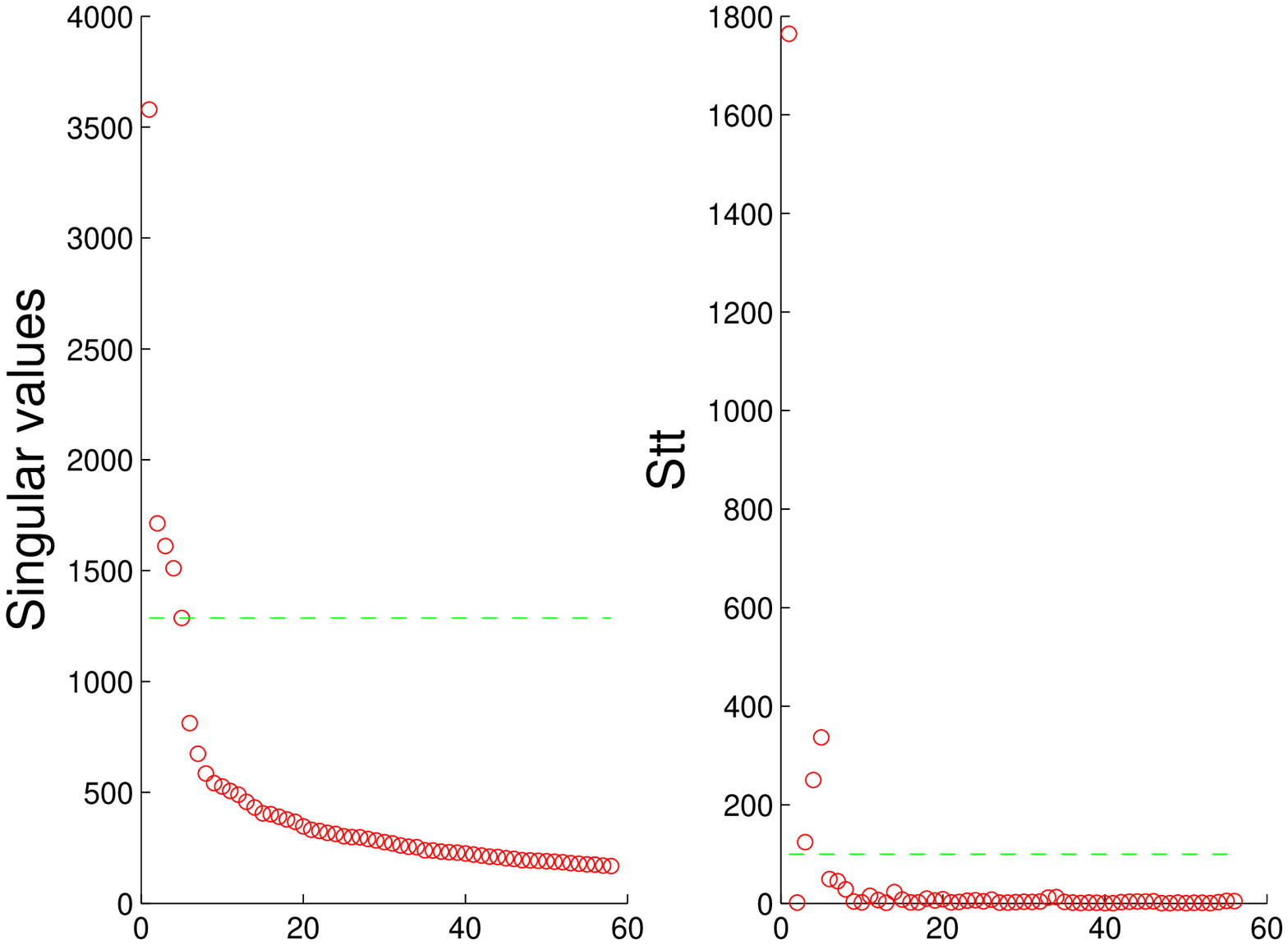}}
\hfil
\subfloat[door: the third outer iteration of LRISD]{\includegraphics[height=4.2cm,width=0.33\textwidth]{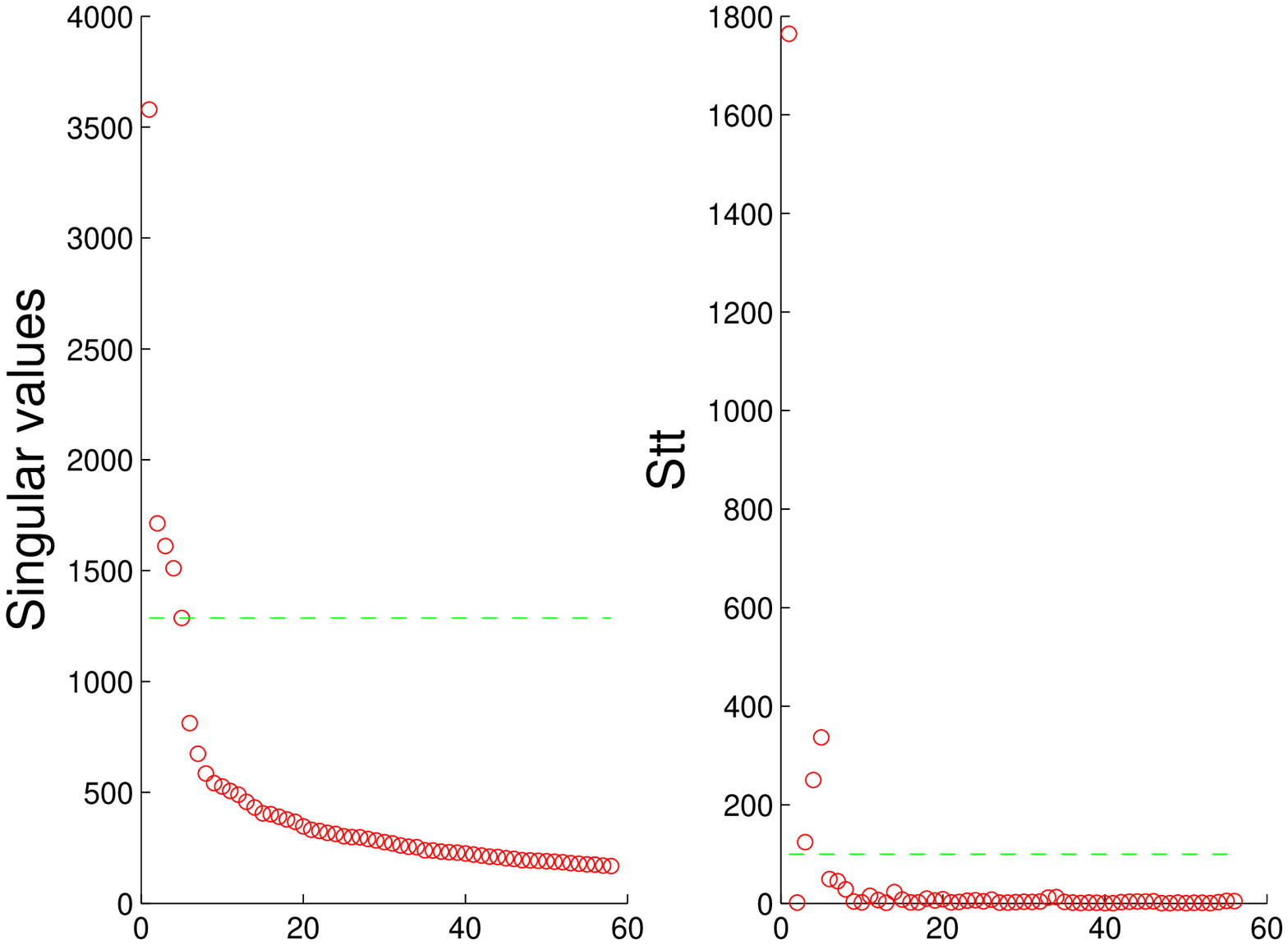}}\\
\subfloat[window: the first outer iteration of LRISD]{\includegraphics[height=4.2cm,width=0.33\textwidth]{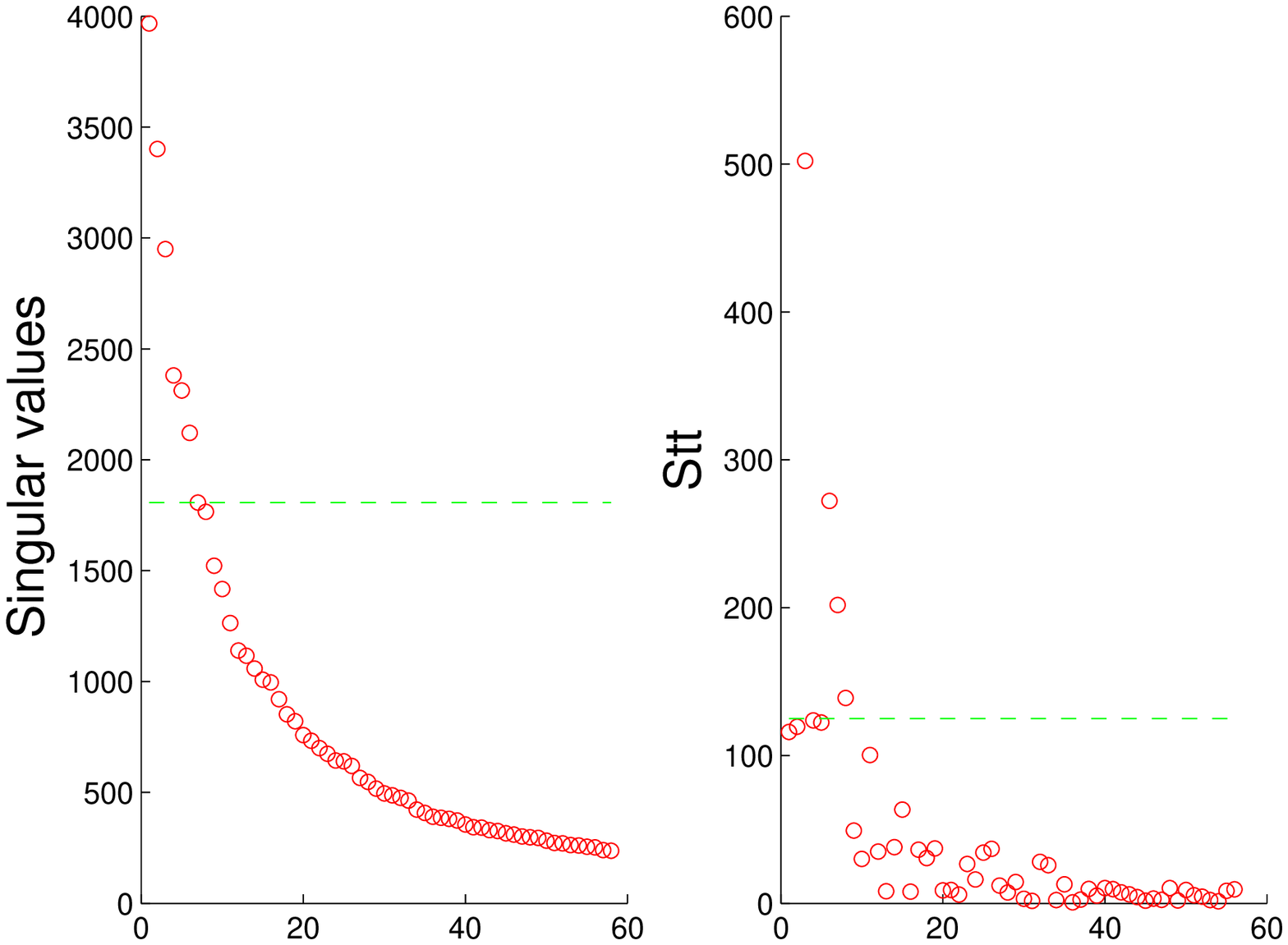}}
\hfil
\subfloat[window: the second outer iteration of LRISD]{\includegraphics[height=4.2cm,width=0.33\textwidth]{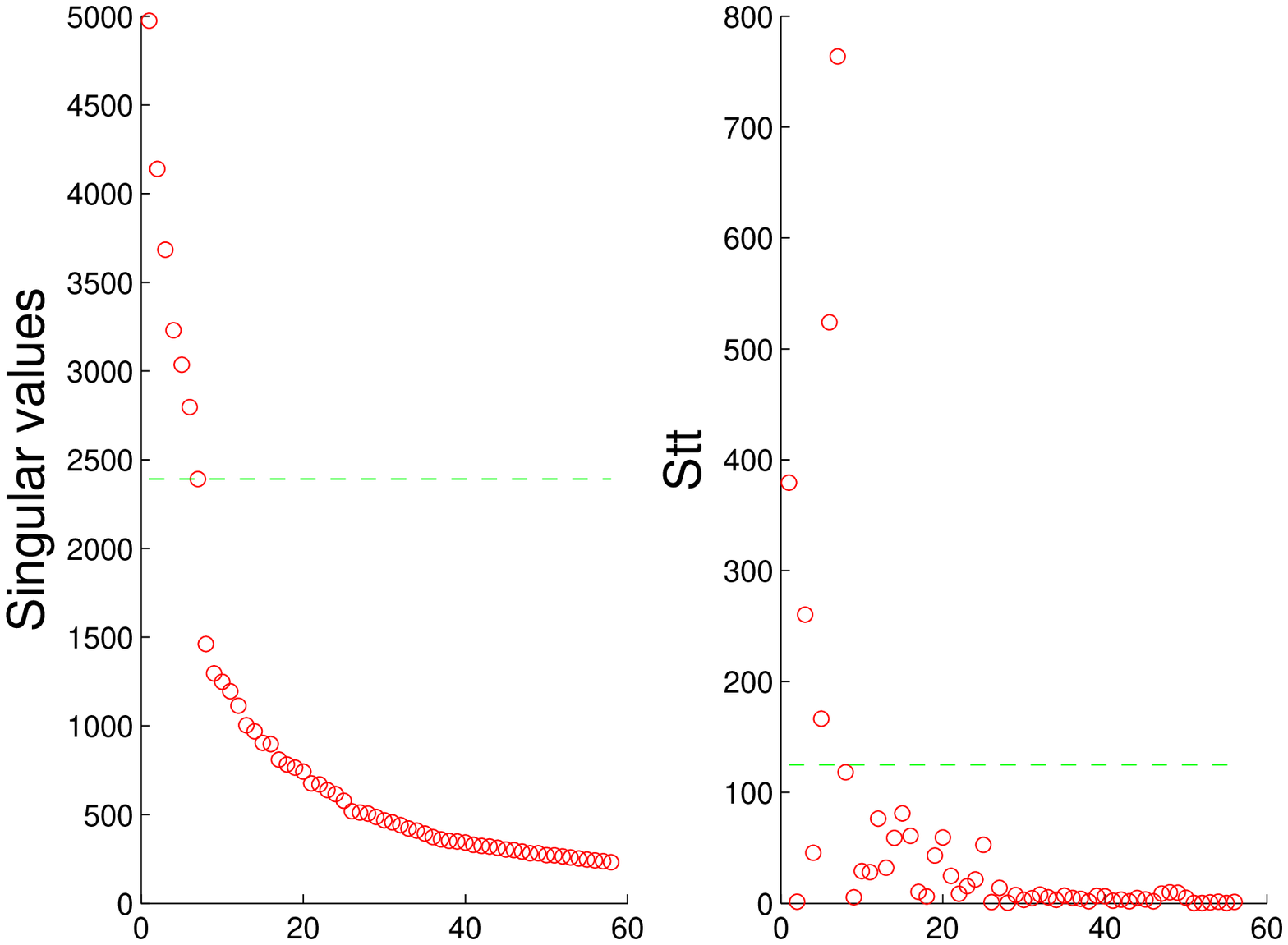}}
\hfil
\subfloat[window: the third outer iteration of LRISD]{\includegraphics[height=4.2cm,width=0.33\textwidth]{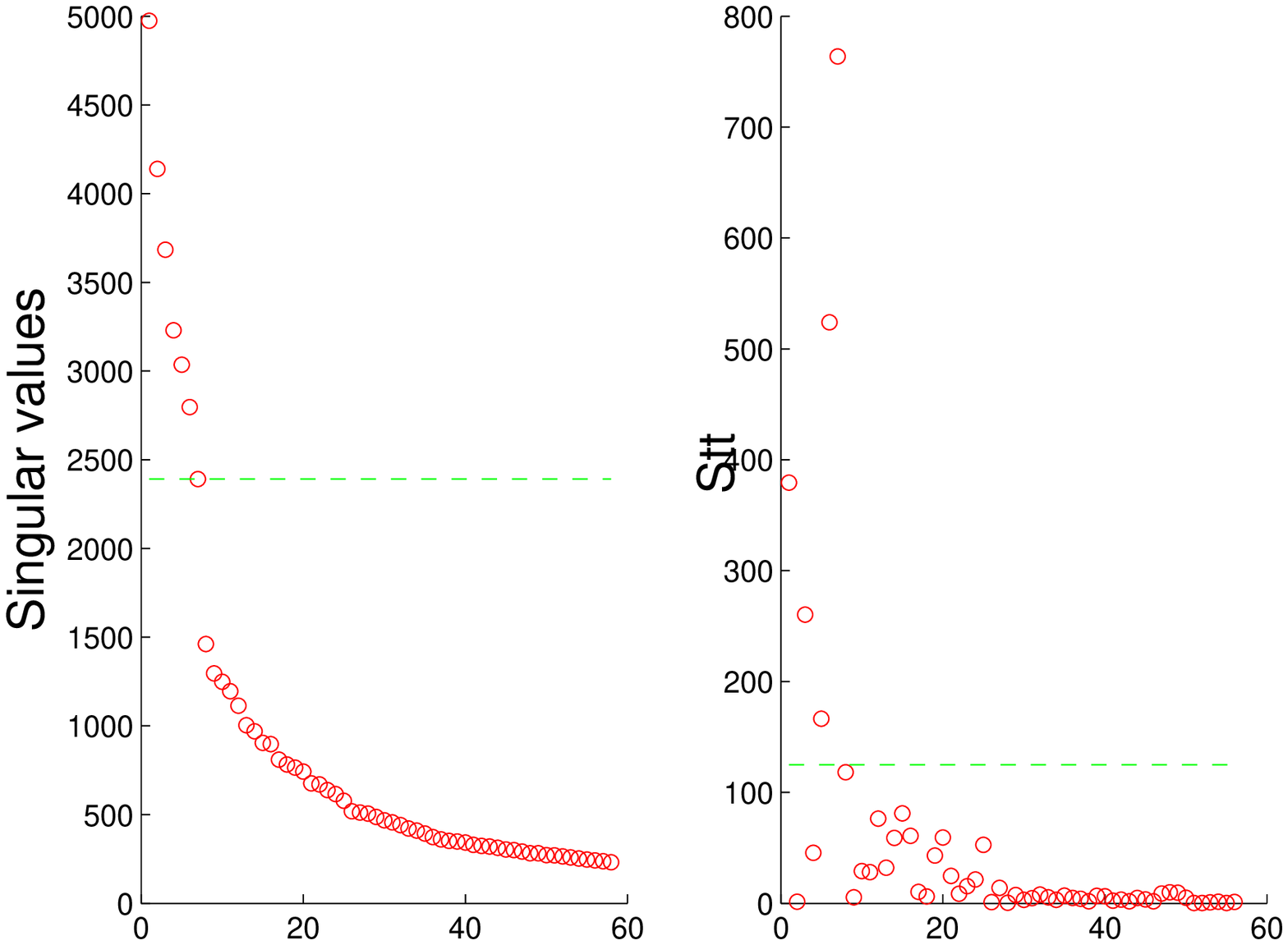}}\\
\subfloat[sea: the first outer iteration of LRISD]{\includegraphics[height=4.2cm,width=0.33\textwidth]{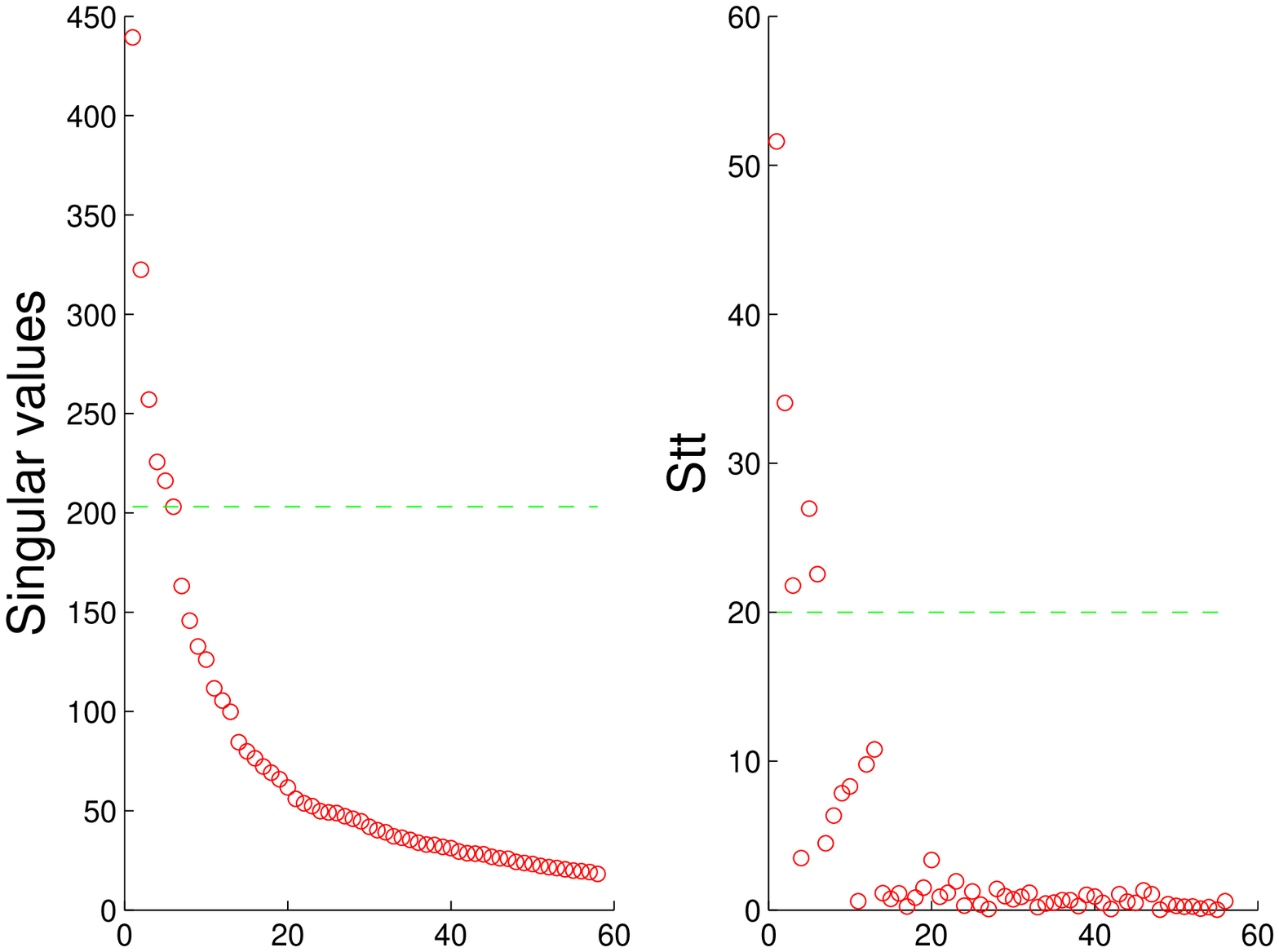}}
\hfil
\subfloat[sea: the second outer iteration of LRISD]{\includegraphics[height=4.2cm,width=0.33\textwidth]{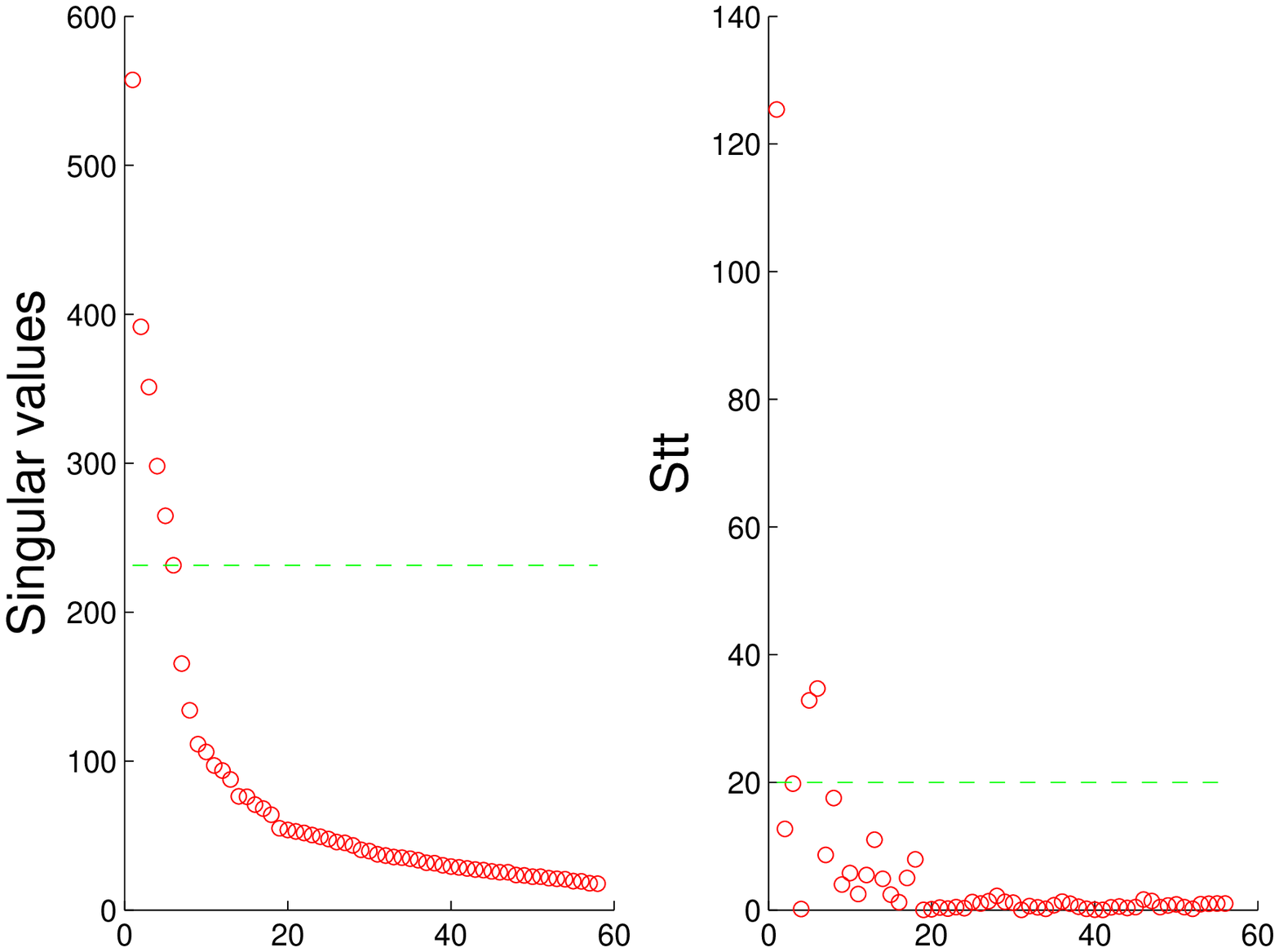}}
\hfil
\subfloat[sea: the third outer iteration of LRISD]{\includegraphics[height=4.2cm,width=0.33\textwidth]{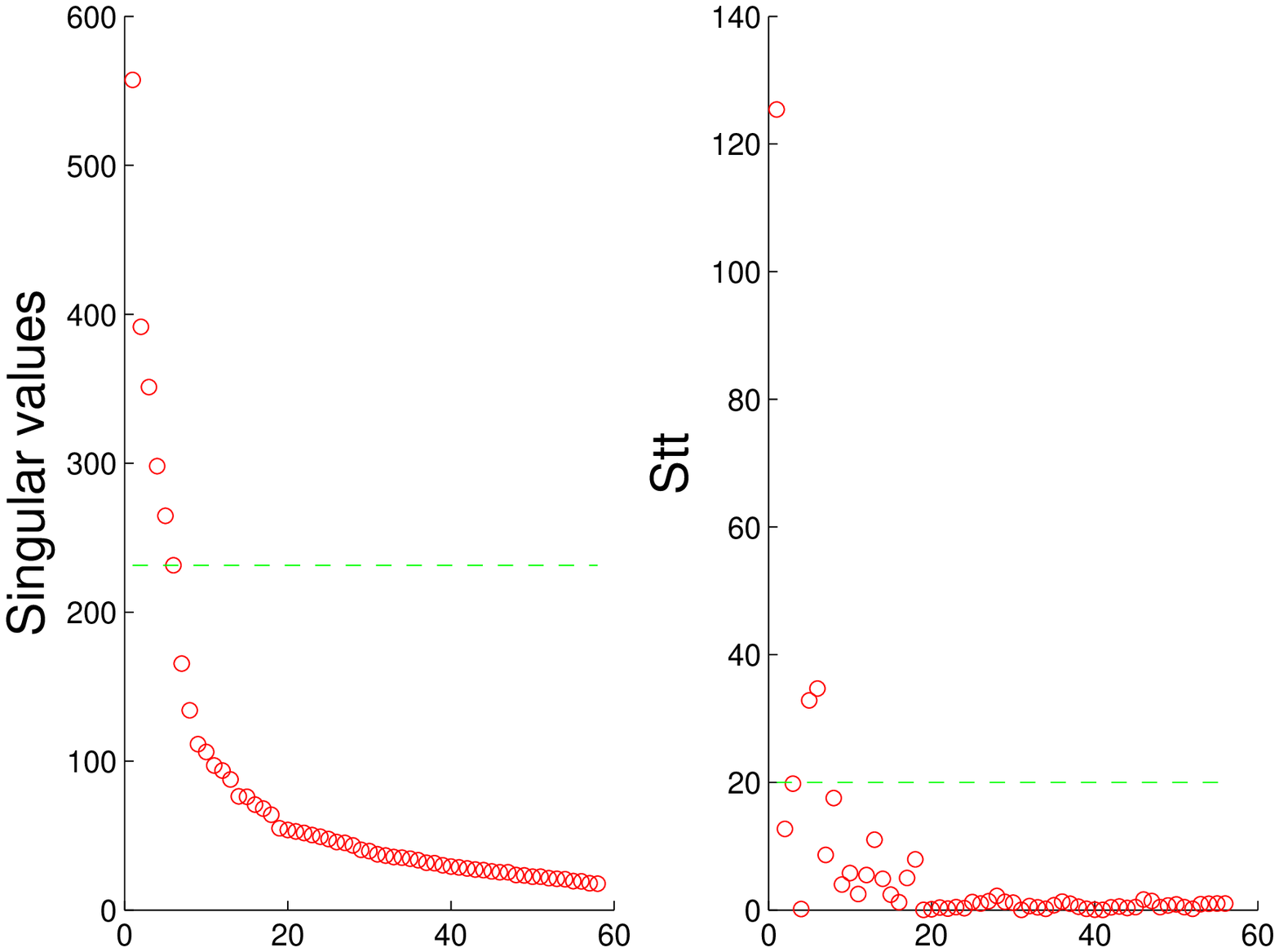}}
\caption{\small We present the process of  SVE in LRISD-ADMM about three images in Fig \ref{fig:6}. At the same time, we note that the first and the second singular values are much larger than others , as well as the values of Stt. To make the results more clear, we omit the first and the second singular values and $Stt$  in each figure. We can find the observed estimated $\tilde{r}$ are 7, 9, 8. Compared to the best $r$, which are 8,10,7, estimated $\tilde{r}$ is approximately equivalent to the best $r$. }
\label{fig:5}
\end{figure*}

\begin{figure*}[!th]
\centering
\subfloat[Original image]{\includegraphics[width=0.23\textwidth]{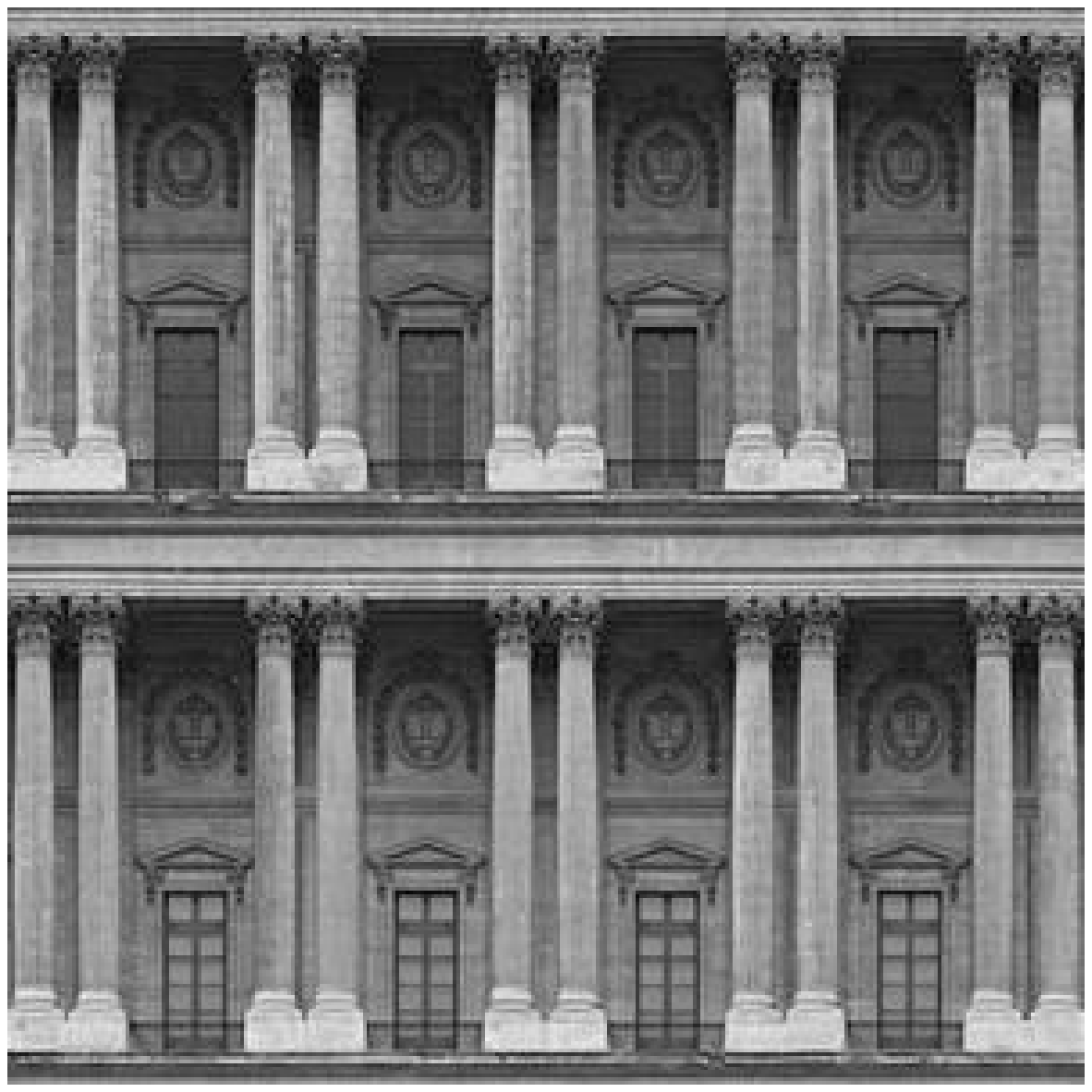}}
\hfil
\subfloat[Masked image]{\includegraphics[width=0.23\textwidth]{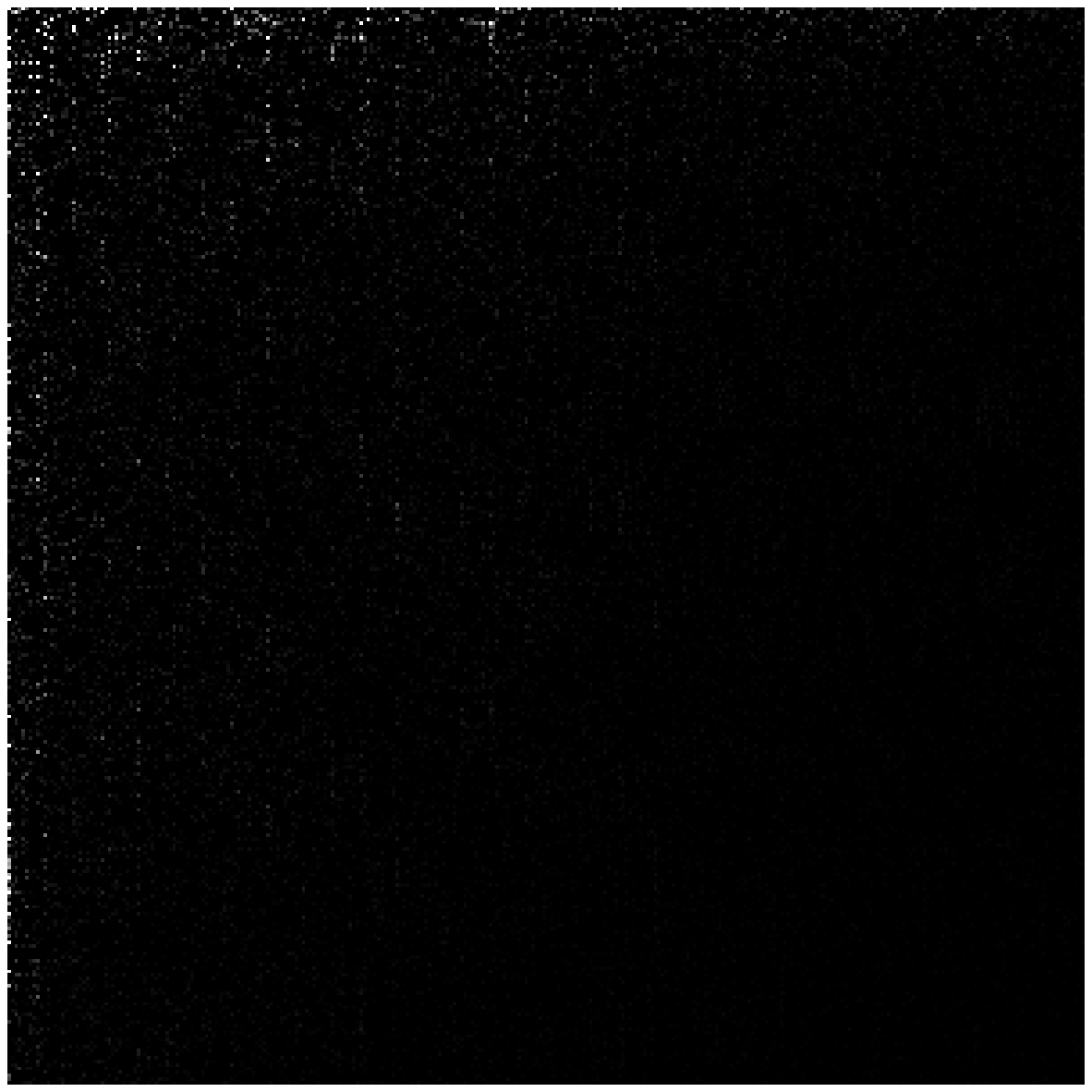}}
\hfil
\subfloat[LR-ADMM PSNR=18.779]{\includegraphics[width=0.23\textwidth]{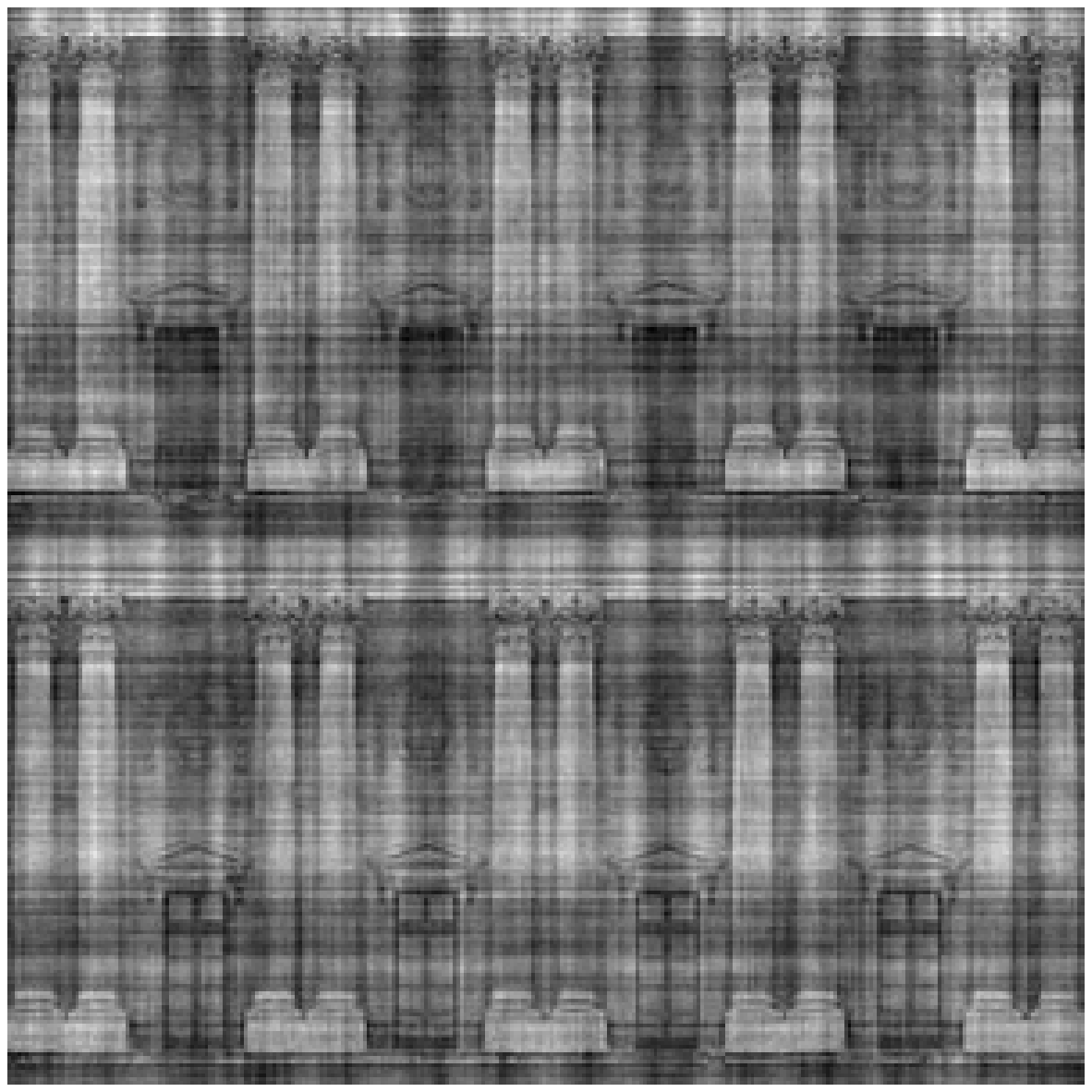}}
\hfil
\subfloat[LRISD-ADMM PSNR=19.109]{\includegraphics[width=0.23\textwidth]{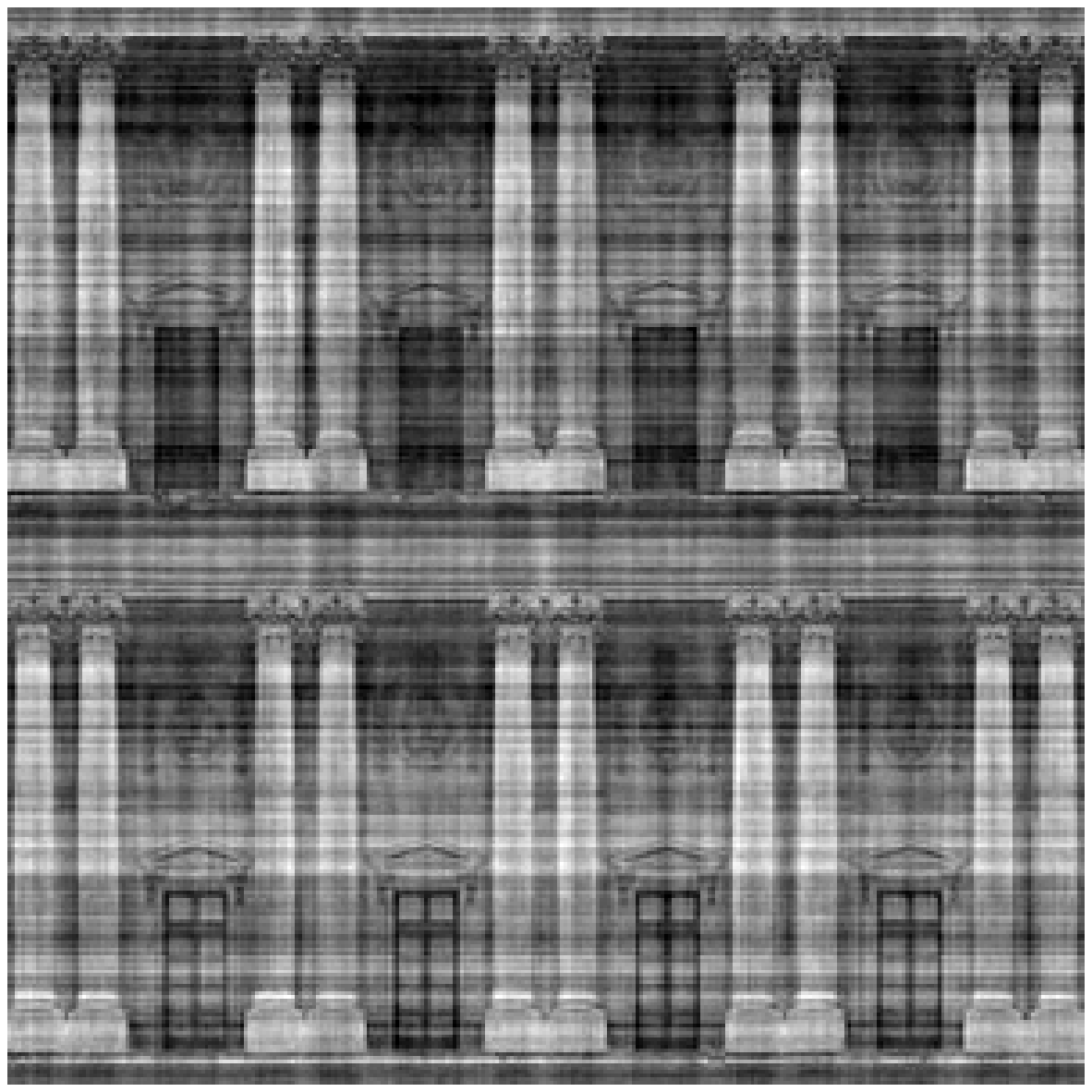}}\\
\subfloat[Original image]{\includegraphics[width=0.23\textwidth]{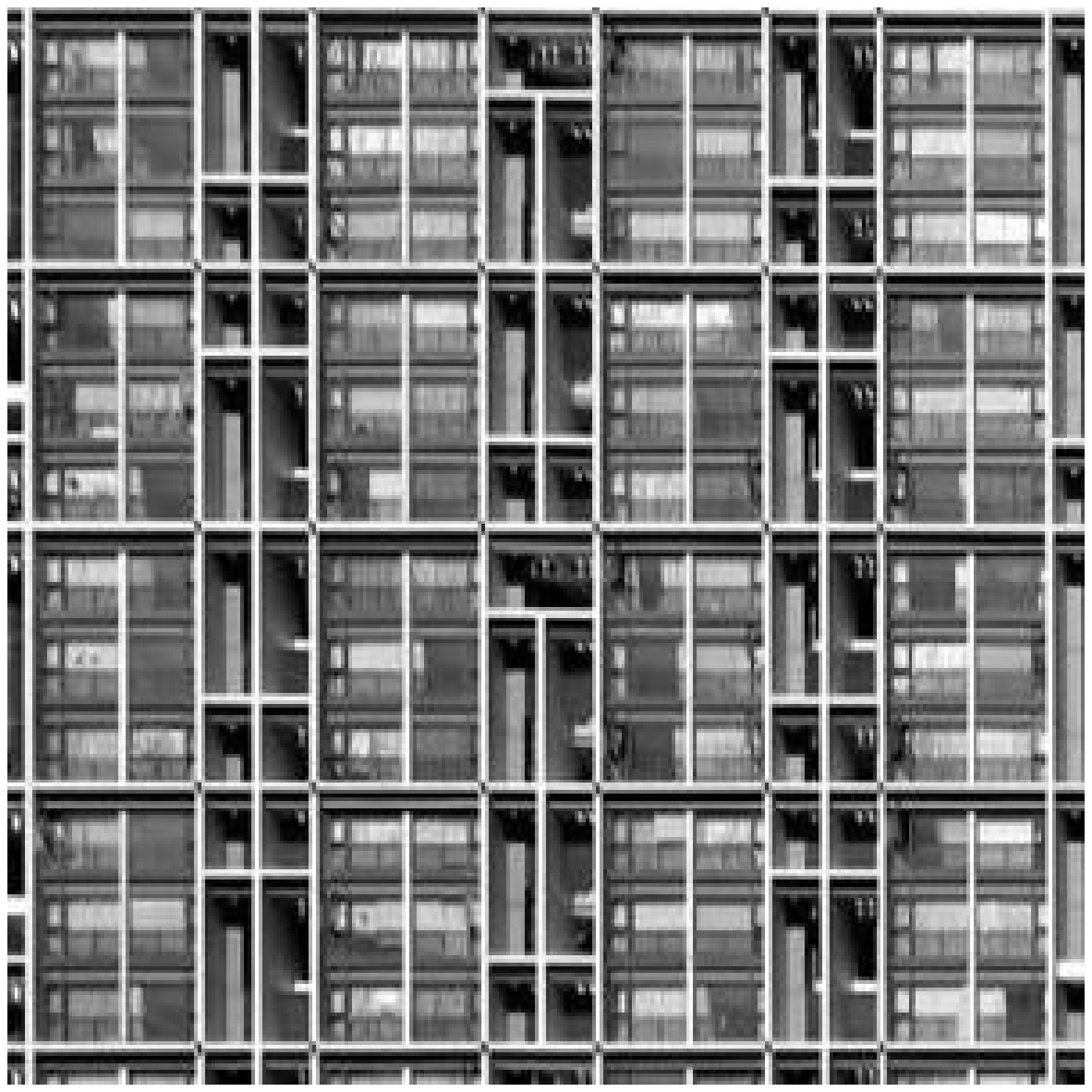}}
\hfil
\subfloat[Masked image]{\includegraphics[width=0.23\textwidth]{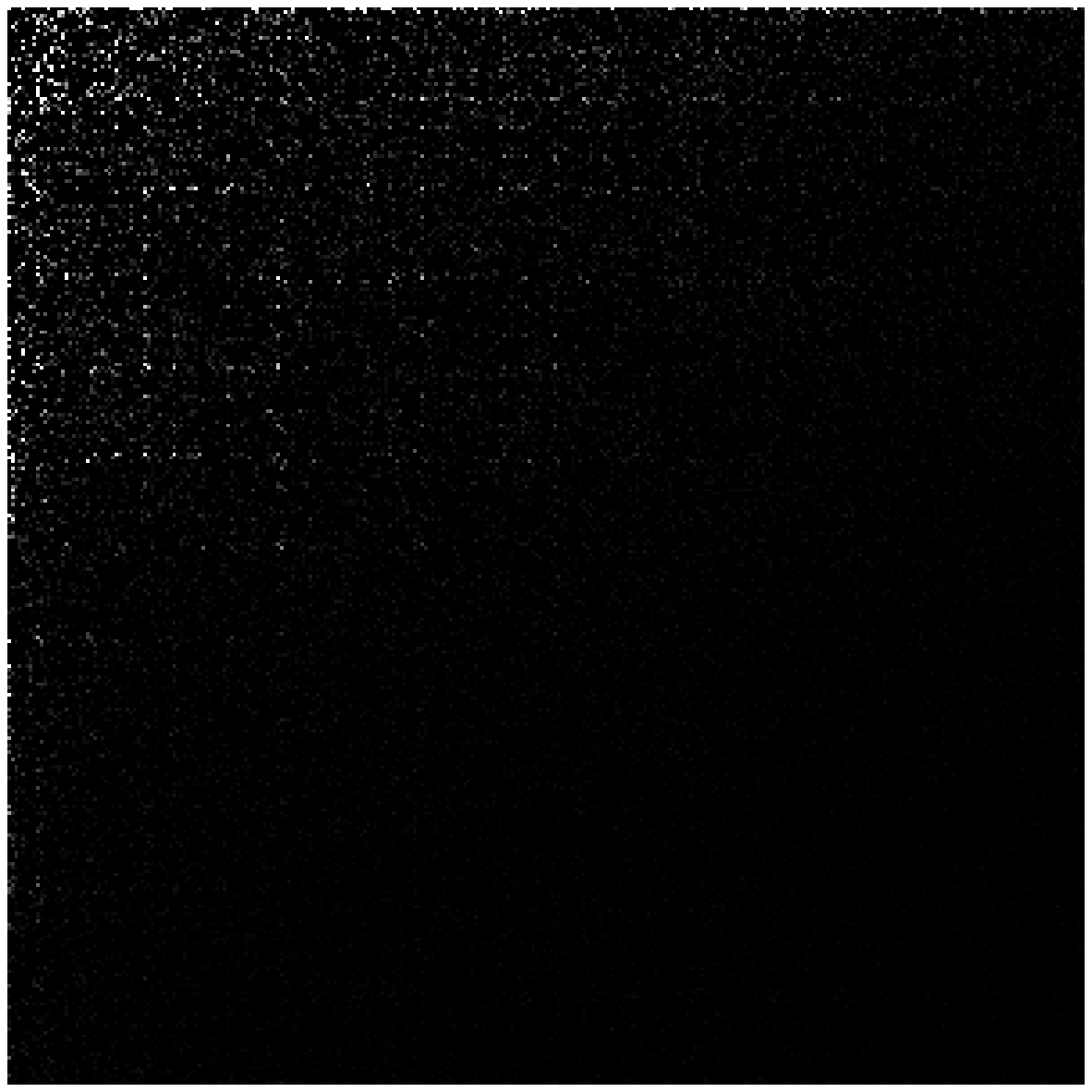}}
\hfil
\subfloat[LR-ADMM PSNR=17.140]{\includegraphics[width=0.23\textwidth]{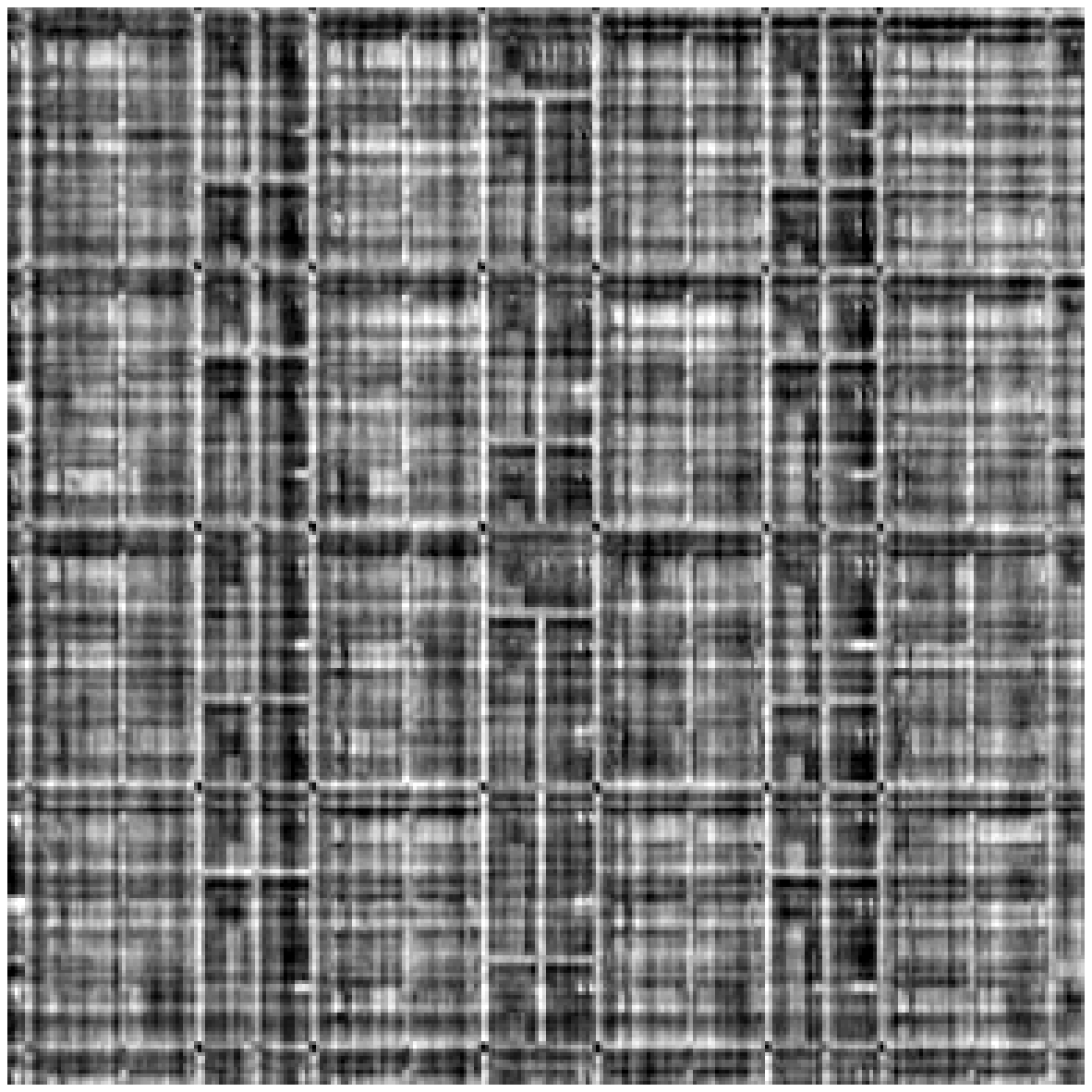}}
\hfil
\subfloat[LRISD-ADMM PSNR=17.608]{\includegraphics[width=0.23\textwidth]{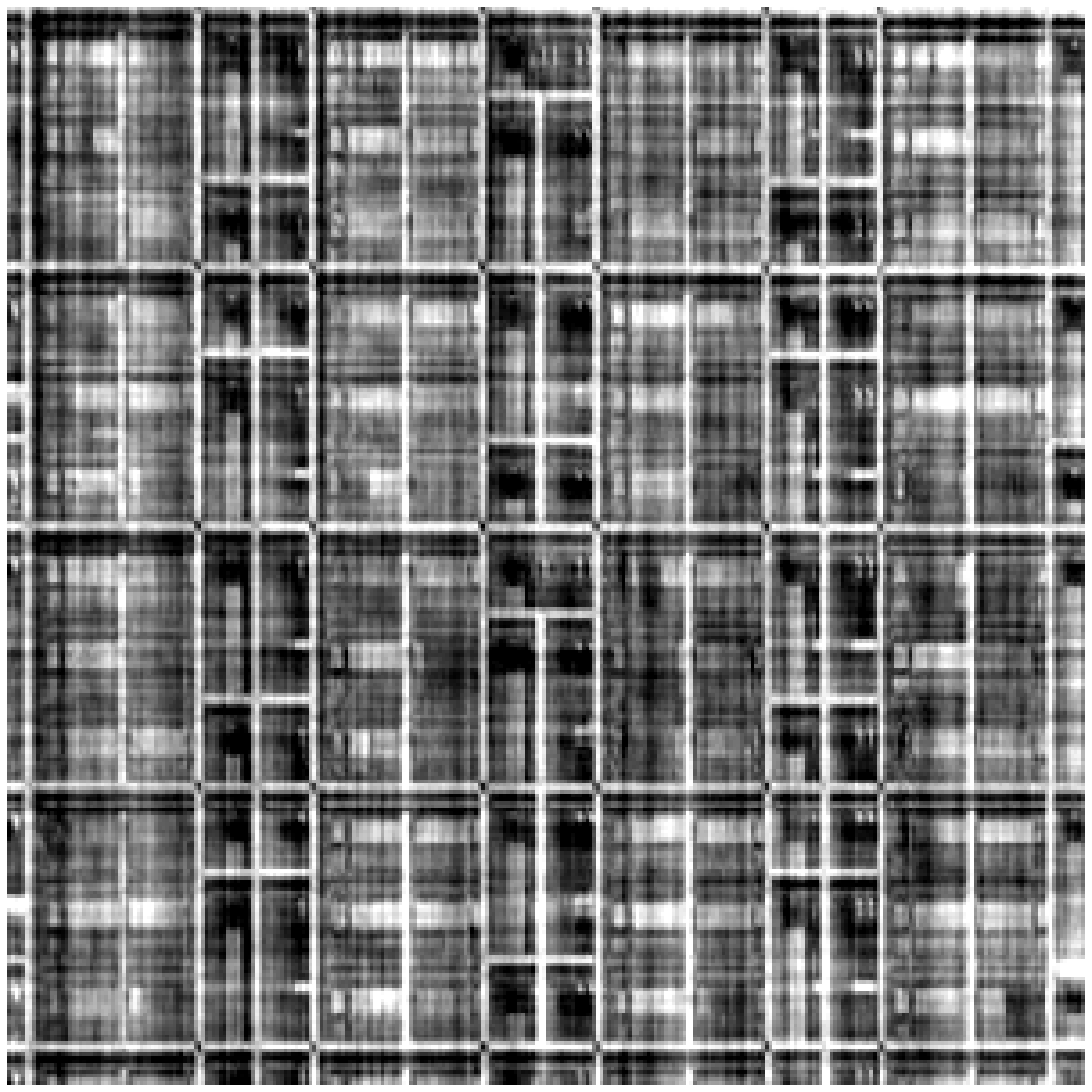}}\\
\hfil
\subfloat[Original image]{\includegraphics[width=0.23\textwidth]{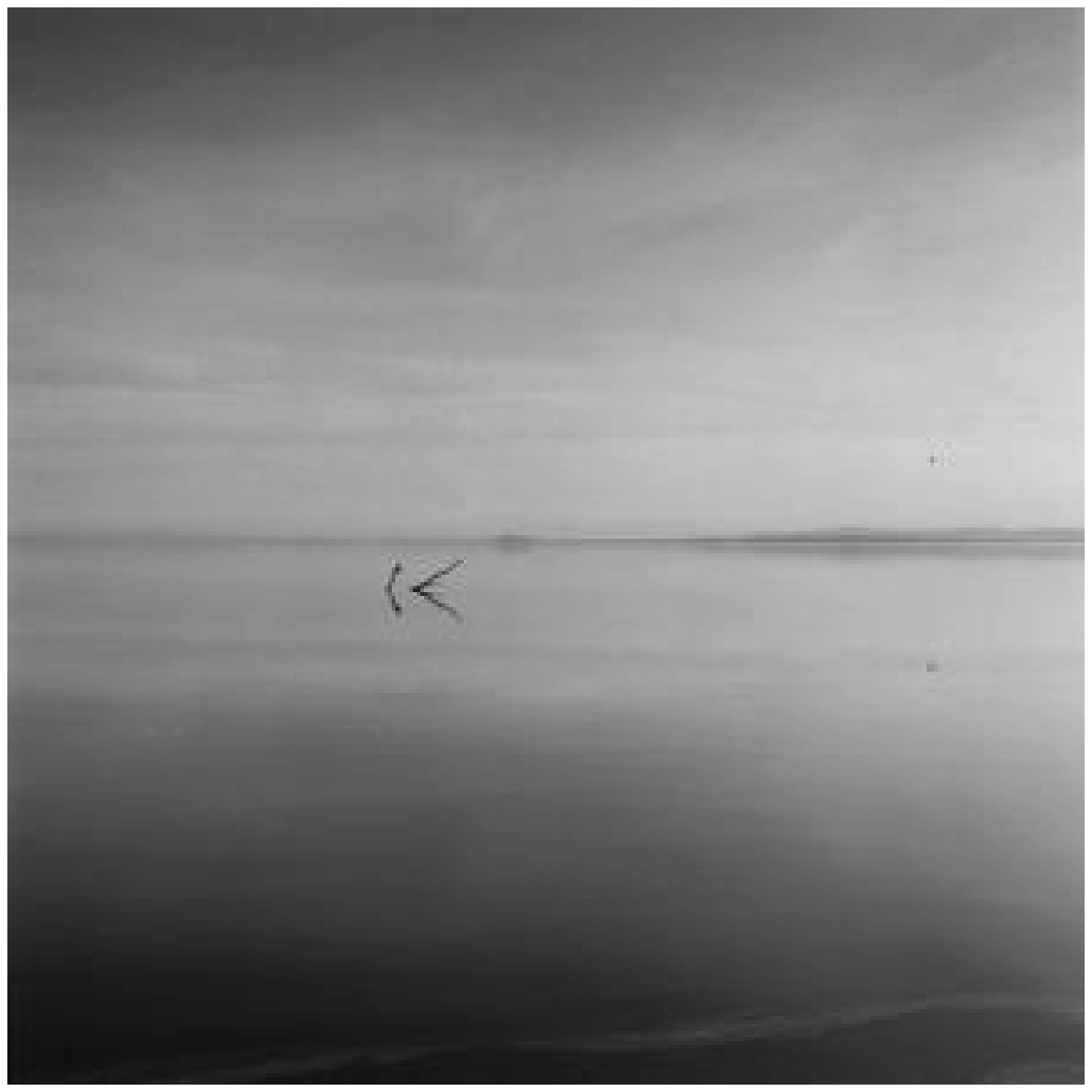}}
\hfil
\subfloat[Masked image]{\includegraphics[width=0.23\textwidth]{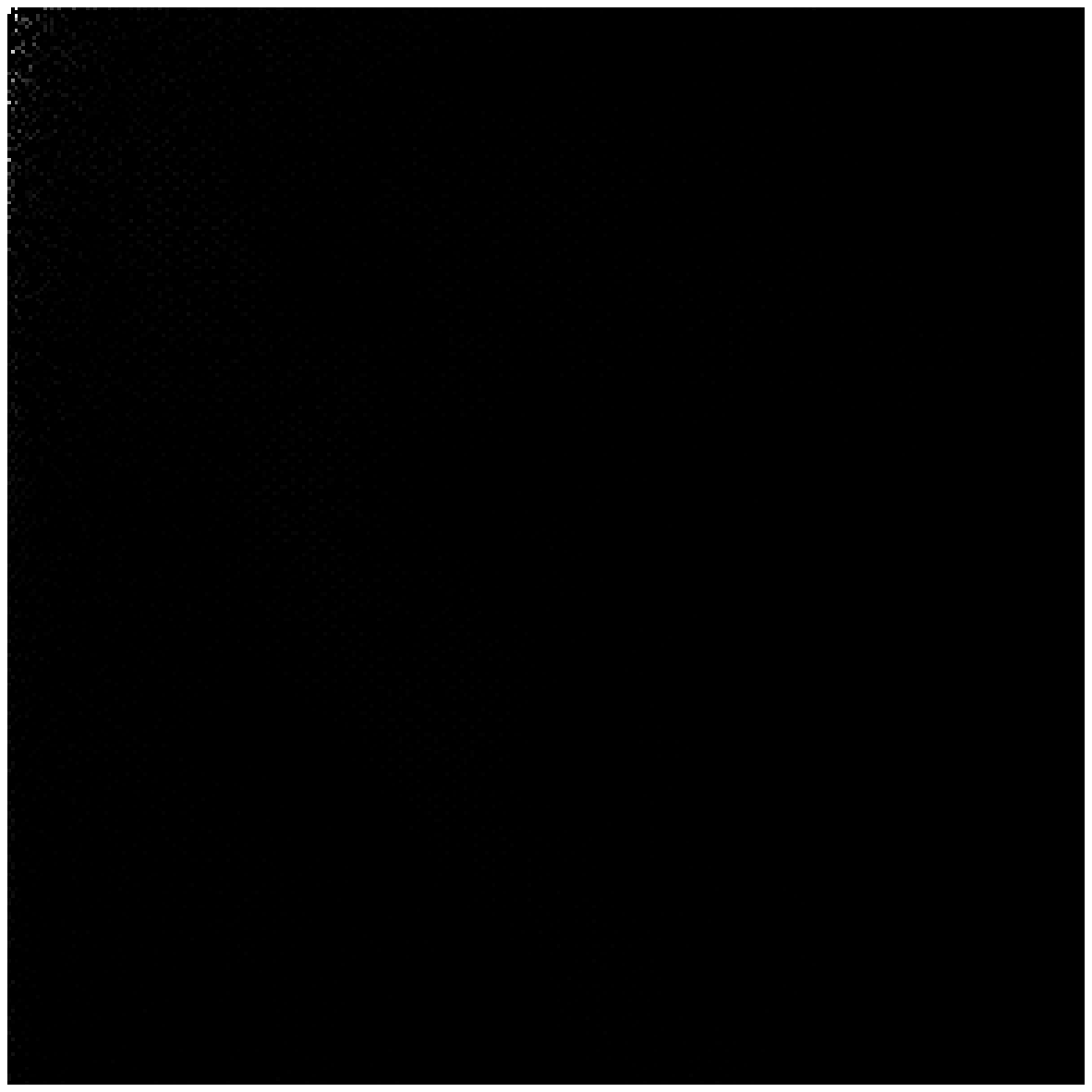}}
\hfil
\subfloat[LR-ADMM PSNR=16.829]{\includegraphics[width=0.23\textwidth]{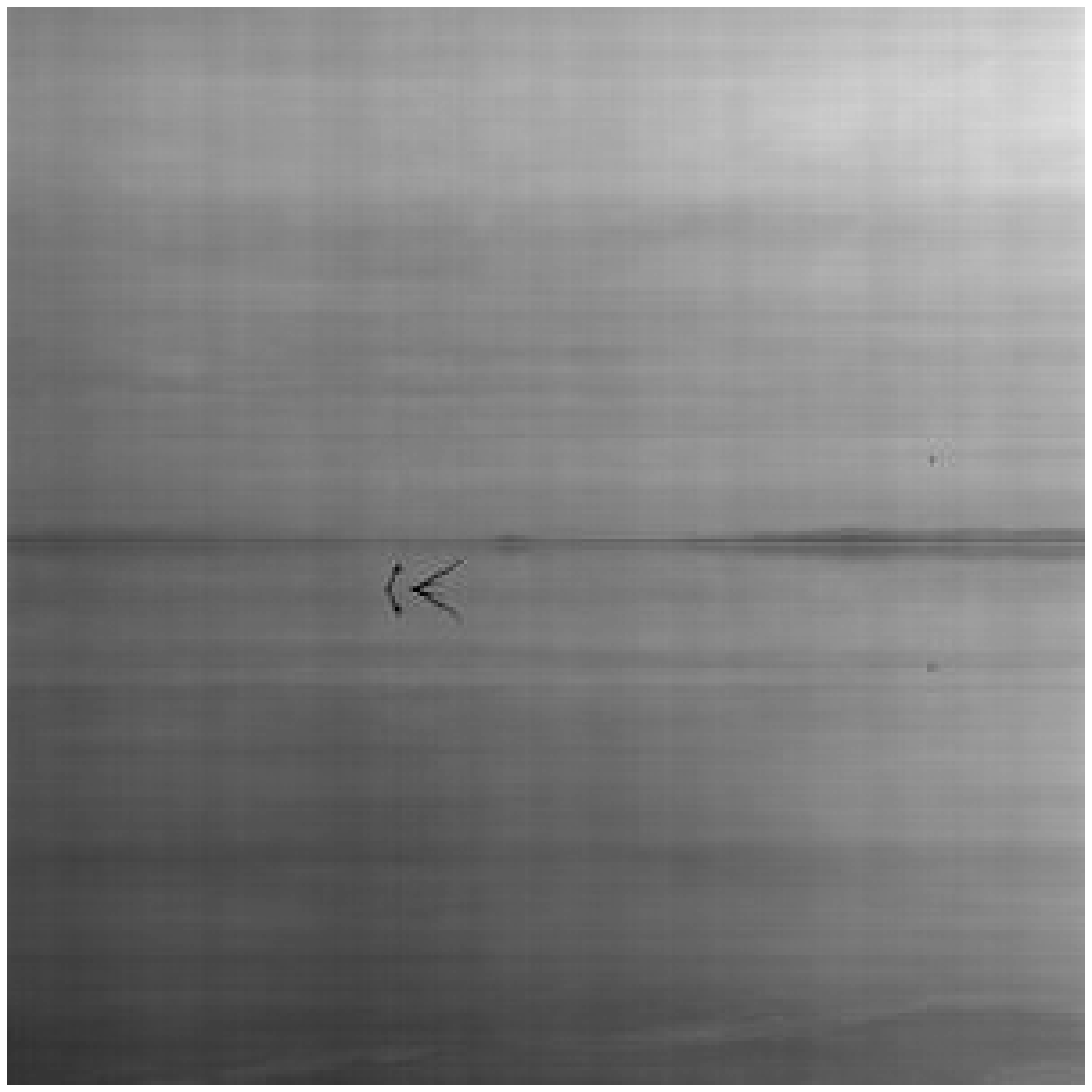}}
\hfil
\subfloat[LRISD-ADMM PSNR=16.960]{\includegraphics[width=0.23\textwidth]{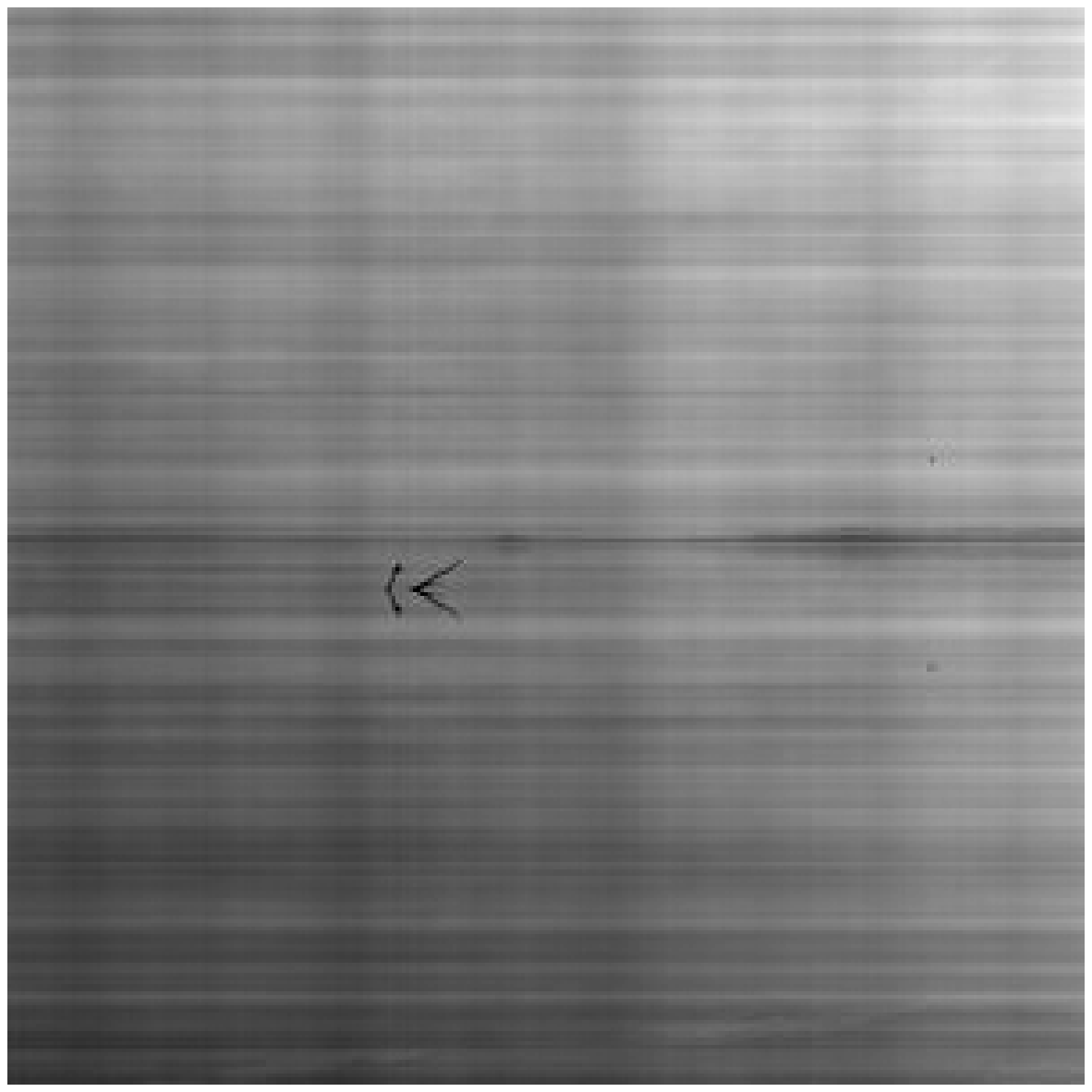}}
\caption{\small Comparison results of  LR-ADMM  and LRISD-ADMM, we use three images here. The masked images is obtained by making partial DCT on the original images. Beside, images recovered by LR-ADMM  and LRISD-ADMM method are displayed.}
\label{fig:6}
\end{figure*}

In this subsection, we test three images: door, window and sea, and compare the recovery images by  LRISD-ADMM and general LR-ADMM on the partial DCT operator. In all tests, we fix $sr=0.6, std=0$. The SVE process during different stages to obtain $\tilde{r}$ is depicted in Fig \ref{fig:5}. For three images, we set $\kappa=100, 125, 20$ and generate $\tilde{r}=8, 10, 7$ in LRISD-ADMM. Moreover, Fig \ref{fig:6} shows the recovery results of the two algorithms.

As illustrated in Fig \ref{fig:5}, it is easy to see SVE returns an stable $\tilde{r}$ in merely three iterations. And, the estimated $\tilde{r}$ is a good estimate to the number of largest few singular values. 
From Fig \ref{fig:6}, it can be seen that LRISD-ADMM outperforms the general LR-ADMM in terms of smaller PSNR. More important, using eyeballs, we can see the
better fidelity of the recoveries of LRISD-ADMM to the true signals, in  terms of better recovering sharp edges. 

\subsection{\bf{A note on $\kappa$}}\label{subsection6F}

We note that the thresholding $\kappa$ plays a critical rule for the efficiency of the proposed SVE. For real visual data, we can use $\kappa=\sqrt{m*n}/(3*s), s=0.5,0.8,1,3,5$. For synthetic data, $\kappa$ is denoted as $\kappa=s*\sqrt{m*n}/30, s=1,2,3$. The above heuristic, which works well in our experiments, is certainly not necessarily optimal; on the other hand, it has been observed that LRISD is not very sensitive to $\kappa$.  Of course, the ``last significant jump'' based thresholding is only one way for estimate the number of the true nonzero (or large) singular values, and 
%
one can try other available effective jump detection methods \citep{wang2010sparse,yasakov1994method,durante2007iterative}. 

\section{Conclusion}\label{Conclusion}

This paper introduces the singular values estimation (SVE) to estimate a appropriate $r$ (in $\|X\|_{r}$) that the estimated rank is (approximately) equivalent to the best rank. In addition, we extend TNNR from matrix completion to the general low-rank cases (we call it LRISD). Both synthetic and real visual data sets are discussed. 
Notice that 
 SVE is not limited to thresholding. Effective support detection guarantees the good performance of LRISD. Therefore future research includes studying specific signal classes and developing more effective support detection methods.


\section*{Acknowledgment}

This work was supported by the Natural Science Foundation of China, Grant
Nos. 11201054, 91330201  and by the Fundamental Research Funds for the Central Universities ZYGX2012J118.


\bibliographystyle{model1-num-names}
\bibliography{wosvd1}







\end{document}